\documentclass[3p, times]{elsarticle}

\usepackage{hyperref}
\usepackage{amsmath}
\usepackage{amsfonts}
\usepackage{natbib}
\usepackage{comment}
\usepackage[ruled,vlined]{algorithm2e}
\usepackage{subcaption}
\usepackage{bm}
\usepackage{enumitem}
\usepackage{lipsum}
\usepackage{booktabs}
\usepackage{threeparttable}
\usepackage{pgfplots}
\pgfplotsset{compat=newest}
\usepgfplotslibrary{external}
\usepgfplotslibrary{patchplots}
\usetikzlibrary{shapes,arrows}
\usetikzlibrary{positioning,calc}
\usetikzlibrary{decorations.pathreplacing,decorations.markings,math,arrows.meta}
\usetikzlibrary{spy}
\usetikzlibrary{pgfplots.groupplots}

\usepackage{eso-pic}
\usepackage{cleveref}
\newcommand{\state}{\mathbf{x}}

\newcommand{\statePred}{\mathbf{\widetilde{x}}}

\newcommand{\param}{\mathbf{p}}

\newcommand{\statedt}{\dot{\mathbf{x}}}

\newcommand{\embedding}{\mathbf{e}}

\newcommand{\dimState}{D_{\mathbf{x}}}

\newcommand{\dimParam}{D_{\mathbf{p}}}

\newcommand{\dimE}{D_{\mathbf{e}}}

\newcommand{\nE}{N_{\mathbf{e}}}

\newcommand{\RR}{\mathbb{R}}

\newcommand{\np}{N_p}
\newcommand{\nics}{N_{ics}}
\newcommand{\ntimesteps}{N_{T}}

\newcommand{\trainData}{\mathbf{X}}

\newcommand{\augState}{\mathbf{u}}

\newcommand{\isl}{\text{ISL}}

\newcommand{\loss}{\mathcal{L}}

\newcommand{\nrmse}{\mathrm{NRMSE}}

\newcommand{\ttt}{\mathrm{TtT}}

\newcommand{\trainnoise}{\sigma_{\text{noise}}}

\newcommand{\nicsTrain}{\nics^{\text{train}}}

\newcommand{\nicsTest}{\nics^{\text{test}}}

\newcommand{\pTrainSet}{P_{\text{train}}}

\newcommand{\pTestInterp}{P_{\text{test}}^{\text{interp}}}
\newcommand{\pTestExtrap}{P_{\text{test}}^{\text{extrap}}}

\newcommand{\hnn}{\operatorname{HNN}}
\usepackage{float}

\usepackage{amsthm}
\newtheorem{proposition}{Proposition}
\newtheorem{remark}{Remark}

\def\equationautorefname~#1\null{Equation~(#1)\null}

\begin{document}

\AddToShipoutPictureBG{%
  \AtPageLowerLeft{%
    \hspace{2.5cm}\raisebox{1.5cm}{%
      \large\textbf{Preprint} \large-- submitted to \textit{Mechanical Systems and Signal Processing}%
    }%
  }%
}

\begin{frontmatter}

\title{
Beyond Static Models: Hypernetworks for Adaptive and Generalizable Forecasting in Complex Parametric Dynamical Systems
}

\author[mainaddress]{Pantelis R. Vlachas \corref{correspondingauthor}}
\cortext[correspondingauthor]{Corresponding author}
\ead{pvlachas@ethz.ch}
\author[mainaddress]{Konstantinos Vlachas}
\ead{vlachas@ibk.baug.ethz.ch}
\author[mainaddress]{Eleni Chatzi}
\ead{chatzi@ibk.baug.ethz.ch}



\address[mainaddress]{
Department of Civil, Environmental, and Geomatic Engineering, ETH Z\"{u}rich, Stefano-Franscini Platz 5, 8049, Zürich, Switzerland
}

\begin{abstract}
Dynamical systems play a key role in modeling, forecasting, and decision-making across a wide range of scientific domains. 
However, variations in system parameters, also referred to as parametric variability, can lead to drastically different model behavior and output, posing challenges for constructing models that generalize across parameter regimes.
In this work, we introduce the Parametric Hypernetwork for Learning Interpolated Networks (PHLieNet), a framework that simultaneously learns: (a) a global mapping from the parameter space to a nonlinear embedding and (b) a mapping from the inferred embedding to the weights of a dynamics propagation network.
The learned embedding serves as a latent representation that modulates a base network, termed the \textit{hypernetwork}, enabling it to generate the weights of a \textit{target network} responsible for forecasting the system's state evolution conditioned on the previous time history.
By interpolating in the space of models rather than observations, PHLieNet facilitates smooth transitions across parameterized system behaviors, enabling a unified model that captures the dynamic behavior across a broad range of system parameterizations.
The performance of the proposed technique is validated in a series of dynamical systems with respect to its ability to extrapolate in time and interpolate and extrapolate in the parameter space, i.e., generalize to dynamics that were unseen during training.
Our approach outperforms state-of-the-art baselines in both short-term forecast accuracy and in capturing long-term dynamical features such as attractor statistics.
\end{abstract}


\end{frontmatter}


\section{Introduction}
\label{sec:intro}

Accurate modeling and inference of the behavior of complex dynamical systems is essential for understanding, predicting, and controlling real-world phenomena across disciplines.
While recent work in data-driven modeling has advanced our ability to learn the governing dynamics of complex systems, typically represented by ordinary and partial differential equations (ODEs and PDEs), most approaches focus mainly on accounting for variations due to the influence of initial conditions~\cite{vinuesa2022enhancing, brunton2024promising}.
Yet, an equally critical aspect is \textit{parametric variability}, which arises from changes in intrinsic system properties or external excitation characteristics~\cite{benner2015survey}.
Examples include the influence of the Reynolds number on flow regime transitions in fluid dynamics~\cite{schultz2013reynolds}, and the role of carbon dioxide levels or solar radiation in shaping long-term climate trends~\cite{eyring2019taking}.
Such examples underscore the ubiquity of parametric systems, where dynamics are governed not only by initial conditions but also by smoothly or abruptly varying parameters.

Traditional physics-based approaches for parametric modeling rely on the derivation of mathematical models based on first principles and include projection-based~\cite{agathos2024accelerating, amsallem2016pebl, benner2020model, barnett2022quadratic} and decomposition-based strategies~\cite{proctor2016dynamic,hesthaven2016certified, vlachas2024parametric, champaney2022engineering}.
Despite recent advances~\cite{barnett2023neural,fresca2022pod,vlachas2025reduced, de2026nonlinear}, both often face limitations when applied in chaotic or strongly nonlinear systems, particularly those exhibiting intricate interactions or multiple parametric dependencies.
In addition, resolving and propagating the dynamics in many real-world applications requires resolving a wide range of scales, rendering the application of equation-based models computationally expensive or even intractable.

In this context, hybrid approaches that integrate data with governing equations have gained traction~\cite{wan2018data}.
Among these, Physics-Informed Neural Networks (PINNs)\cite{raissi2019physics,cuomo2022scientific,karniadakis2021physics} and their meta-learning extensions, such as Meta-PDE\cite{qin2022meta}, HyperPINN~\cite{de2021hyperpinn}, and Meta-Auto-Decoder frameworks~\cite{huang2022meta}, represent prominent efforts to encode the PDE structure directly into the learning process, enhancing interpretability.
Diffusion-based generative models have also been proposed as a means of incorporating physics-driven constraints into probabilistic modeling~\cite{shysheya2024conditional}.
More recently, the Optimizing a Discrete Loss (ODIL) framework has been introduced, which minimizes a cost function for discrete PDE approximations and outperforms PINNs in computational speed and accuracy~\cite{karnakov2024solving, amoudruz2025bayesian}.
Such hybrid approaches often rely on explicit knowledge of the underlying dynamics or the availability of the Jacobian, and may require intrusive access to integration schemes during training.
As a result, their applicability can be limited in high-dimensional systems, or in settings involving complex boundary conditions and broad parametric variability~\cite{de2023physics, huang2023limitations, vlachas2025utility}.
These issues motivate the search for more flexible, non-intrusive alternatives capable of generalizing across diverse system configurations.

In that sense, data-driven approaches remain an established pathway, particularly in the context of autoregressive time-series modeling.
Classical architectures such as Reservoir Computers~\cite{pathak2017using, pathak2018model, zhai2024reconstructing}, Recurrent Neural Networks (RNNs) including Long Short-Term Memory (LSTM) networks~\cite{hochreiter1997long} and Gated Recurrent Units (GRUs)~\cite{vlachas2020backpropagation, vlachas2018data, vlachas2022multiscale, kivcic2023adaptive}, and Transformers~\cite{geneva2022transformers, duthe2025mechanistic} are commonly employed while operator-learning approaches such as DeepONets~\cite{lu2019deeponet} and Neural Operators~\cite{kovachki2023neural, li2020fourier, fanaskov2023spectral, raonic2023convolutional, raonic2023convolutionalICLR} have recently gained ground.

Despite their promise, most data-driven methods train a single forecasting model per parametrization and struggle to generalize or extrapolate reliably across unseen system parameters and dynamical regimes.
Recent works in deep learning for parametric PDEs~\cite{wagner2023stacked, franco2023deep, karbasia2025parametric}, PINNs~\cite{cho2024parameterized}, Neural ODEs~\cite{farenga2025latent}, and Echo State Networks (ESNs)~\cite{luo2024reconstructing, langer2004modeling, roy2022model} also identify this gap and attempt to address it by stacking vector embeddings that combine state and parameter information~\cite{karbasia2025parametric} or by directly augmenting the state with the parameter, following earlier practices~\cite{langer2004modeling}.
Generative approaches have also been explored for surrogate modeling of parameterized nonlinear dynamical systems~\cite{li2023generative, rostamijavanani2023dyncgan}, and meta-learning strategies have been proposed for adaptation to new bifurcation parameter values~\cite{li2025bifurcation}.
However, these methods still rely on shared model weights across all parameter configurations, which limits expressiveness and generalization when dynamics vary significantly across parameter space.
We argue that this pitfall might evolve to a limiting factor hindering generalization, especially if the dynamics of the problem exhibit a wide variability, e.g., from simple oscillatory to fully chaotic behavior.

\begin{figure}
\centering
\includegraphics[width=0.9\columnwidth]{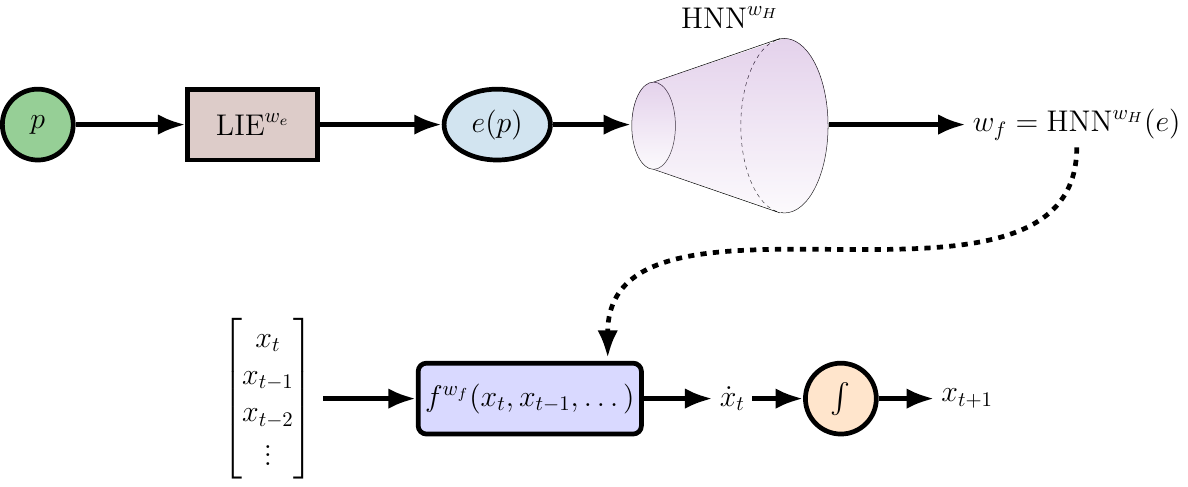}
\caption{
PHLieNet framework: The parameter \( \param \) is passed through the Learned Interpolated Embedding (LIE) layer to produce an embedding $e( \param )$.
This embedding is used by the Hypernetwork to generate the weights of a target network (e.g., a causal dilated CNN or LSTM), which is then used to model and integrate the system’s temporal dynamics.
}
\label{fig:phlienet}
\end{figure}

To overcome these limitations, we introduce \textit{PHLieNet}, a \textit{Parametric Hypernetwork for Learning Interpolated Networks}, a novel framework that explicitly conditions a hypernetwork on a learned embedding of the system parameters. 
A graphical abstract is offered in~\autoref{fig:phlienet}.
PHLieNet maps each parameter vector to a continuous latent embedding by interpolating over learned embedding vectors associated with fixed anchor points.
This embedding is then passed to a hypernetwork, which generates the weights of a forecaster network.
The forecaster models the temporal evolution of the system’s state, conditioned on short-term history.
By dynamically adapting the weights of the target network to reflect the input system parameters, PHLieNet provides a unified and flexible modeling framework capable of representing a wide range of dynamical regimes.
Compared to state augmentation methods that interpolate implicitly in state space, PHLieNet interpolates in the space of models so that each interpolated parameter yields a dedicated forecaster, i.e., an approximator of the corresponding dynamics propagation operator.
Crucially, the proposed approach allows differentiation through the hypernetwork with respect to the input parameters, enabling the computation of parameter-aware gradients.
This capability allows PHLieNet to support gradient-based training and inference over both state and parameter spaces.
This stands in contrast to most state-of-the-art methods, which either require training separate models for each parametrization or lack mechanisms for explicit parametric generalization.
The key contributions are:
\begin{itemize}
    \item A novel framework, PHLieNet, that conditions a hypernetwork on a learned embedding of the system parameters to generate the weights of a target forecasting network, enabling a single unified model to capture a diversity of dynamical regimes without retraining for each parametrization.
    \item A learned interpolated embedding based on RBF anchor interpolation that structures the parameter space without explicit supervision, enabling both interpolation and extrapolation to unseen parameter values.
    \item A formal existence result showing that, under smoothness of the parametric dynamics, there exists a hypernetwork that approximates the optimal weight mapping to arbitrary precision.
    \item Systematic benchmarking on five dynamical systems, demonstrating consistent improvements over state augmentation and parameter-agnostic baselines in short-term forecasting accuracy and long-term attractor statistics.
\end{itemize}

PHLieNet relies on hypernetworks to output the parameters of the dynamics' forecaster, known as the target network~\cite{ha2016hypernetworks}.
Originally introduced in the context of meta-learning \cite{finn2017model, ravi2017optimization}, hypernetworks have been used to generate initial weights or learning rules that enable rapid adaptation in low-data regimes or few-shot learning scenarios.
Their success in this domain has spurred broader adoption across applications such as neural architecture search~\cite{zoph2016neural, pham2018efficient, liu2018darts} and across a range of parametric tasks, including image retouching \cite{chai2020supervised}, style transfer \cite{alaluf2022hyperstyle}, and differentiable pruning \cite{li2020dhp}.
Despite this growing interest, the application of hypernetworks to the modeling of dynamical systems, particularly those governed by ODEs and PDEs, remains relatively nascent~\cite{chauhan2024brief}.
Early works demonstrate promising directions, including dynamic convolutions, implemented as hypernetworks, which have achieved promising results in short-range weather forecasting~\cite{klein2015dynamic}.
Furthermore, Berman et al. \cite{berman2024colora} proposed CoLoRA, which adapts low-rank weights of neural networks for new parameters and initial conditions, and Zheng et al. \cite{zheng2024hypercan} introduced HyperCAN for modeling mechanical meta-materials under varying conditions.
More recent innovations include Hypersolvers\cite{poli2020hypersolvers}, which leverage hypernetworks to approximate higher-order terms in PDE solvers, and HyperPINNs\cite{majumdar2023hyperlora, cho2023hypernetwork}, which combine hypernetworks with physics-informed learning to improve generalization across varying PDE conditions.

Closest to our problem setting are frameworks such as Context-Informed Dynamics Adaptation (CoDA)~\cite{kirchmeyer2022generalizing}, which uses a hypernetwork to condition a dynamics model on inferred environment-specific context vectors, and LEADS, by Yin et al.~\cite{yin2021leads}, which explicitly decomposes dynamics into shared and environment-specific components to generalize across environments.
Such context-based methods that infer latent representations from observed trajectories aim for parameter-agnostic generalization but are limited when precise parametric information is available and can be explicitly leveraged.
In addition, they are tailored for settings where parametric information is inferred from data.
In contrast, PHLieNet operates under the assumption that parameter values are explicitly available, leveraging them directly to enable more precise and flexible modeling across diverse dynamical regimes.
By directly incorporating parametric variations into the embedding, PHLieNet ensures continuous and accurate forecasting across parameter space without compromising adaptability.
This design supports a principled and scalable means of interpolating between distinct dynamical regimes within a single unified framework.

A related direction is the work of Guo et al.~\cite{guo2025parametric}, who propose learning parametric Koopman decompositions by jointly training a shared observable dictionary and a parametric family of projected Koopman operators.
While this framework provides a principled linearization of parametric dynamics and is well-suited for control applications, it requires the existence of a finite-dimensional Koopman-invariant subspace, which is generally unavailable for systems with continuous spectra such as chaotic attractors.
PHLieNet avoids this limitation by operating directly in state space, and is evaluated on systems spanning the full range of dynamical regimes, including strongly chaotic ones.

Therefore, the proposed PHLieNet framework, supported by recent theoretical work that demonstrates the advantages of hypernetworks in terms of modularity and expressiveness~\cite{galanti2020modularity}, provides a scalable alternative for data-driven inference on complex dynamical systems with parametric variability. 
By avoiding precomputed reduced spaces, explicit physical constraints, and fixed model architectures, PHLieNet adapts flexibly to a broad range of system behaviors, from fixed points and periodic orbits to chaos.
We benchmark its performance on a diverse set of dynamical systems and against parameter-agnostic temporal dynamics models and parameter-augmented models.  
Across all systems, PHLieNet consistently outperforms or matches state-of-the-art models.
Moreover, we demonstrate that PHLieNet is expressive and capable of learning the complete spectrum of dynamics across parametric regimes, enabling time extrapolation and generalization to unseen parameters.
These results highlight the robustness and flexibility of PHLieNet, advancing the state of the art in data-driven, parametric modeling of dynamical systems.

This paper is organized as follows: 
In \Cref{sec:methods}, the problem formulation is presented, along with the considered baseline models. \Cref{sec:phlinet} presents the proposed PHLieNet framework and~\Cref{sec:metrics} the comparison metrics used. 
In \Cref{sec:applications}, we describe the benchmark dynamical systems, the models used for comparison, and the numerical results that highlight the efficiency and effectiveness of PHLieNet.
\Cref{sec:discussion} concludes the paper by summarizing our contributions, offering insights and suggesting directions for future research.
\section{Methods}
\label{sec:methods}

\subsection{Parametric Dynamical Systems}

A parametric dynamical system can be represented by a system of equations whose evolution depends on a parameter vector \( \param \in \RR^{\dimParam} \), whose components may include physical constants, external conditions, or control inputs. 
Such systems can be formulated in continuous time as either ordinary differential equations (ODEs) or partial differential equations (PDEs).
For ODEs, the state \( \state(t) \in \RR^{\dimState} \) evolves according to:
\begin{equation}
\frac{d\state}{dt} = f(\state(t) , \param),
\label{eq:ode}
\end{equation}
where \( f: \RR^{\dimState} \times \RR^{\dimParam} \to \RR^{\dimState}  \) defines the dynamics of the system as a function of the state \( \state \) and parameters \( \param \).

In the discrete-time setting, the evolution of the system is modeled as a sequence of state transitions governed by a nonlinear update function:
\begin{equation}
\state_{t+1} = \Phi(\state_t, \param),
\label{eq:discrete}
\end{equation}
where \( \Phi: \RR^{\dimState} \times \RR^{\dimParam} \to \RR^{\dimState} \) represents a possibly learned or explicitly defined discrete-time transition map.  
This formulation encompasses a wide range of integration schemes, including explicit or implicit methods (e.g., Euler or Runge-Kutta~\cite{liu2022hierarchical}), as well as data-driven alternatives such as Neural ODEs~\cite{chen2018neural}.  
Our method is compatible with any such formulation. For simplicity and clarity, we adopt a first-order integration scheme in this work, expressed as:
\begin{equation}
\state_{t+1} = \state_t + \Delta t \cdot f(\state_t, \param),
\end{equation}
where \( \state_t \) is the state at time \( t \), \( \param \) is the parameter vector, \( f(\state_t, \param) \) is the time derivative (i.e., the gradient of the state), and \( \Delta t \) is the discretization step.
This formulation corresponds to an explicit Euler integration of the continuous-time dynamics.
Parameter vector \( \param \) can vary, resulting in a diverse set of phenomena such as fixed points, periodic orbits, and chaotic dynamics.
This variation enables the study of how changes in \( \param \) influence the evolution of the system.
A key challenge in parametric dynamical systems is efficiently capturing the relationship between \( \param \) and the resulting dynamics, especially across a wide range of parameter values or when \( \param \) itself varies over time.

\subsection{Learning Temporal Dynamics}

In forecasting dynamical systems, we want to learn an approximation of $f$, using a parametrized model $f^{w_f}$ by minimizing some reconstruction or prediction error.
The variable $w_f$ represents the parameters of the approximator $f$.
In case of a neural network, for example, $w_f$ is the set of all weights and biases of the network.
To address non-Markovian effects, account for missing information, and improve performance in long-term forecasting, the approximator often incorporates information from the previous history of the state.
Models that are explicitly designed to process sequential data, such as RNNs\cite{sutskever2013training, graves2013generating} and Temporal Convolutional Neural Networks with Causal Dilated convolutions (TCNN-CD)\cite{bai2018empirical}, are natural choices for this task because they can effectively learn temporal patterns and dependencies in time series. 
In what follows, we use the term Temporal Dynamic Networks (TDNs) to refer to such models, grouping together architectures specifically designed for learning from sequential data.
Such approaches align with Takens' theorem~\cite{takens1981dynamical}, which demonstrates that a system's dynamics can be reconstructed from a time-delayed embedding of its state under certain conditions.

In the case of TDNs, the state evolution is approximated by:
\begin{equation}
\tilde{\state}_{t+1} = 
\int \tilde{\statedt}_{t} dt,
\quad
 \tilde{\statedt}_{t}=
f^{w_f}(
\underbrace{
\state_t, \dots, \state_{t-\isl+1}
}_{\text{history}} \, ; \, \param).
\label{eq:tdn}
\end{equation}
where $\tilde{\bullet}$ denotes inferred quantities, the networks are used to approximate the dynamics (time derivative of the state), and we truncate the dependence on the previous states after $\isl$ timesteps (state-less formulation).

To account for the influence of parametric variability, expressed by $\param$, the network needs to be fitted to trajectories from the parametrized dynamics.
Let us assume we have response data from a set of parameters $P_{train}=\{p_1, \dots, p_{\np}\}$ with total $|P_{train}|=N_{P}$ parametrizations. 
For each parameterization, we assume $\nics$ trajectories, each one representing different initial conditions (after eliminating initial transients) and consisting of $\ntimesteps$ timesteps.
Thus, the train data are $\trainData \in \RR^{\np\times \nics \times \ntimesteps}$.
In turn, the neural network employed as an approximator $f$ is trained to minimize the prediction loss across time.
Specifically, for a given batch of a trajectory of the data $x_{t-\isl+1}^{j,k}, \dots, x_t^{j,k}, x_{t+1}^{j,k}$, corresponding to parameter $p_j \in P_{train}$, and initial condition $k \in \{1, \dots, \nics \}$, the loss is defined as:
\begin{equation}
    \loss_{j, k, t} = 
    \|
    \statedt_{t} - 
    \tilde{\statedt}_{t}
    \|^2 =
    \|
    \statedt_{t} - 
    f^{w_f}(
    \state_t, \dots, \state_{t-\isl+1}
    ; \param)
    \|^2.
\end{equation}
The network parameters are optimized by minimizing the loss over the entire parameter set, across all initial conditions and timesteps, as follows:
\begin{equation}
w_f = \arg \min_{\substack{p_j \in P_{train} \\ k \in \{1, \dots, \nics\} \\ t \in \{1, \dots, \ntimesteps\}}} \loss_{j, k, t}
.
\label{eq:min}
\end{equation}
Given a sufficiently diverse parameter set $P_{train}$ to capture the system’s behavior, the trained network can be used to forecast unseen dynamics, extrapolate in time, and even generalize to unseen parameters.
The latter is a significantly more challenging task, as varying parameters can profoundly alter the attractor structure and overall dynamics.

\subsubsection{Recurrent Neural Networks}

A natural choice to approximate the time derivative \( f^{w_f}(\state_t, \param) \) is an LSTM network. 
LSTMs are particularly effective at capturing long-range dependencies through a gated mechanism that controls the flow of information over time steps~\cite{hochreiter1997long}.
Such models have been successfully applied in learning dynamical system representations~\cite{vlachas2020backpropagation}.
The LSTM updates its hidden state \( \mathbf{h}_t \) and the cell state \( \mathbf{c}_t \) at each time step \( t \) based on the previous states \( \mathbf{h}_{t-1} \), \( \mathbf{c}_{t-1} \), and the current input \( \state_t \).
The update equations are given by:
\begin{align}
    \mathbf{i}_t &= \sigma(\mathbf{W}_i \state_t + \mathbf{U}_i \mathbf{h}_{t-1} + \mathbf{b}_i), \\
    \mathbf{f}_t &= \sigma(\mathbf{W}_f \state_t + \mathbf{U}_f \mathbf{h}_{t-1} + \mathbf{b}_f), \\
    \mathbf{o}_t &= \sigma(\mathbf{W}_o \state_t + \mathbf{U}_o \mathbf{h}_{t-1} + \mathbf{b}_o), \\
    \tilde{\mathbf{c}}_t &= \tanh(\mathbf{W}_c \state_t + \mathbf{U}_c \mathbf{h}_{t-1} + \mathbf{b}_c), \\
    \mathbf{c}_t &= \mathbf{f}_t \odot \mathbf{c}_{t-1} + \mathbf{i}_t \odot \tilde{\mathbf{c}}_t, \\
    \mathbf{h}_t &= \mathbf{o}_t \odot \tanh(\mathbf{c}_t),
\end{align}
where \( \sigma \) denotes the sigmoid activation function and \( \odot \) represents element-wise multiplication.
The vectors \( \mathbf{i}_t \), \( \mathbf{f}_t \), and \( \mathbf{o}_t \) are the input, forget, and output gates, respectively.
All parameters \( \mathbf{W}_\ast \), \( \mathbf{U}_\ast \), and \( \mathbf{b}_\ast \) are learned during training and collectively constitute the parameter set \( w_f\), which is not to be confused with the parametric dependencies of the model $\param$.
The LSTM approximates the time derivative of the state $f^{w_f}$ as in~\Cref{eq:tdn}.
During training, the model minimizes the loss defined in~\Cref{eq:min} across all parameterizations and initial conditions, allowing \( f^{w_f} \) to learn the parameter-dependent dynamics of the system.
\begin{figure}
\centering
\includegraphics[width=0.4\columnwidth]{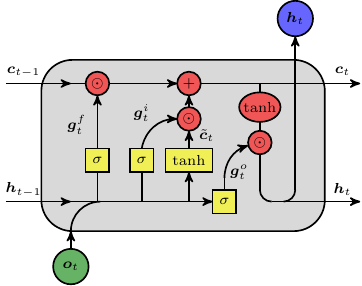}
\caption{
Information flow of a Long Short-Term Memory (LSTM) Cell.
}
\label{fig:lstm}
\end{figure}

\subsubsection{Causal Dilated Temporal CNN (TCNN-CD)}

Another effective solution to model sequential data, while ensuring causality, is a TCNN-CD.
This architecture leverages dilated convolutions to efficiently capture long-range dependencies without relying on recurrent structures.
Specifically, each convolutional layer uses a dilation factor that grows exponentially with the layer index:
\[
d_i = 2^i, \quad i=0,1,\dots,L-1,
\]
where \( L \) is the number of layers.
This exponentially increasing dilation pattern ensures that the receptive field grows rapidly, enabling the model to capture temporal dependencies over large windows.
For a 1D temporal convolution at time step \( t \), the output \( y_t \) is computed as:
\[
y_t = \sum_{i=0}^{k-1} w_i \cdot x_{t - i \cdot d},
\]
where \( w_i \) are the learned convolutional weights, \( k \) is the kernel size, and \( d \) is the dilation factor.
This formulation allows each output \( y_t \) to aggregate information from a causal receptive field of past states \((x_t, x_{t-1}, \dots, x_{t-\isl})\), without accessing future inputs.
Causal padding is used in each convolutional layer to prevent information leakage from future time steps. For a convolutional kernel of size \( k \), the causal padding at layer \( i \) is computed as:
\[
\text{padding}_i = d_i \cdot (k - 1).
\]
The total receptive field \( R \) of the network is then:
\[
R = 1 + \sum_{i=0}^{L-1} (k-1) \cdot d_i.
\]
After each convolution, a smooth nonlinearity, the Sigmoid Linear Unit (SiLU), is applied:
\[
\text{SiLU}(x) = x \cdot \sigma(x),
\]
where \(\sigma(x)\) is the sigmoid function.
Unlike recurrent networks, TCNN-CD omits fully connected layers, relying instead on a final \( 1 \times 1 \) convolution to project the learned features to the output dimension.
This approach reduces the parameter count while maintaining sufficient expressiveness for temporal data modeling.

\begin{figure*}[!htb]
    \begin{subfigure}[b]{0.485\textwidth}   
    \centering 
    \includegraphics[width=0.99\columnwidth, height=0.50\columnwidth]{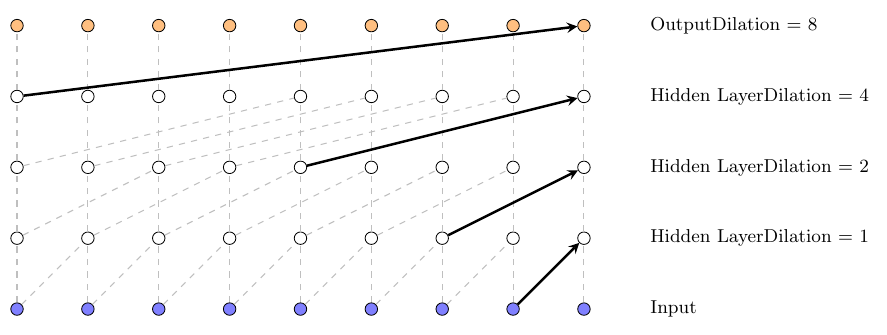}
    \caption[]%
    {Visualization of the exponentially increasing dilation pattern.
    Each hidden layer employs a convolutional kernel with a dilation factor \( d_i = 2^i \), which enables the network to efficiently capture long-range dependencies across input sequences.
    The final output aggregates information from a broad receptive field that spans multiple temporal scales.} 
    \label{fig:dilation}
    \end{subfigure}
    \quad
    \begin{subfigure}[b]{0.485\textwidth}   
    \centering 
    \includegraphics[width=0.99\columnwidth, height=0.50\columnwidth]{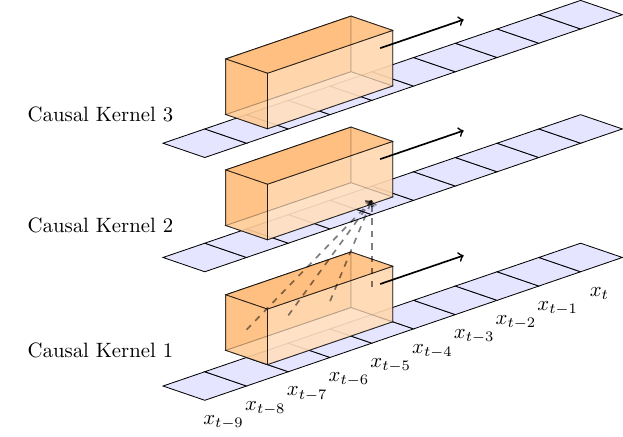}
    \caption[]%
    {Illustration of the causal convolutional kernels.
    Each kernel processes the current and past input states, thereby preserving the temporal order and preventing information leakage from future time steps.
    Different layers have different dilation factors, allowing the receptive field to expand and integrate long-range dependencies while respecting causality.}
    \label{fig:causalkernels}
    \end{subfigure}
    \caption{Illustration of the exponentially increasing dilation pattern and the causal convolutional kernels in the TCNN-CD.}
    \label{fig:TDCN}
\end{figure*}

In this work, the number of layers \( L \) is automatically determined by the length of the input sequence and the size of the kernel to ensure that the receptive field covers the necessary temporal context.
To determine the minimum number of layers \( L \) required to cover an input sequence of length \( \isl \), we analytically invert the receptive field formula.
Given a kernel size \( k > 1 \), the receptive field grows as \( R = 1 + (k - 1)(2^L - 1) \).
Solving for \( L \), we obtain:
\[
L = \left\lceil \log_2\left( \frac{\isl - 1}{k - 1} + 1 \right) \right\rceil.
\]
This ensures that the receptive field spans at least \( \isl \) time steps, allowing the network to access the full temporal context during training.
For example, with \( k = 5 \), this results in \( L = 3 \) layers for sequences of length \( \isl = 16 \), and \( L = 4 \) layers for \( \isl = 32 \).
More information on the hyperparameters of the models used is reported in~\ref{sec:appendix:parameters}.
The dilation pattern is illustrated in~\Cref{fig:dilation}.
The causal temporal kernel is illustrated in~\Cref{fig:causalkernels}.

\subsection{Modeling Parametric Dependencies}

The main challenge addressed in this work lies in formulating an expressive functional representation for~\Cref{eq:tdn}. 
We distinguish between three modeling paradigms: (i) a trivial, parameter-agnostic formulation; (ii) the established approach of state augmentation for parametric modeling, as reviewed in~\autoref{sec:intro}; and (iii) the proposed PHLieNet framework, which introduces a principled alternative.

\subsubsection{Parametric-Agnostic Case}
\label{sec:atdn}

A straightforward way to handle parametric dependency is to treat all parametric dynamics uniformly, assuming that $\param$ does not significantly change the functional form, effectively ignoring the parametric dependency.
Alternatively, we may approximate the complete $f^{w_f}$ without assuming explicit knowledge or dependence on $\param$.
This form is referred to as the parametric-agnostic model.
The functional form then becomes:
\begin{equation}
\tilde{\state}_{t+1} = \state_t + \Delta t \cdot f^{w_f}(\state_t, \dots, \state_{t-\isl+1}).
\label{eq:atdn}
\end{equation}
Any temporal dynamics model such as an LSTM, a GRU, or a TCNN-CD can then be used to model~\Cref{eq:atdn}.

\subsubsection{State Augmentation}
\label{sec:ptdn}

Another straightforward way to handle parametric dependency is to augment the hidden state.
As a result, $f^{w_f}$ becomes a neural network that receives as input a vector with the state concatenated to the parameters.
Thus, the augmented state is given by:
\begin{equation}
\augState_{t} = \begin{bmatrix}
\state_t \\ \param
\end{bmatrix} \in \RR^{\dimState+\dimParam}
\label{eq:xaug}
\end{equation}
and the state evolution is approximated by
\begin{equation}
\tilde{\state}_{t+1} = \state_t + \Delta t \cdot f^{w_f}(\augState_t, \dots, \augState_{t-\isl+1}).
\label{eq:ptdn}
\end{equation}

\section{PHLieNet: Parametric Hypernetwork for Learning Interpolated Networks}
\label{sec:phlinet}

In systems where the dynamics heavily depend on the external parameters $\param$, learning a single set of network coefficients $w_f$, like proposed in~\Cref{sec:atdn}, may not adequately capture the full range of behaviors induced by different values of $\param$.
Moreover, appending the parameters to the state of the system, as in~\Cref{sec:ptdn}, may not be adequate, as the parameters might affect the structural form of $f$.
Instead, concatenation hinders flexibility in the expressiveness of $f$.
To address this challenge, in this work, we utilize a hypernetwork which dynamically generates the coefficients $w_f$ of the network(s) used to model $f$ conditioned on the input parameter vector $\param$.

Hypernetworks, as introduced by Ha et al.~\cite{ha2016hypernetworks}, provide a framework for generating the coefficients $w_f$ of another neural network, meaning the corresponding weights and biases.
Instead of directly learning a function \( f \) that models the system's dynamics for each possible value of \( \param \), a hypernetwork can be used to generate the coefficients of \( f \) conditioned on \( \param \).
However, directly conditioning the hypernetwork on the raw parameters \( \param \) presents significant challenges, as the network might struggle to distinguish between qualitatively different dynamical regimes, especially when transitions are nonlinear or discontinuous.
This leads to poor generalization across regimes and necessitates some form of representation learning or clustering to structure the parameter space.
A related approach using linear RNNs and a linear embedding of the parameter vector was proposed in~\cite{brenner2024learning}.
Extending such approaches to nonlinear systems and nonlinear target networks remains an open challenge.

In this work, we adopt a different approach.  
We capture the parametric dependence of the system through a structured two-stage mechanism, although the entire architecture is trained end-to-end.  
First, the input parameters \( \param \) are mapped to a continuous embedding via continuous interpolation over a set of learned anchor embeddings using radial basis functions (RBFs).
Second, the continuous embedding is passed to a hypernetwork that generates the coefficients of the target network.  
We refer to this method as \textit{Parametric Hypernetwork for Learning Interpolated Networks (PHLieNet)}.  
A detailed description of its implementation follows. 

\subsection{Step 1: Learned Interpolated Embedding}

Let \( \{ \param^{(i)} \}_{i=1}^{\nE} \) be a set of anchor parameter vectors and let \( \{ \embedding^{(i)} \}_{i=1}^{\nE} \subset \mathbb{R}^{\dimE} \) be their corresponding learned embeddings.
The embeddings are learned, so the matrix \( w_e = [\mathbf{e}^{(1)}, \dots, \mathbf{e}^{(\nE)}] \) represents the weights of this layer.
Given a new parameter vector realization \( \param^j \), we compute the interpolation weights \( \{ \alpha_i(\param^j) \}_{i=1}^{\nE} \) such that:
\begin{equation}
\sum_{i=1}^{\nE} \alpha_i(\param^j) = 1, \quad \alpha_i \geq 0,
\end{equation}
and define the embedding of \( \param^j \) as:
\begin{equation}
\embedding(\param^j) = \sum_{i=1}^{\nE} \alpha_i(\param^j) \, \embedding^{(i)}.
\label{eq:embedding}
\end{equation}

\begin{figure}
\centering
\includegraphics[width=0.4\columnwidth]{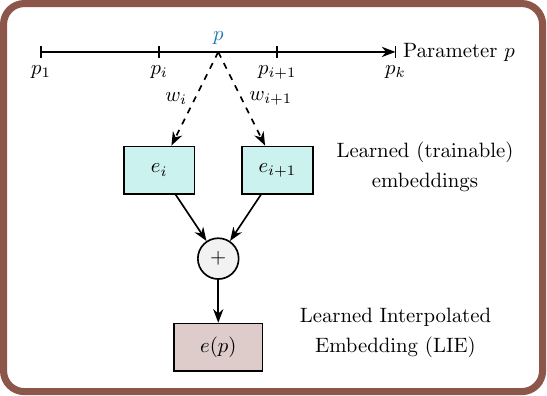}
\caption{
Overview of the learned interpolation mechanism used in PHLieNet.
An input parameter vector \( \param \) is used to compute interpolation weights \( \{ a_i(\param) \} \) over a set of anchor points \( \{ \param^{(i)} \} \).
Each anchor is associated with a learned embedding \( \embedding^{(i)} \).
The final embedding \( \embedding(\param) = \sum_i a_i(\param) \embedding^{(i)} \) is a convex combination of the learned embeddings, which is then used as input to the hypernetwork to generate the coefficients of the target network.
This structure enables generalization across parameter space by smoothly interpolating between known regimes.
}
\label{fig:lie}
\end{figure}

Specifically, the interpolation weights are computed using an RBF kernel with a softmax normalization.
Given a parameter realization \( \param^j \), we compute the squared distances to each anchor position and apply a Gaussian kernel:
\begin{equation}
\alpha_i(\param^j) = \frac{\exp\!\Big(-\frac{\| \param^j - \param^{(i)} \|^2}{2\sigma^2}\Big)}{\sum_{k=1}^{\nE} \exp\!\Big(-\frac{\| \param^j - \param^{(k)} \|^2}{2\sigma^2}\Big)},
\label{eq:rbf_weights}
\end{equation}
where \( \sigma > 0 \) is the bandwidth parameter controlling the locality of the interpolation.
The softmax normalization ensures that the weights satisfy the convexity conditions \( \alpha_i \geq 0 \) and \( \sum_i \alpha_i = 1 \).
Importantly, all \( \nE \) anchors contribute to every embedding, with contributions decaying smoothly with distance in parameter space.
This yields a continuous and differentiable embedding \( \embedding(\param^j) \) as defined in~\Cref{eq:embedding}, which allows the method to interpolate and, to some extent, extrapolate across diverse parameterized dynamical regimes.
In higher-dimensional parameter spaces, the same RBF mechanism generalizes naturally, as the Gaussian kernel operates on the norm of the parameter vector without modification.
The learned interpolated embedding layer is illustrated in~\Cref{fig:lie}.

The number of anchor embeddings \( \nE \) determines the trade-off between the expressiveness and generalization ability of the PHLieNet framework.
On the one hand, \( \nE \) must be sufficiently smaller than the number of parameter values included in the training data, ensuring that the network is forced to learn meaningful interpolations in embedding space rather than memorizing the dynamics associated with each training parameter. 
This promotes generalization and encourages the model to capture shared structure across parametric regimes. 
On the other hand, if \( \nE \) is too small, the resulting embedding space may lack the capacity to represent the diversity of dynamics present in the dataset, particularly in systems exhibiting rich or highly nonlinear behaviors. 
In such cases, the model may fail to generate target networks with sufficient representational power, limiting its ability to accurately forecast dynamics across parameter space. 
Therefore, \( \nE \) must be chosen to balance the trade-off between interpolation and expressiveness.

\subsection{Step 2: Parameter Generation via Hypernetwork.}

The second stage of our framework involves the generation of the coefficients $w_f$ of the target temporal dynamics model using a hypernetwork.  
The hypernetwork, denoted by \( \hnn \), takes as input the embedding \( \mathbf{e}(\param^j) \in \mathbb{R}^{\dimE} \) produced in Step 1 and outputs the parameters \( w_f \in \mathbb{R}^{|w_f|} \) of network \( f \).  
Formally, this mapping is defined as:
\begin{equation}
w_f = \hnn \left(\mathbf{e}
(
\param^j
); w_{H} \right),
\label{eq:hnn1}
\end{equation}
where \( \hnn: \mathbb{R}^{\dimE} \to \mathbb{R}^{|w_f|} \) is the hypernetwork parameterized by coefficients \( w_H \), and \( {|w_f|} \) denotes the number of weights of the target model \( f \).
The target temporal dynamics network \( f^{w_f} \), with coefficients $w_f$ generated by the hypernetwork, is then used to model the time derivative of the system's state based on a history of observations, as in~\Cref{eq:tdn}:
\begin{equation}
\tilde{\statedt}_t = f^{w_f}
\Big(
\state_t, \dots, \state_{t-\isl+1}
\Big) 
= f^{\hnn \big(
\mathbf{e}(\param^j); \, w_H
\big)
}
\Big(
\state_t, \dots, \state_{t-\isl+1}
\Big) 
.
\label{eq:hnntdn}
\end{equation}
The proposed PHLieNet framework is illustrated in~\Cref{fig:phlienet}.
By conditioning the coefficients $w_f$ of the temporal dynamics model on the system parameters \( \param \), the hypernetwork enables flexible and continuous adaptation to a wide range of dynamical regimes.  
This architecture allows a single, unified model to generalize across different parameter configurations, eliminating the need to train separate models for each regime.
More details on the hypernetwork architecture are provided in~\ref{sec:appendix:hypernetwork}.

\begin{figure}
\centering
\includegraphics[width=0.9\columnwidth]{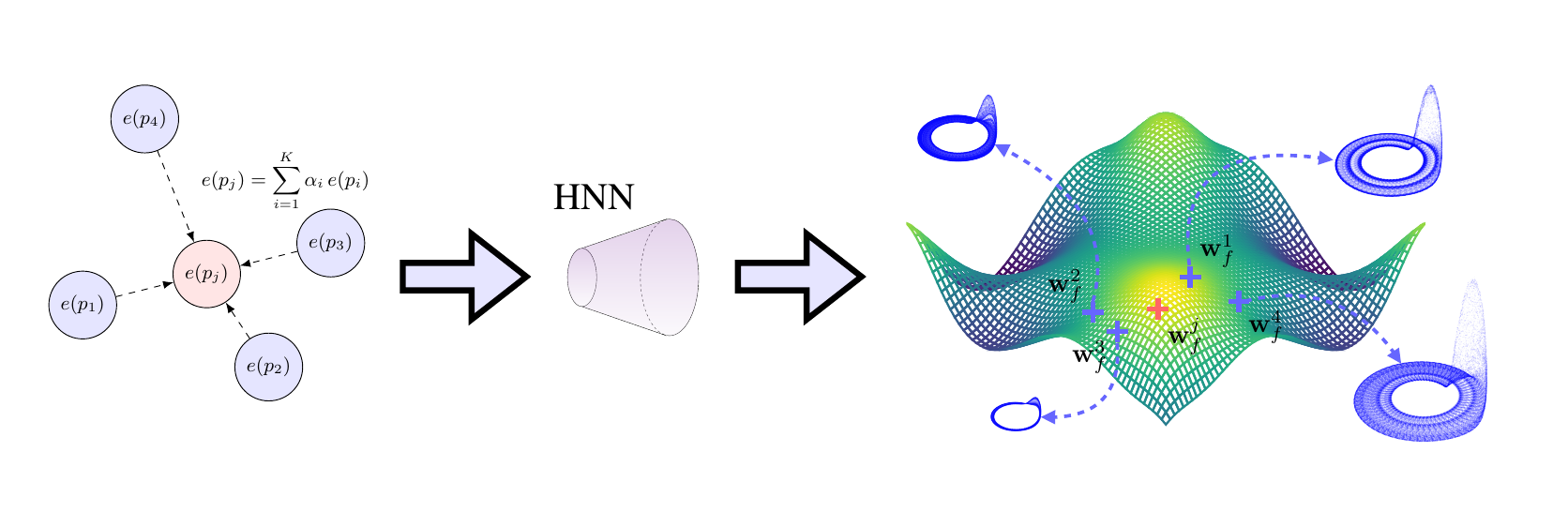}
\caption{
Overview of the three-step modeling process: (1) Linear combination of task embeddings in the shared embedding space; (2) Transformation of the embedding through a hypernetwork to generate corresponding model weights; (3) Nonlinear interpolation in the model (weight) space, which induces interpolation in the resulting phase space dynamics.
This procedure induces smooth transitions across parametric tasks while preserving expressive dynamical behavior.
}
\label{fig:interpolation_pipeline}
\end{figure}

Figure~\ref{fig:interpolation_pipeline} offers a methodological perspective on PHLieNet, emphasizing interpolation in the weight space.
The process begins with the linear combination of task-specific embeddings, representing different dynamical regimes, in a shared latent space.
These interpolated embeddings are then mapped by a hypernetwork to generate model weights, effectively performing interpolation in the weight space. 
Crucially, this nonlinear interpolation induces meaningful transitions in the phase space dynamics of the resulting models.
By focusing on the weight space (model space), rather than the state or parametric space, this approach enables coherent and controllable blending of dynamical behaviors across tasks or parameter regimes.

\subsection{Why Interpolate in Weight Space?}
\label{sec:whyinterpolation}

To understand the motivation for the proposed interpolation in the weight space, we here shortly discuss the limitations of alternative approaches as presented in~\cref{sec:methods}, and analyze their behavior in inference post-training.
The best performing approach is state augmentation.
A state-augmented network as in~\Cref{sec:ptdn} employs a fixed set of weights shared across all parameters.
The parameter enters explicitly as an additional input.
The network has to learn the mapping from the parameter to the state space behavior implicitly through its activations at every forward pass.
All convolutional filters, recurrent transition matrices, and biases are shared.
As a consequence, the network must partition its capacity across qualitatively different regimes.

To illustrate the representational limitation of state augmentation, consider a simple linear parametric system $\statedt = W( \param )  \state$, where the transition matrix $W(\param)$ depends smoothly on the parameter $\param$.
For instance, different values of $\param$ may yield different eigenvalues, corresponding to qualitatively different stability properties or oscillation frequencies.
A state-augmented recurrent network with fixed weights processes the concatenated input $[\state_t, \param]$ at each timestep, producing a hidden state update of the form $h_{t+1} = \sigma(W_1 [\state_t, p]^\top + W_2 \mathbf{h}_t)$, where $W_1$, $W_2$ are learned once and shared across all values of $\param$.
In this scenario, the parameter can shift the operating point of the gates but it cannot influence the transition matrices that govern characteristics of the dynamics (e.g. Lyapunov exponents, eigenvalues, etc.).

In contrast, PHLieNets learn a different set of weights per parameter value, directly influencing the transition function.
This design reflects a fundamental distinction: the system parameter $\param$ and the state are quantities of different nature.
The state evolves in time and is governed by the dynamics, while the parameter defines which dynamics are active.
State augmentation conflates these two modalities by concatenating them into a single input vector.
In PHLieNets this distinction is clear:
\begin{itemize}
\item The target network operates purely on temporal state space dynamics.
\item Each parametric regime gets its own set of convolutional filters and nonlinear response characteristics.
\item The parametric dependence, thus, modulates the entire computational graph. 
\end{itemize}

Moreover, PHLieNets learn to generate weights that define a complete dynamical system valid for all initial conditions and time horizons.
Interpolating in the weight space produces a new dynamical system, a new function, for the intermediate parameter.
Under smoothness of the parametric dynamics, this interpolation is principled as we explain in~\ref{sec:appendix:learnability}.
\section{Evaluation Metrics}
\label{sec:metrics}

To evaluate the forecasting performance of parametric models across different dynamical systems, we employ complementary metrics that capture different aspects of predictive accuracy.
Namely, we employ the time evolution of the normalized root-mean-square error (NRMSE), the Time-to-Threshold (TtT), and the power spectrum error.
These metrics assess short-term prediction accuracy and long-term spectral fidelity, and are briefly presented below.

\subsection{Time Evolution of the Normalized RMSE}
\label{sec:nrmse_evolution}

To evaluate the accuracy of model predictions over time, we computed the NRMSE as a function of time.
The NRMSE at time $t$ is defined as:
\begin{equation}
    \nrmse(t) =
    \frac{
        \| \statePred(t) - \state(t) \|_2
    }{
        \sqrt{ \sigma^2 } + \varepsilon
    },
\end{equation}
where $\statePred(t)$ and $\state(t) \in \RR^{\dimState}$ are the predicted and true states, and $\sigma^2$ is the variance of the ground truth states for the given parameter value, aggregated across initial conditions, times, and dimensions.
The small constant $\varepsilon$ ensures numerical stability.
We calculate the mean of the NRMSE across initial conditions to characterize the predictive performance of different models.
This time-resolved error curve provides information on how the accuracy degrades during extrapolation in time.

\subsection{Time-to-Threshold (TtT)}
\label{sec:ttt}

The Time-to-Threshold (TtT) metric quantifies the duration for which the predicted trajectory remains within an acceptable error margin relative to the ground truth.
We define the TtT based on the NRMSE defined in~\Cref{sec:nrmse_evolution}.
It measures the maximum continuous time during which the normalized error stays below a specified threshold $\theta_{\text{rel}}$.
The Time-to-Threshold $\ttt$ is given by:
\begin{equation}
    \boxed{
    \ttt_{ 
	    \theta_{\text{rel}}
    }
    = \max \left\{
    t \;\middle|\;
    \nrmse(t') < \theta_{\text{rel}}
    \quad \forall\, t' \leq t
    \right\} \cdot \Delta t
    },
\end{equation}
where $\theta_{\text{rel}}$ is the predefined relative error threshold and $\Delta t$ is the simulation time step.
In practice, the TtT is calculated as the maximum continuous time before the relative error first exceeds the threshold $\theta_{\text{rel}}$, averaged (or otherwise aggregated) over multiple initial conditions to obtain a robust measure of predictive performance.
We report TtT in two complementary ways.
In the NRMSE evolution figures, $\ttt_{0.2}$ is annotated in the legend: the NRMSE curves of all initial conditions and all test parameters are first averaged into a single mean curve, and the threshold crossing of that curve is reported.
In the dedicated TtT bar plots, the NRMSE curves are averaged over the initial conditions of each test parameter separately, and $\ttt_{0.2}$ is the threshold crossing of that per-parameter mean curve.
The bar plot then reports the mean $\pm$ standard deviation of these per-parameter values over the test parameter set, exposing the spread of prediction horizons across dynamical regimes.

\subsection{Power Spectrum Error}

The power spectrum error measures the discrepancy between the frequency content of predicted and true trajectories.  
For each dimension, we compute the power spectral density (PSD) using a real-valued Fast Fourier Transform (FFT).  
Given a signal \( x(t) \in \RR \), which is a component of the state evolution $\state \in \RR^{\dimState}$ sampled at frequency \( f_s = 1 / \Delta t \), the frequency spectrum in decibels (dB) is calculated as:
\begin{equation}
\text{PSD}(f) = 20 \log_{10} \left( \frac{2}{N} |\mathcal{F}[x](f)| \right),
\end{equation}
where \( \mathcal{F}[x](f) \) denotes the FFT of the signal \( x(t) \), and \( N \) is the number of time steps.  
The power spectrum error is then defined as the average \( \ell_1 \)-distance between the predicted and true spectra across dimensions:
\begin{equation}
\text{Spectrum Error} = \frac{1}{\dimState} \sum_{d=1}^{\dimState} \frac{1}{F} \sum_{f=1}^F \left| \text{PSD}_{\text{pred}}^{(d)}(f) - \text{PSD}_{\text{true}}^{(d)}(f) \right|,
\end{equation}
where \( F \) is the number of frequency bins. In practice, this error is further averaged over initial conditions.

\section{Applications}
\label{sec:applications}

Our framework is implemented in PyTorch~\cite{paszke2019pytorch}.
We build the PHLieNet implementation on the hypernetwork library~\cite{sudhakaran2022}, extending it to support parametric weight generation.
Our runs are conducted on the Euler supercomputing cluster at ETH Zurich.
Each run uses an RTX 3090 GPU, with 8 CPUs per task and 4 GB of RAM per CPU.
For benchmarking, we consider the networks summarized in \autoref{tab:networks}.

Although the architecture of the TCNN-CD is quite effective, as it exploits temporal invariances in the data, it is not straightforward to design a state-augmented TCNN-CD.
In fact, for complex architectures where the input modality requires spatial invariance (handled by convolutions), it is not straightforward to incorporate other modalities.
Here, PHLieNet offers a natural alternative: the TCNN-CD serves as the target network, while the hypernetwork modulates its kernels based on the system parameter.
Thus, in the following, we use the TCNN-CD as the target network, benefiting from parameter-based modulation via the hypernetwork and from the strong temporal modeling capacity of the TCNN-CD.
Regarding hyperparameters, we did not perform an exhaustive search for optimal values.
Our aim is not to achieve the best possible performance, but to demonstrate the viability of PHLieNet as a modeling framework.

\begin{table}[!h]
	\caption{Reference table for compared networks.} \label{tab:networks}
	    \centering{
		\begin{tabular}{ p{3cm} p{13cm}}
    		\hline
    		Reference name & Description\\
    		\hline
    		FFNN-P & A feedforward neural network, augmented on the state with the parameter. \\
    		LSTM-A & An LSTM neural network agnostic to the parameter.\\
    		LSTM-P & An LSTM neural network, augmented on the state with the parameter. \\
    		TCNN-CD-A & A causal dilated temporal CNN agnostic to the parameter.\\
    		PHLieNet\textsubscript{$(n_e, d_e)$} & PHLieNet with $n_e$ RBF anchors and embedding dimension $d_e$. The specific values of $(n_e, d_e)$ are chosen per system and listed in the corresponding results section.\\
    		\hline
	    \end{tabular}}
\end{table}

The networks are trained on a dataset that contains trajectories $\trainData \in \RR^{\np \times \nics \times \ntimesteps \times \dimState}$ generated from a set of parameter values denoted as \( \pTrainSet \), with cardinality $\np$.
For validation, we use the same set of parameters \( \pTrainSet \) but with trajectories initialized from different initial conditions.
Each model variant is trained with 5 independent random seeds, and the seed yielding the lowest validation loss is selected for evaluation.
We evaluate the methods in the auto-regressive forecasting (AR) setting.
For testing, we consider two distinct tasks.
In the interpolation task, we evaluate autoregressive prediction on parameter values within the training range, so \( \pTestInterp \subseteq \pTrainSet \).
In the extrapolation task, we assess the ability of the networks to generalize to unseen parameter values outside the training range.
Here, trajectories are sampled from a testing set of parameters \( \pTestExtrap \) with \( \pTestExtrap \cap \pTrainSet = \emptyset \).

We apply PHLieNet to a diverse set of dynamical systems spanning a broad range of behaviors and modeling challenges.
These systems include the Van der Pol oscillator, the Rössler attractor, the Finance system, and the Lorenz 3D system, spanning nonlinear oscillations and chaotic attractors across varying dynamical regimes.
Results for two additional systems (Chua's circuit and the Duffing oscillator) are reported in~\ref{sec:appendix:chuaduffing}.
For each system, we report results on the interpolation task, and additionally on the extrapolation task where noted.
The following sections provide details on each system and the corresponding experimental setup.
\subsection{The Van der Pol Oscillator}\label{sec:vdp}

The Van der Pol oscillator is a second-order nonlinear dynamical system, originally introduced by Balthasar van der Pol in the 1920s while studying electrical circuits containing vacuum tubes~\cite{van1926lxxxviii}.
It is governed by the following second-order differential equation:
\begin{equation}
\frac{d^2 x_1}{dt^2} - \mu(1 - x_1^2)\frac{d x_1}{dt} + x_1 = 0,
\label{eq:vanderpol}
\end{equation}
which can be rewritten as a system of first-order ODEs:
\begin{align}
\frac{d x_1}{dt} &= x_2, \\
\frac{d x_2}{dt} &= \mu (1 - x_1^2) x_2 - x_1,
\end{align}
where \( \state = [x_1, x_2]^T \) is the state variable with dimensionality \( \dimState=2 \) and \( \mu \in \mathbb{R}^+ \) is a scalar parameter that controls the degree of nonlinearity and the damping intensity.
The system exhibits qualitatively different behaviors depending on the value of $\mu$.
For small $\mu$, it behaves like a near-harmonic oscillator, whereas for larger $\mu$, it transitions to nonlinear relaxation oscillations with slow dynamics punctuated by rapid jumps.
This rich variety of behaviors makes the Van der Pol oscillator a widely used model for studying non-equilibrium phenomena in biological, chemical, and engineering systems~\cite{strogatz2018nonlinear}.

In our experiments, we vary the parameter \( \mu \) continuously in the range $[1.0,\, 8.0]$, sampling $N_\text{train}=50$ values using Sobol quasi-random sequences to ensure uniform coverage of the training regime.
We use an RK45 integrator with a solver time step \( \delta t = 0.001 \) and sample observations every \( \Delta t = 0.05 \) time units.
For each parameter value, we simulate \( \nicsTrain = 10 \) initial conditions, each integrated up to $t_\text{end}=20$ time units.
The noise level during training is set to $\trainnoise=5\%$.
For interpolation evaluation, we sample 10 parameter values within $[1.0,\, 8.0]$ using a held-out Sobol seed, simulating \( \nicsTest = 10 \) trajectories per parameter value up to $t_\text{end}=50$ time units.
For extrapolation evaluation, we use 10 parameter values outside the training range: $\mu \in \{0.4, 0.5, 0.6, 0.7, 0.8, 8.3, 8.6, 8.9, 9.2, 9.4\}$ (5 below and 5 above the training interval), simulated with the same protocol.
This setup allows us to assess both interpolation within the training regime and robustness to genuinely unseen dynamical regimes.

\begin{figure}[!hbt]
    \centering
    \includegraphics[width=0.75\textwidth]{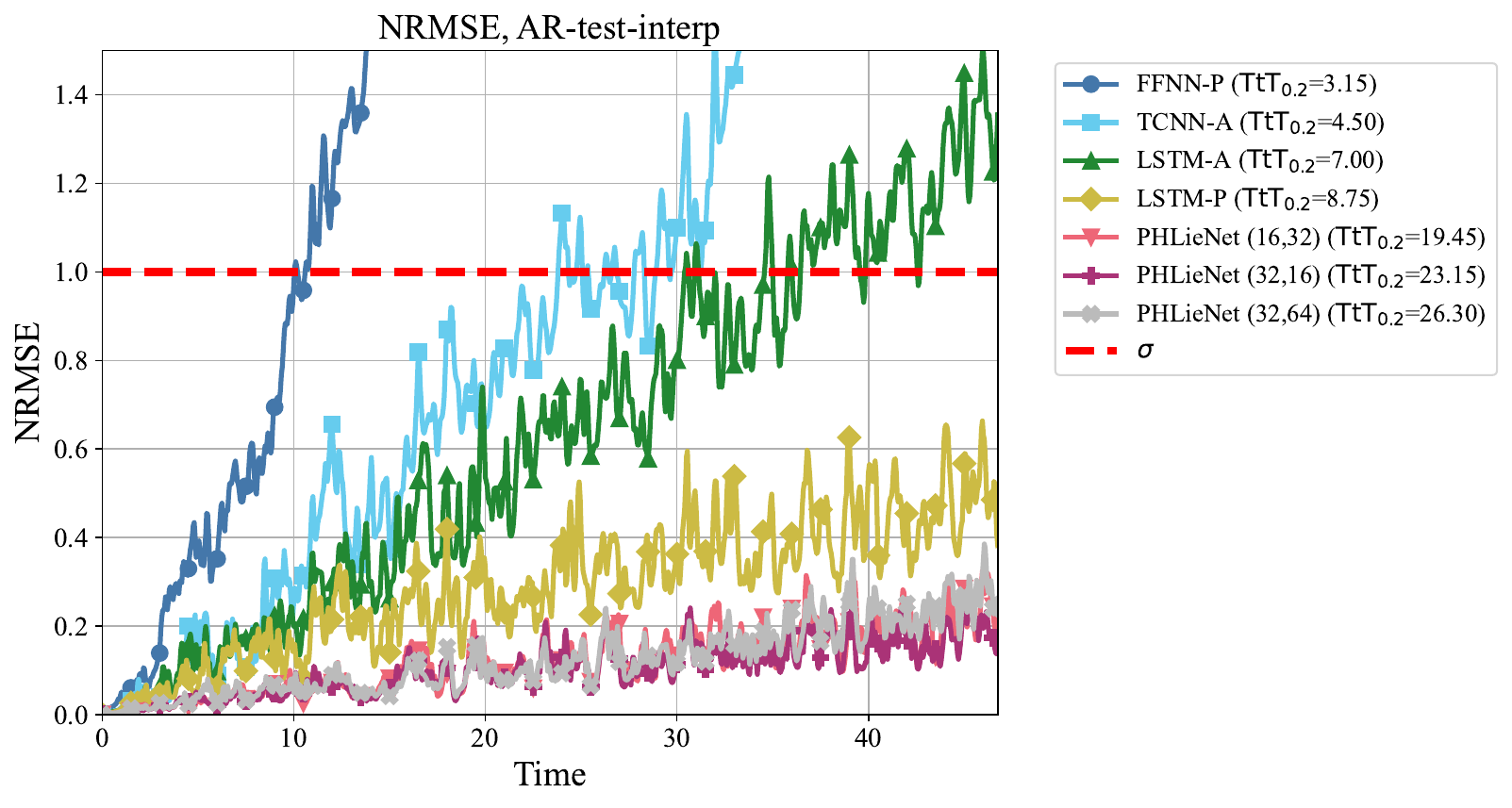}
    \caption{Normalized Root Mean Squared Error (NRMSE) evolution in time for the interpolation task.}
    \label{fig:vdp:interp:rmse_evolution}
\end{figure}

\begin{figure}[!htb]
    \centering
    \begin{subfigure}{0.45\textwidth}
        \centering
        \includegraphics[width=\textwidth]{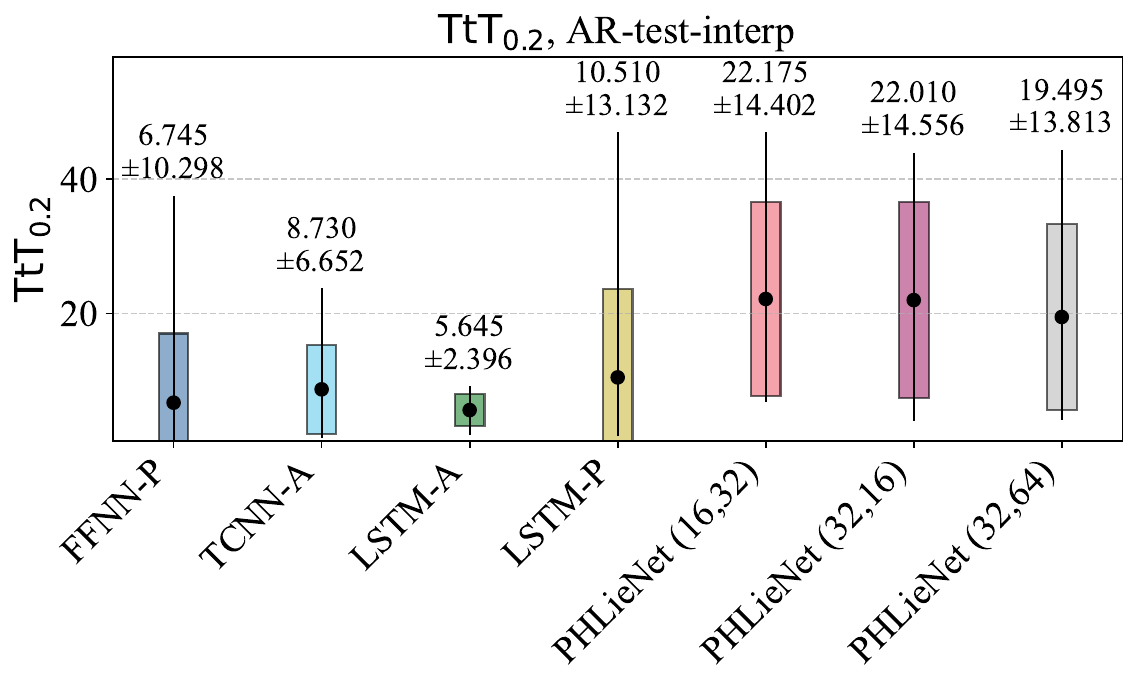}
        \caption{Time-to-Threshold (TtT) metric.}
        \label{fig:vdp:interp:ttt}
    \end{subfigure}
    \quad
    \begin{subfigure}{0.45\textwidth}
        \centering
        \includegraphics[width=\textwidth]{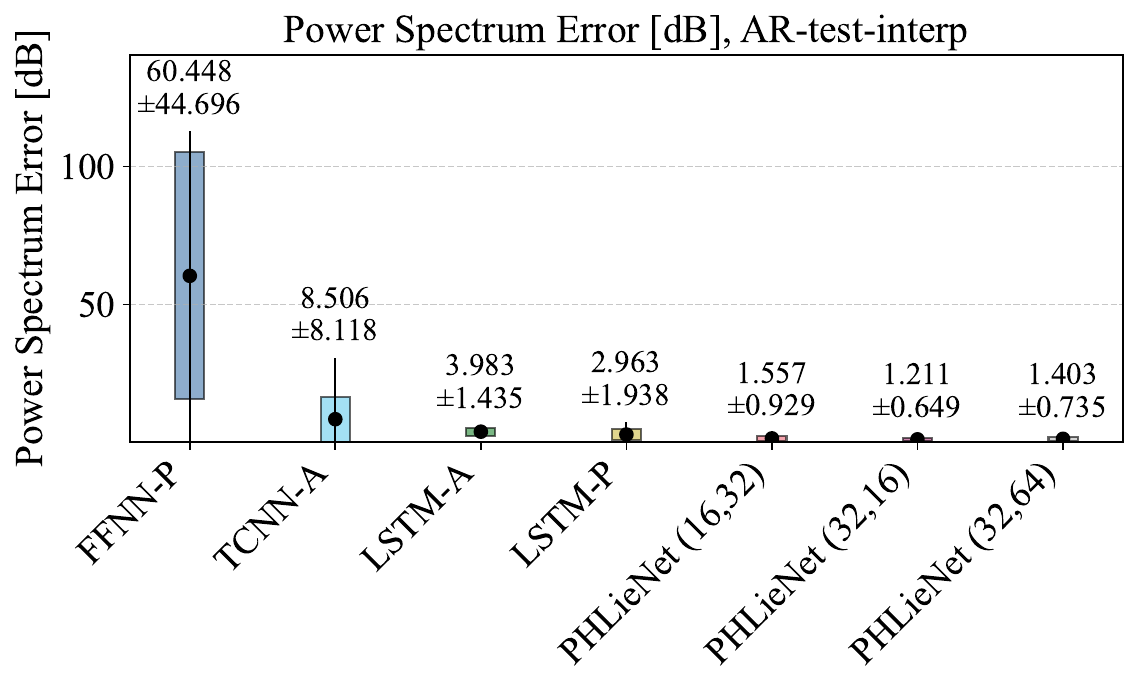}
        \caption{Power spectrum error.}
        \label{fig:vdp:interp:psd_error}
    \end{subfigure}
    \caption{
        Performance on the Van der Pol oscillator for the interpolation task.
        (a) Time-to-Threshold (TtT) metric.
        (b) Power spectrum error.
    }
    \label{fig:vdp:interp}
\end{figure}

In~\Cref{fig:vdp:interp:rmse_evolution,fig:vdp:interp}, we report results on the interpolation task.
The NRMSE evolution in~\Cref{fig:vdp:interp:rmse_evolution} shows that all three PHLieNet variants maintain a slower error increase than the baselines.
The aggregate $\ttt_{0.2}$, computed over all initial conditions and parameter values, is annotated in the legend for each model.
The per-parameter $\ttt_{0.2}$ values in~\Cref{fig:vdp:interp:ttt}, computed separately for each test parameter and averaged over initial conditions, confirm this advantage quantitatively.
PHLieNet\textsubscript{(32,16)} achieves a mean of 22.0 time units, and the other two variants reach 22.2 and 19.5, all substantially above LSTM-P (10.5), LSTM-A (5.6), and TCNN-A (8.7).
The power spectrum error in~\Cref{fig:vdp:interp:psd_error} confirms this: PHLieNet variants attain the lowest error values (1.21--1.56), well below LSTM-P (2.96), LSTM-A (3.98), and TCNN-A (8.51), demonstrating faithful spectral reconstruction within the training range.

\begin{figure}[!hbt]
    \centering
    \includegraphics[width=0.75\textwidth]{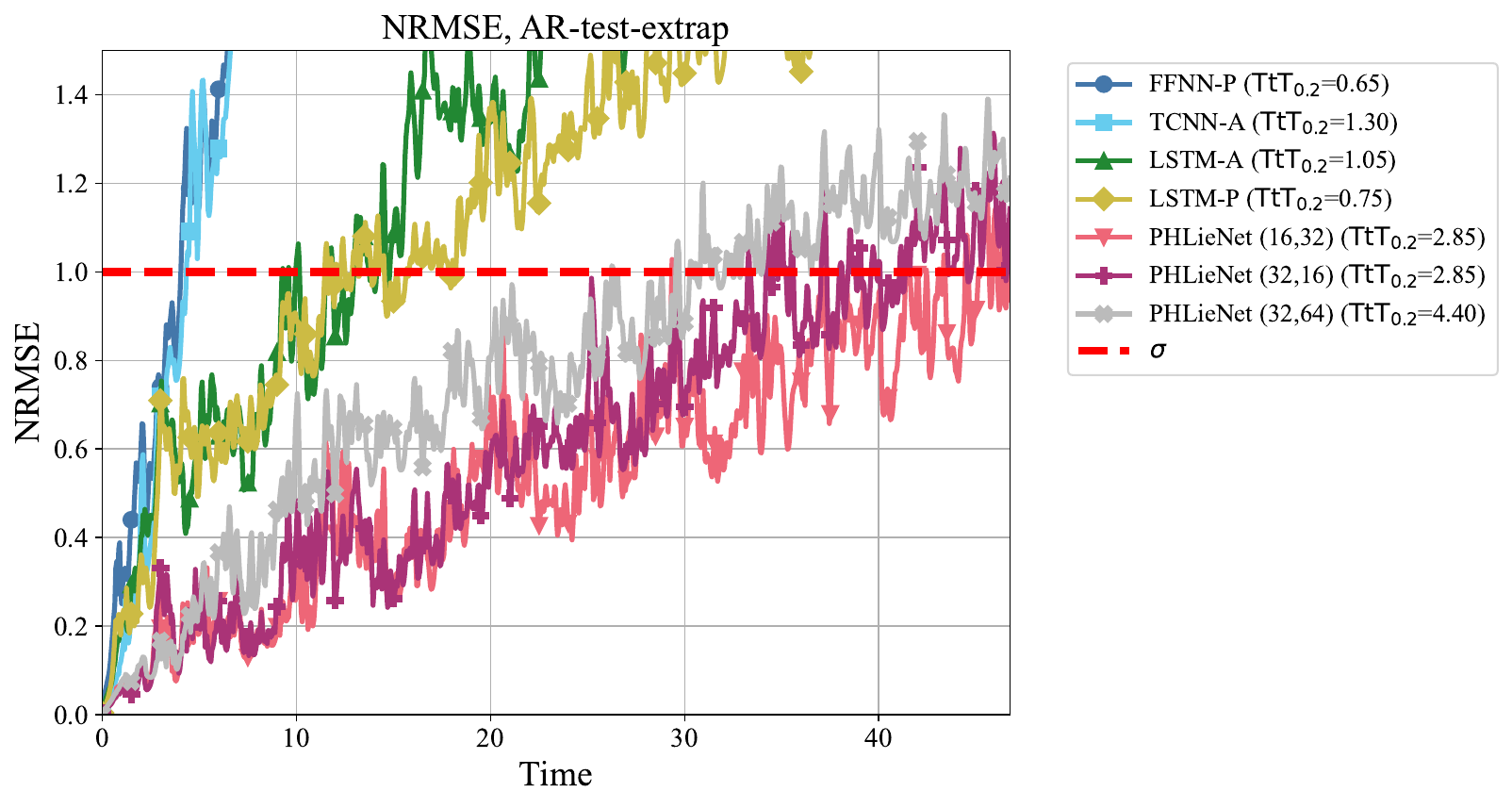}
    \caption{Normalized Root Mean Squared Error (NRMSE) evolution in time for the extrapolation task.}
    \label{fig:vdp:extrap:rmse_evolution}
\end{figure}

\begin{figure}[H]
    \centering
    \begin{subfigure}{0.45\textwidth}
        \centering
        \includegraphics[width=\textwidth]{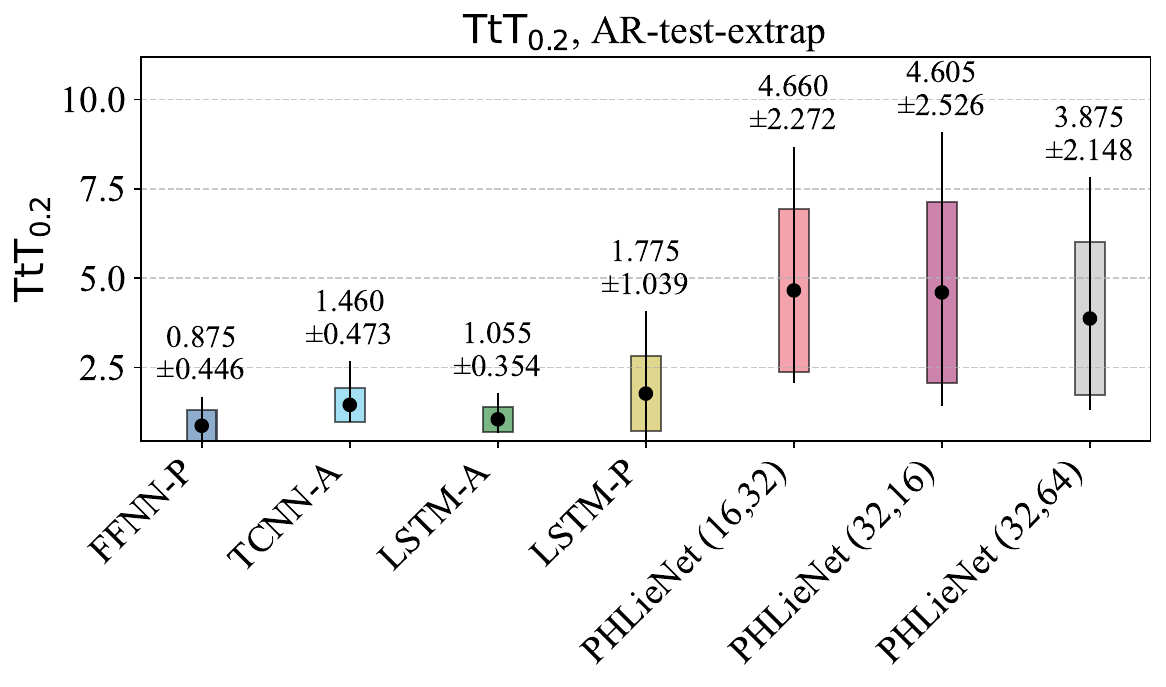}
        \caption{Time-to-Threshold (TtT) metric.}
        \label{fig:vdp:extrap:ttt}
    \end{subfigure}
    \quad
    \begin{subfigure}{0.45\textwidth}
        \centering
        \includegraphics[width=\textwidth]{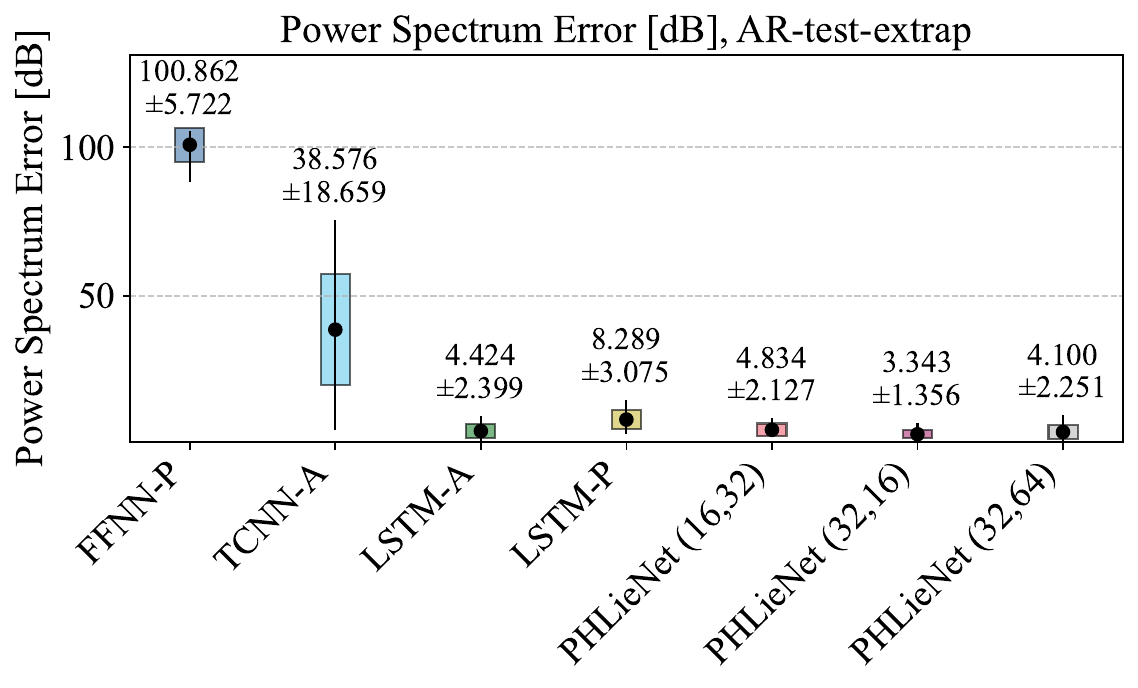}
        \caption{Power spectrum error.}
        \label{fig:vdp:extrap:psd_error}
    \end{subfigure}
    \caption{
        Performance on the Van der Pol oscillator for the extrapolation task.
        (a) Time-to-Threshold (TtT) metric.
        (b) Power spectrum error.
    }
    \label{fig:vdp:extrap}
\end{figure}

In~\Cref{fig:vdp:extrap:rmse_evolution,fig:vdp:extrap}, we evaluate performance on the more challenging extrapolation task, where parameter values lie outside the training range.
The NRMSE evolution in~\Cref{fig:vdp:extrap:rmse_evolution} shows that PHLieNet variants exhibit a consistently slower error growth than all baselines, even for genuinely unseen $\mu$ values.
The aggregate $\ttt_{0.2}$ annotated in the legend reflects this ordering.
The per-parameter $\ttt_{0.2}$ values in~\Cref{fig:vdp:extrap:ttt} quantify the advantage across individual test conditions: all three PHLieNet variants outperform all baselines, with PHLieNet\textsubscript{(16,32)} and PHLieNet\textsubscript{(32,16)} achieving means of 4.66 and 4.61 time units respectively, and PHLieNet\textsubscript{(32,64)} reaching 3.88, all substantially above LSTM-P (1.78) and the parameter-agnostic models.
The smooth limit-cycle structure of the Van der Pol oscillator enables effective weight-space extrapolation.
The power spectrum error in~\Cref{fig:vdp:extrap:psd_error} confirms this: PHLieNet\textsubscript{(32,16)} achieves the lowest spectral error (3.34), and all three PHLieNet variants compare favourably against LSTM-P (8.29).
\subsection{The Rössler System}
\label{sec:rossler}

The Rössler system, introduced by Otto Rössler in 1976 \cite{rossler1976equation}, is a set of three coupled nonlinear ordinary differential equations (ODEs) that exhibit chaotic behavior.
Rössler initially developed this system as a simplified model to explore chaos in continuous dynamical systems.
Its simplicity, both in terms of form and computational requirements, has made it a widely studied example in chaos theory.

The system is defined by the following set of ODEs:
\begin{align}
\dot{x_1} &= -x_2 - x_3, \\
\dot{x_2} &= x_1 + a x_2, \\
\dot{x_3} &= b + x_3(x_1 - c),
\end{align}
where \( \state=[ x_1, x_2, x_3]^T \in \RR^3 \) is the state and \( a, b, c \) are scalar parameters controlling the dynamics.
The parameters \(a\), \(b\), and \(c\) in the Rössler system play distinct roles in shaping its dynamics: \(a\) controls the linear damping in the \(y\)-equation, \(b\) introduces a constant drift in the \(z\)-equation, and \(c\) modulates the nonlinearity in the \(z\)-equation through coupling with \(x\).
As these parameters are varied, the system transitions from simple periodic oscillations to chaotic behavior, which is often visualized in the form of attractors.
For specific parameter values, the system generates a fractal attractor known as the Rössler attractor, one of the most iconic examples of deterministic chaos.
The Rössler system exhibits rich dynamical behavior, including bifurcations, periodic orbits, and chaotic attractors, depending on the parameter values.
The system has been applied across fields such as chemical reactions, biological systems, and electronics, where chaos plays a critical role~\cite{strogatz2018nonlinear}.
This makes it a valuable testbed for both theoretical studies and practical applications.

In our experiments, we fix \( a = b = 0.2 \) and vary \( c \) continuously in $[3.0,\, 9.0]$, sampling $N_\text{train}=50$ values via Sobol quasi-random sequences to explore different regimes, from periodic motion to deterministic chaos.
We simulate trajectories using a fourth-order Runge--Kutta integrator (\texttt{RK45}) with a solver time step \( \delta t = 0.001 \), and sample the solution every \( \Delta t = 0.1 \) time units.
A transient of $t_\text{trans}=300$ time units is discarded before recording each trajectory.
For each parameter value, we simulate \( \nicsTrain = 10 \) initial conditions, each integrated up to $t_\text{end}=20$ time units.
The noise level during training is set to $\trainnoise=5\%$.

For interpolation evaluation, we sample 10 parameter values within $[3.0,\, 9.0]$ using a held-out Sobol seed and simulate \( \nicsTest=10 \) trajectories per value up to $t_\text{end}=100$ time units.
For extrapolation evaluation, we use 10 parameter values outside the training range: $c \in \{1.8, 2.0, 2.3, 2.5, 2.8, 9.2, 9.5, 9.7, 10.0, 10.2\}$ (5 below and 5 above the training interval), simulated with the same protocol.

\begin{figure}[!hbt]
    \centering
    \includegraphics[width=0.75\textwidth]{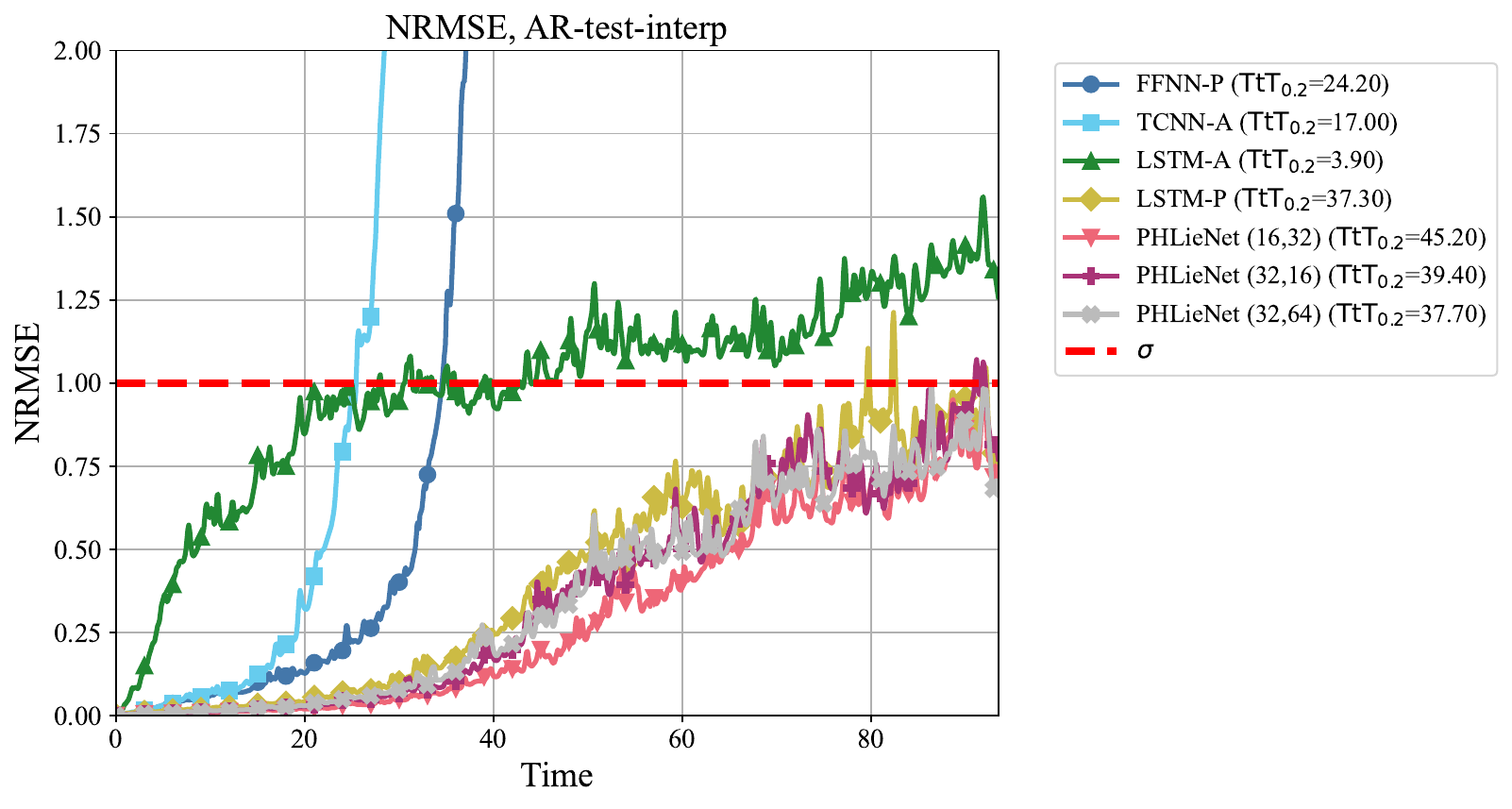}
    \caption{Normalized Root Mean Squared Error (NRMSE) evolution in time for the interpolation task.}
    \label{fig:roessler:interp:rmse_evolution}
\end{figure}

\begin{figure}[H]
    \centering
    \begin{subfigure}{0.45\textwidth}
        \centering
        \includegraphics[width=\textwidth]{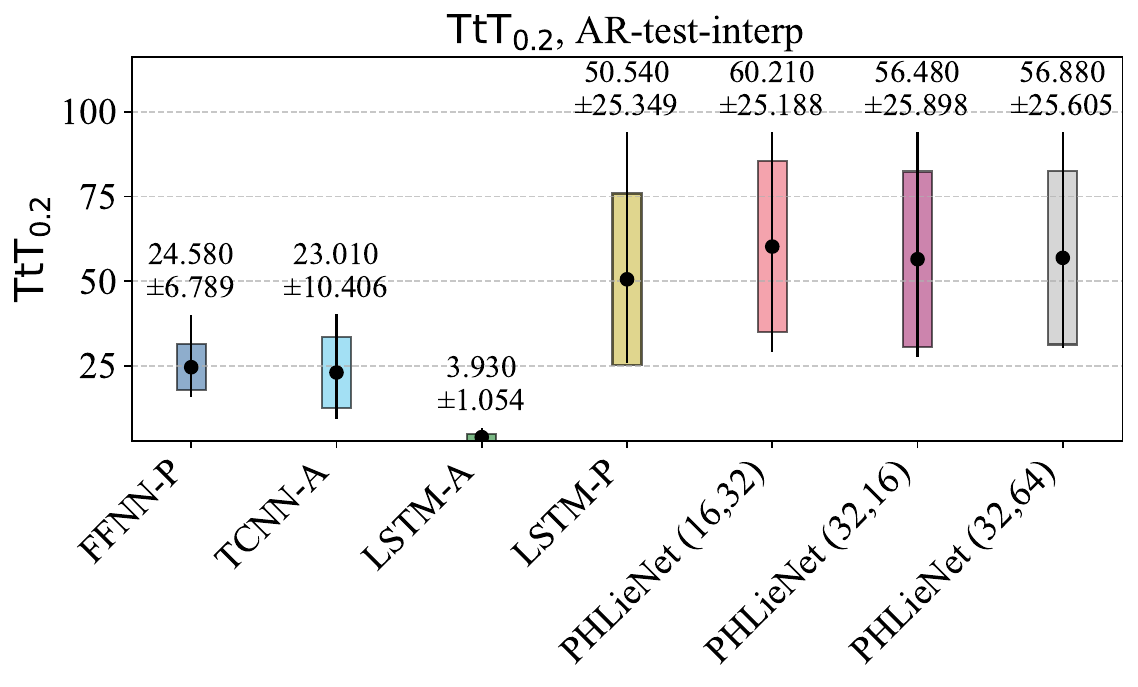}
        \caption{Time-to-Threshold (TtT) metric.}
        \label{fig:roessler:interp:ttt}
    \end{subfigure}
    \quad
    \begin{subfigure}{0.45\textwidth}
        \centering
        \includegraphics[width=\textwidth]{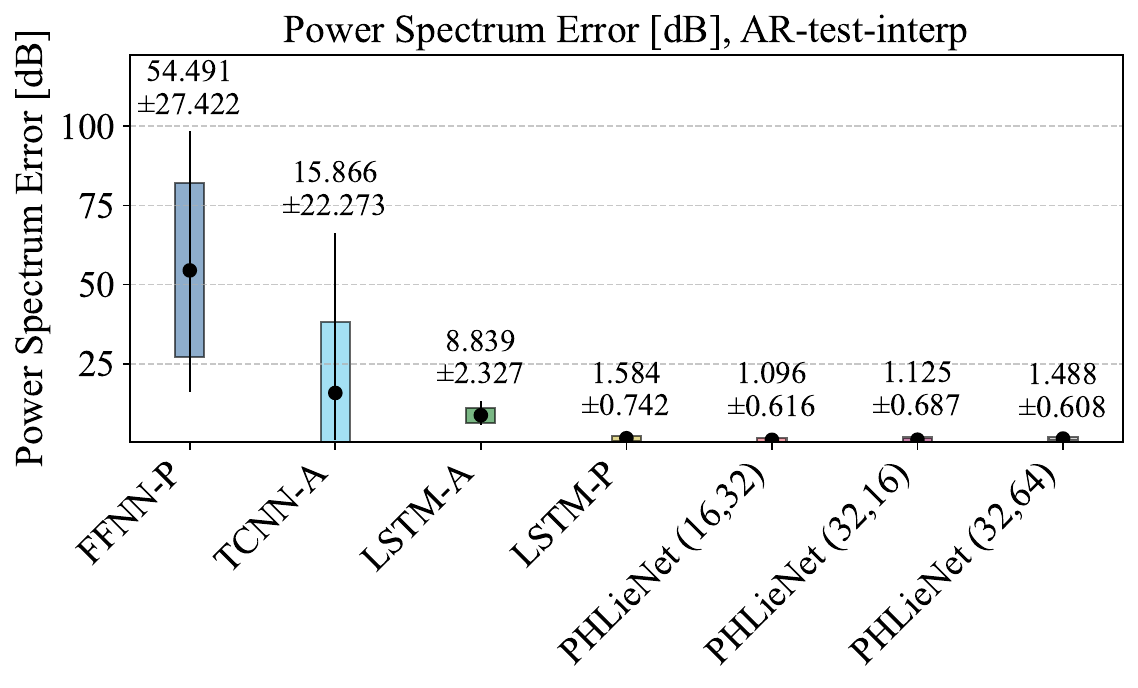}
        \caption{Power spectrum error.}
        \label{fig:roessler:interp:psd_error}
    \end{subfigure}
    \caption{
        Model performance on the Rössler system for the interpolation task.
        (a) Time-to-Threshold (TtT) metric.
        (b) Power spectrum error.
    }
    \label{fig:roessler:interp}
\end{figure}

In~\Cref{fig:roessler:interp:rmse_evolution,fig:roessler:interp}, we report results on the interpolation task.
The NRMSE evolution in~\Cref{fig:roessler:interp:rmse_evolution} shows that all three PHLieNet variants maintain a slower error increase than the baselines.
The aggregate $\ttt_{0.2}$ annotated in the legend reflects this ordering.
The per-parameter $\ttt_{0.2}$ values in~\Cref{fig:roessler:interp:ttt} confirm this quantitatively: PHLieNet\textsubscript{(16,32)} leads with a mean of 60.2 time units, followed by PHLieNet\textsubscript{(32,64)} (56.9) and PHLieNet\textsubscript{(32,16)} (56.5), all substantially above LSTM-P (50.5) and the parameter-agnostic baselines LSTM-A (3.9) and TCNN-A (23.0).
The power spectrum error in~\Cref{fig:roessler:interp:psd_error} reinforces this: PHLieNet variants achieve the lowest values (1.10--1.49), below LSTM-P (1.58), confirming faithful spectral reconstruction within the training range.

\begin{figure}[H]
    \centering
    \includegraphics[width=0.75\textwidth]{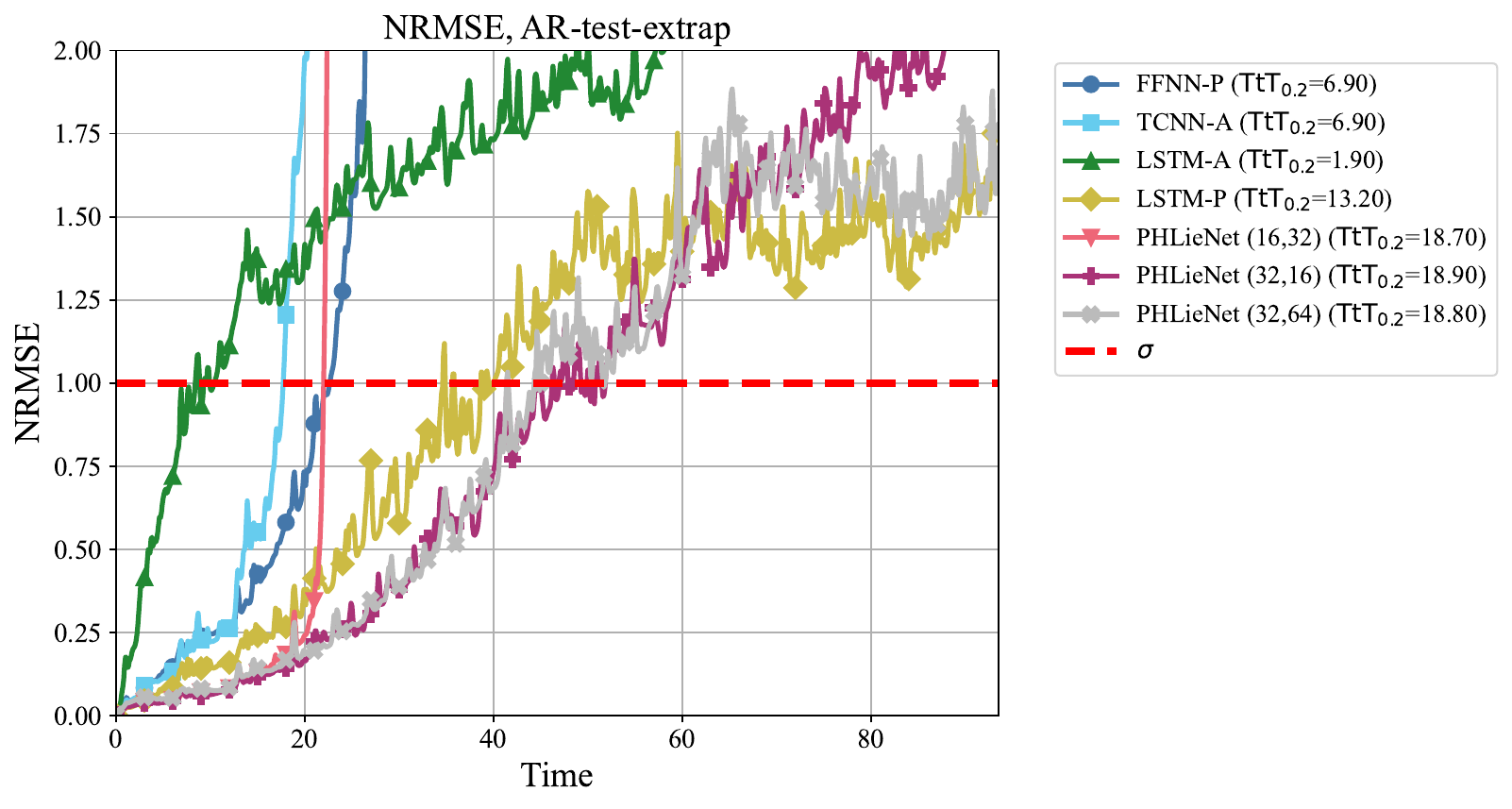}
    \caption{Normalized Root Mean Squared Error (NRMSE) evolution in time for the extrapolation task.}
    \label{fig:roessler:extrap:rmse_evolution}
\end{figure}

\begin{figure}[H]
    \centering
    \begin{subfigure}{0.45\textwidth}
        \centering
        \includegraphics[width=\textwidth]{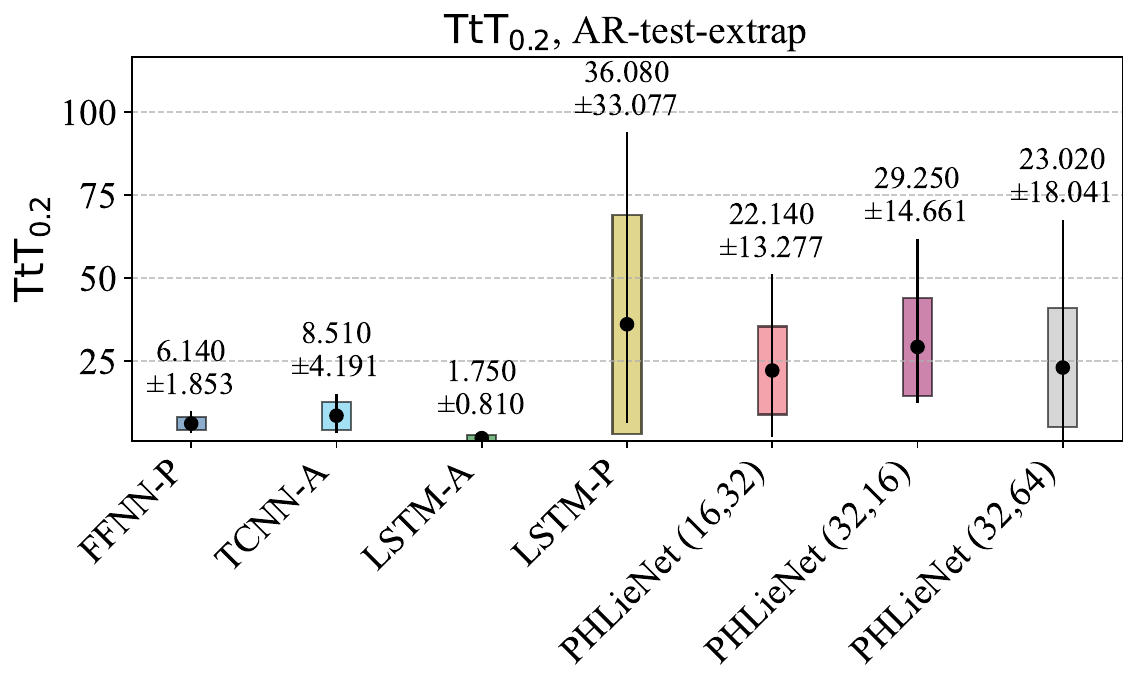}
        \caption{Time-to-Threshold (TtT) metric.}
        \label{fig:roessler:extrap:ttt}
    \end{subfigure}
    \quad
    \begin{subfigure}{0.45\textwidth}
        \centering
        \includegraphics[width=\textwidth]{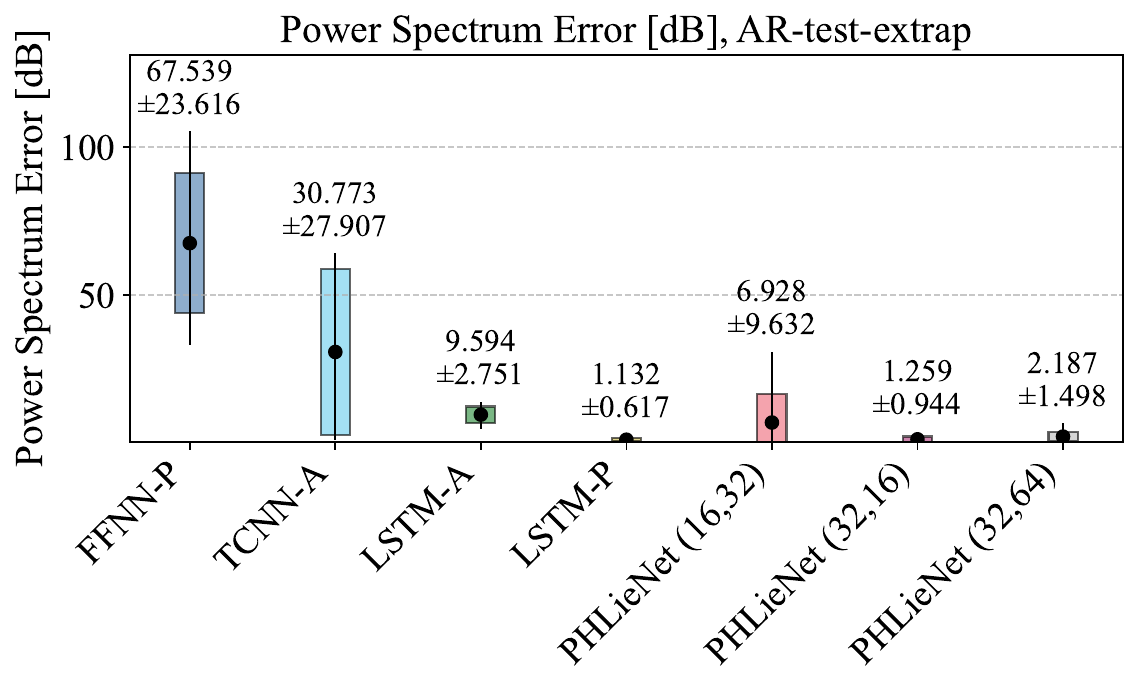}
        \caption{Power spectrum error.}
        \label{fig:roessler:extrap:psd_error}
    \end{subfigure}
    \caption{
        Model performance on the Rössler system for the extrapolation task.
        (a) Time-to-Threshold (TtT) metric.
        (b) Power spectrum error.
    }
    \label{fig:roessler:extrap}
\end{figure}

On the parameter extrapolation task in~\Cref{fig:roessler:extrap:rmse_evolution,fig:roessler:extrap}, the picture is more nuanced.
The NRMSE evolution in~\Cref{fig:roessler:extrap:rmse_evolution} shows that the $n_e{=}32$ PHLieNet variants maintain a slower error growth than the $n_e{=}16$ variant.
The aggregate $\ttt_{0.2}$ in the legend reflects this.
The per-parameter $\ttt_{0.2}$ values in~\Cref{fig:roessler:extrap:ttt} reveal that LSTM-P leads on this split with a mean of 36.1 time units.
PHLieNet\textsubscript{(32,16)} achieves the highest mean among PHLieNet variants at 29.3, followed by PHLieNet\textsubscript{(32,64)} (23.0) and PHLieNet\textsubscript{(16,32)} (22.1).
This suggests that a larger number of RBF anchors ($n_e{=}32$) provides more robust parameter-space coverage under extrapolation.
The power spectrum error in~\Cref{fig:roessler:extrap:psd_error} shows a similar trend: LSTM-P achieves the lowest spectral error (1.13), with PHLieNet\textsubscript{(32,16)} close behind (1.26), while the other PHLieNet variants and all agnostic baselines show substantially higher values.

\subsection{The Finance System}
\label{sec:finance}

The Finance system is a three-dimensional nonlinear dynamical system introduced by Ma and Chen~\cite{ma2001finance}.
It models the interaction between interest rate, investment demand, and price index in a macroeconomic setting.
Its equations are:
\begin{align}
    \dot{x}_1 &= \Bigl(\tfrac{1}{b} - a\Bigr) x_1 + x_3 + x_1 x_2, \\
    \dot{x}_2 &= -b x_2 - x_1^2, \\
    \dot{x}_3 &= -x_1 - c x_3,
\end{align}
where $\state = [x_1, x_2, x_3]^T \in \RR^3$ represents the interest rate, investment demand, and price index respectively.
The scalar parameter $a > 0$ is the savings rate, $b > 0$ is the per-unit investment cost, and $c > 0$ is the elasticity of commercial demand.
As $a$ varies, the system transitions between fixed-point, periodic, and chaotic regimes, which makes it a rich testbed for parameter-dependent forecasting.
In our experiments, $a$ is sampled continuously from $[1.0,\, 3.5]$ using $N_\text{train}=50$ Sobol quasi-random points.
We fix $b = 0.2$ and $c = 1.0$.
We use a Runge--Kutta integrator (\texttt{RK45}) with solver step $\delta t = 0.001$ and sampling interval $\Delta t = 0.1$ time units.
A transient of $t_\text{trans}=200$ time units is discarded before recording each trajectory.
For each parameter value, we simulate $\nicsTrain=10$ initial conditions, each integrated up to $t_\text{end}=100$ time units.
The noise level during training is set to $\trainnoise=5\%$.

\begin{figure}[!hbt]
    \centering
    \includegraphics[width=0.70\textwidth]{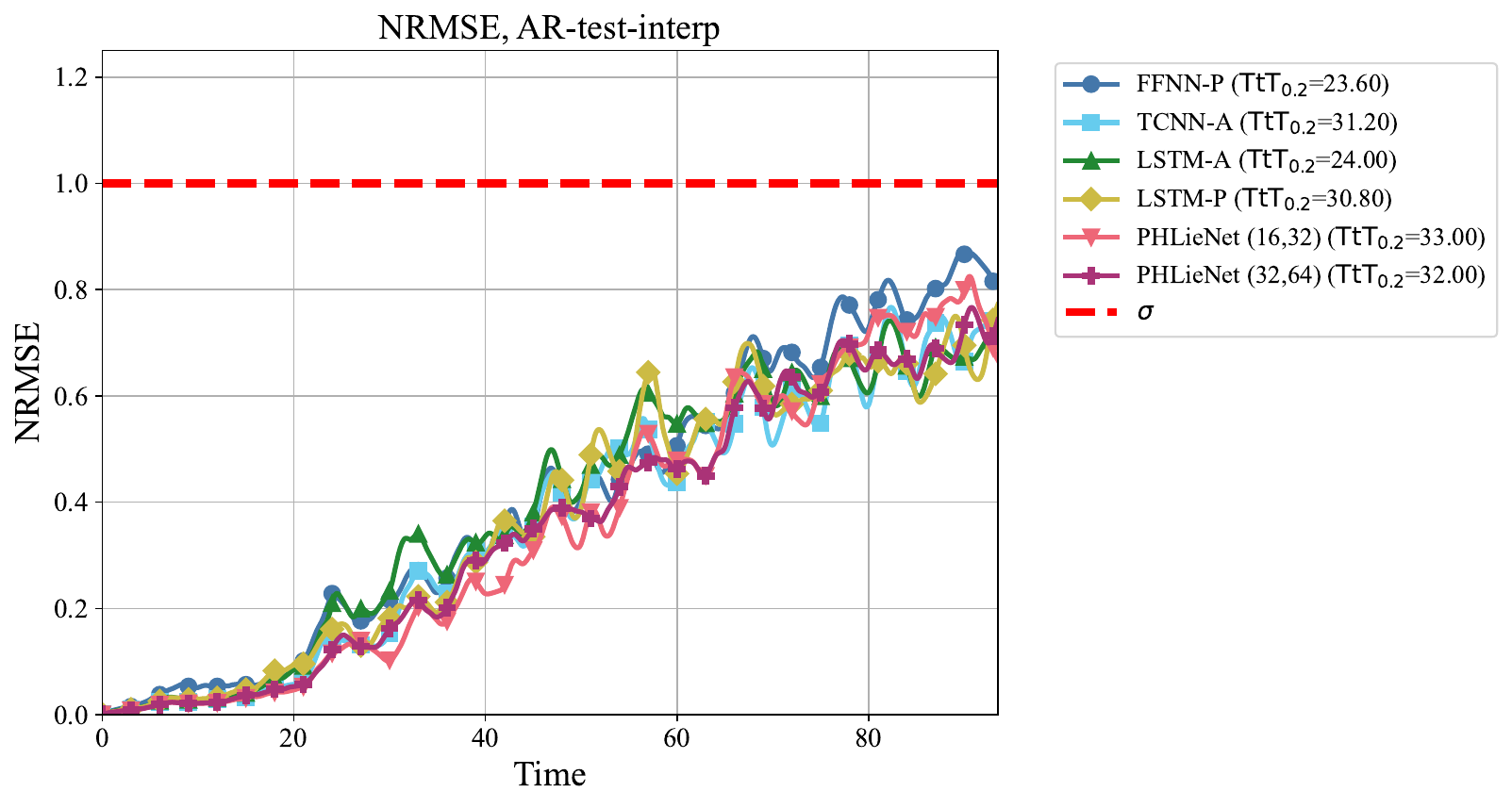}
    \caption{Normalized Root Mean Squared Error (NRMSE) evolution in time for the interpolation task.}
    \label{fig:finance:interp:rmse_evolution}
\end{figure}

For interpolation evaluation, we sample 10 parameter values within $[1.0,\, 3.5]$ using a held-out Sobol seed and simulate \( \nicsTest=10 \) trajectories per value up to $t_\text{end}=100$ time units.
For extrapolation evaluation, we use 10 parameter values outside the training range: $a \in \{0.55, 0.65, 0.75, 0.85, 0.95, 3.55, 3.60, 3.65, 3.70, 3.75\}$ (5 below and 5 above the training interval), simulated with the same protocol.

\begin{figure}[!hbt]
    \centering
    \begin{subfigure}{0.45\textwidth}
        \centering
        \includegraphics[width=\textwidth]{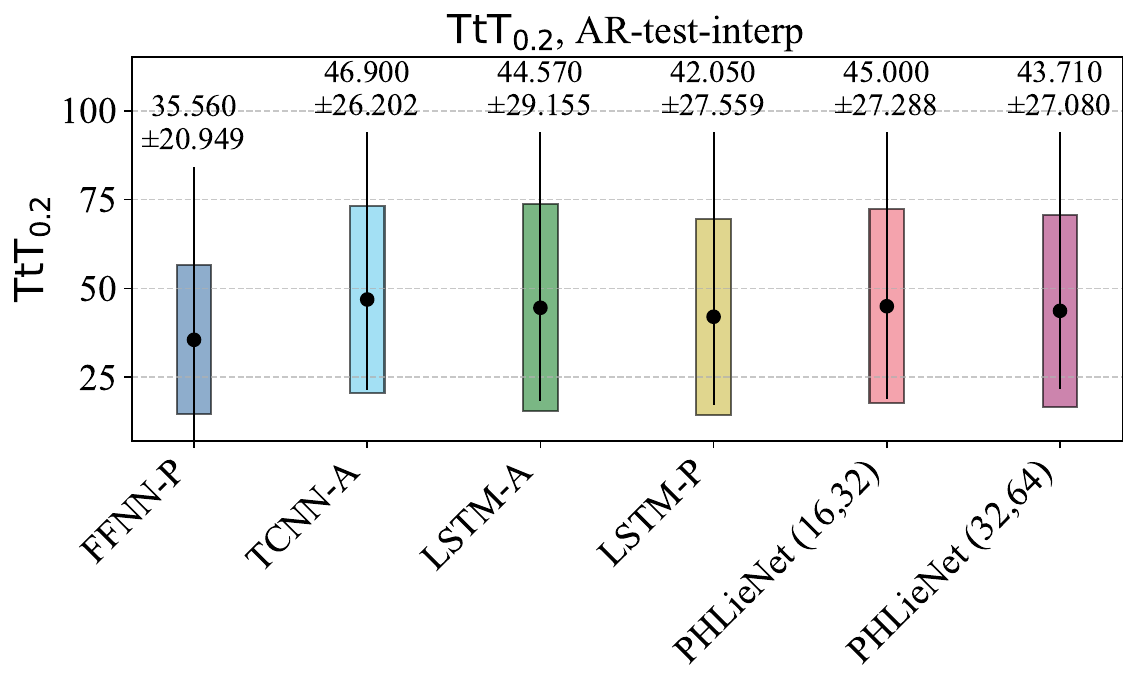}
        \caption{Time-to-Threshold (TtT) metric.}
        \label{fig:finance:interp:ttt}
    \end{subfigure}
    \quad
    \begin{subfigure}{0.45\textwidth}
        \centering
        \includegraphics[width=\textwidth]{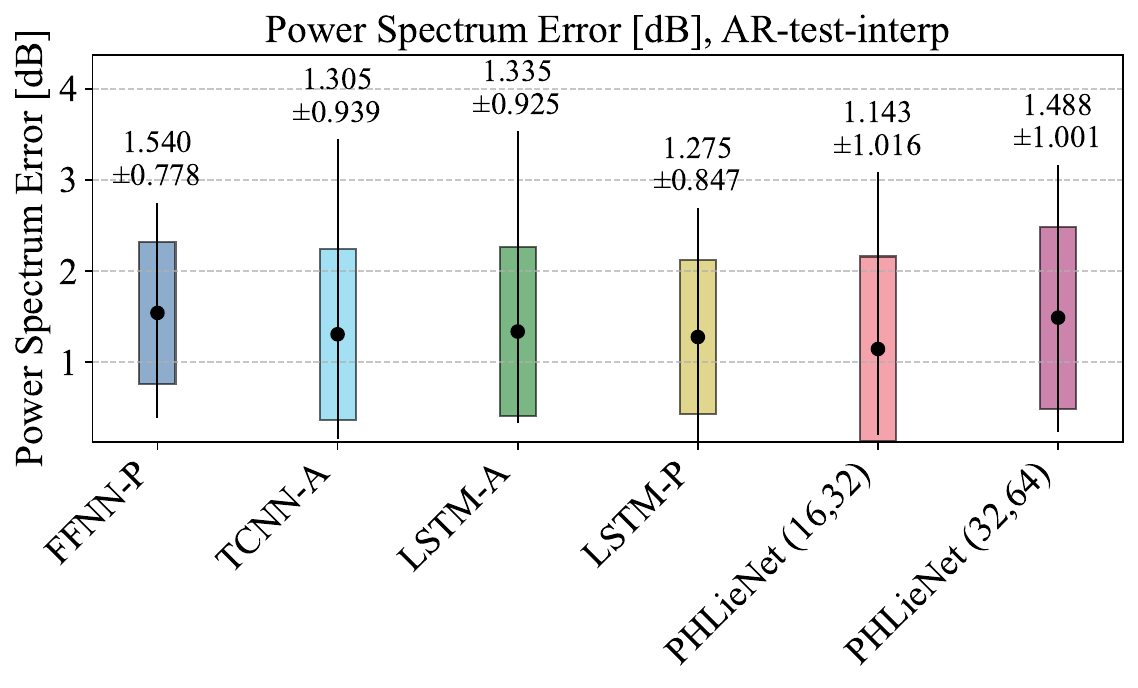}
        \caption{Power spectrum error.}
        \label{fig:finance:interp:psd_error}
    \end{subfigure}
    \caption{
        Model performance on the Finance system for the interpolation task.
        (a) Time-to-Threshold (TtT) metric.
        (b) Power spectrum error.
    }
    \label{fig:finance:interp}
\end{figure}

In~\Cref{fig:finance:interp:rmse_evolution,fig:finance:interp}, we benchmark performance on the interpolation task.
The NRMSE evolution in~\Cref{fig:finance:interp:rmse_evolution} shows PHLieNet\textsubscript{(16,32)} maintaining the slowest error growth among all models.
The aggregate $\ttt_{0.2}$ values annotated in the legend reflect this ordering: PHLieNet\textsubscript{(16,32)} 45.0, LSTM-P 42.1, PHLieNet\textsubscript{(32,64)} 32.0, TCNN-A 31.2, with the agnostic baselines trailing.
The per-parameter $\ttt_{0.2}$ values in~\Cref{fig:finance:interp:ttt} reveal a tighter spread: TCNN-A (46.9) and PHLieNet\textsubscript{(16,32)} (45.0) lead, followed by LSTM-A (44.6), PHLieNet\textsubscript{(32,64)} (43.7), and LSTM-P (42.1), with FFNN-P trailing (35.6).
The large standard deviation ($\approx 27$) across test parameter values reflects the heterogeneous difficulty of the parameter range.
The power spectrum error in~\Cref{fig:finance:interp:psd_error} reinforces PHLieNet\textsubscript{(16,32)}'s advantage: it achieves the lowest spectral error of all models (1.14), below LSTM-P (1.28) and all other baselines.

\begin{figure}[!hbt]
    \centering
    \includegraphics[width=0.75\textwidth]{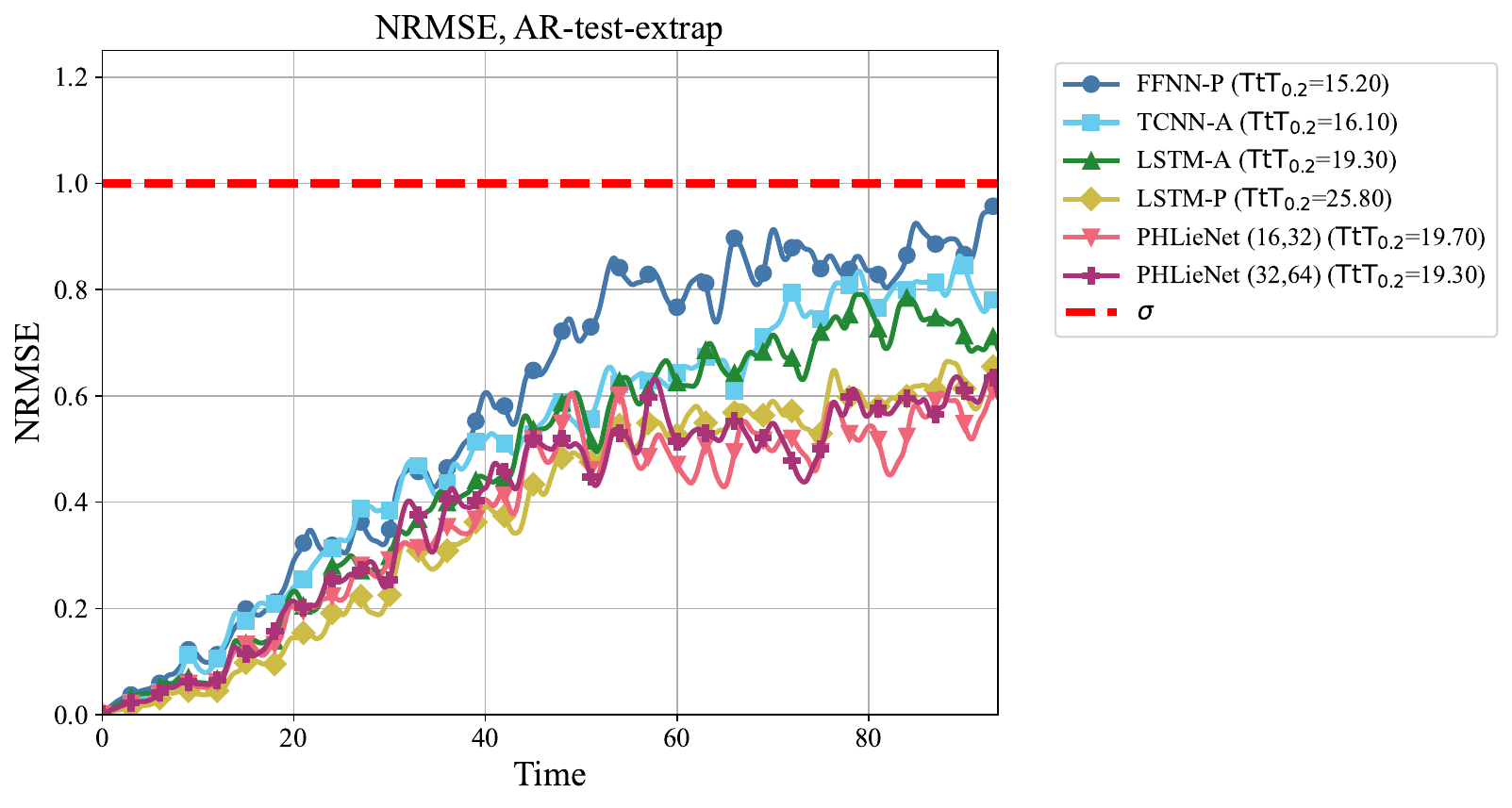}
    \caption{Normalized Root Mean Squared Error (NRMSE) evolution in time for the extrapolation task.}
    \label{fig:finance:extrap:rmse_evolution}
\end{figure}

\begin{figure}[!hbt]
    \centering
    \begin{subfigure}{0.45\textwidth}
        \centering
        \includegraphics[width=\textwidth]{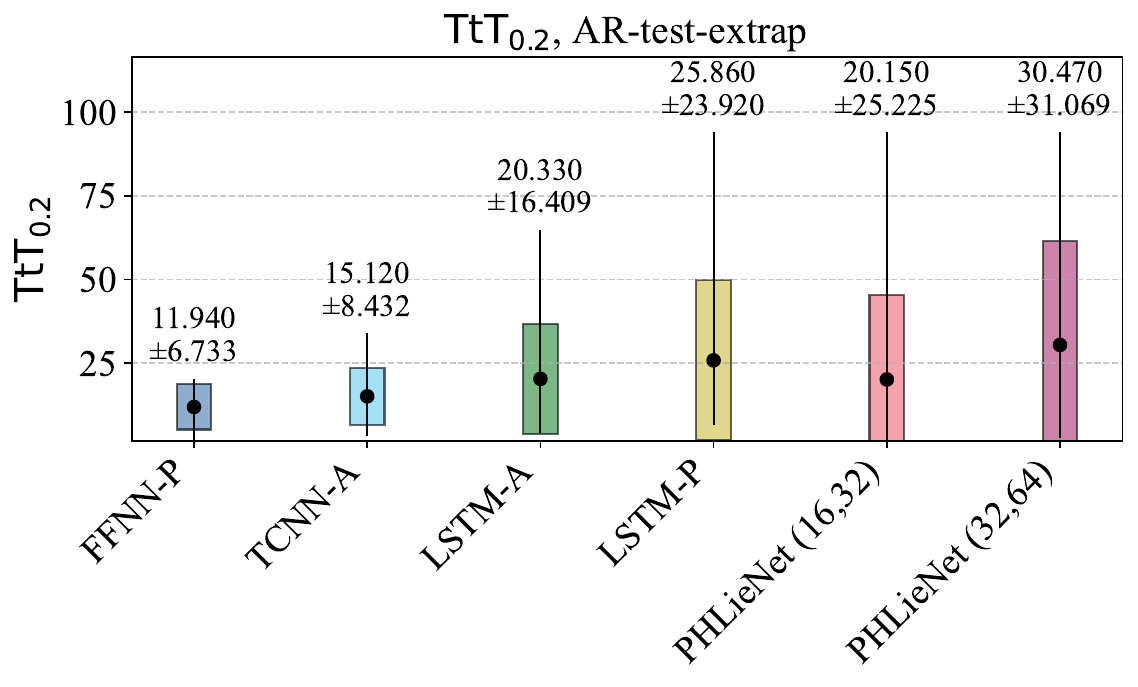}
        \caption{Time-to-Threshold (TtT) metric.}
        \label{fig:finance:extrap:ttt}
    \end{subfigure}
    \quad
    \begin{subfigure}{0.45\textwidth}
        \centering
        \includegraphics[width=\textwidth]{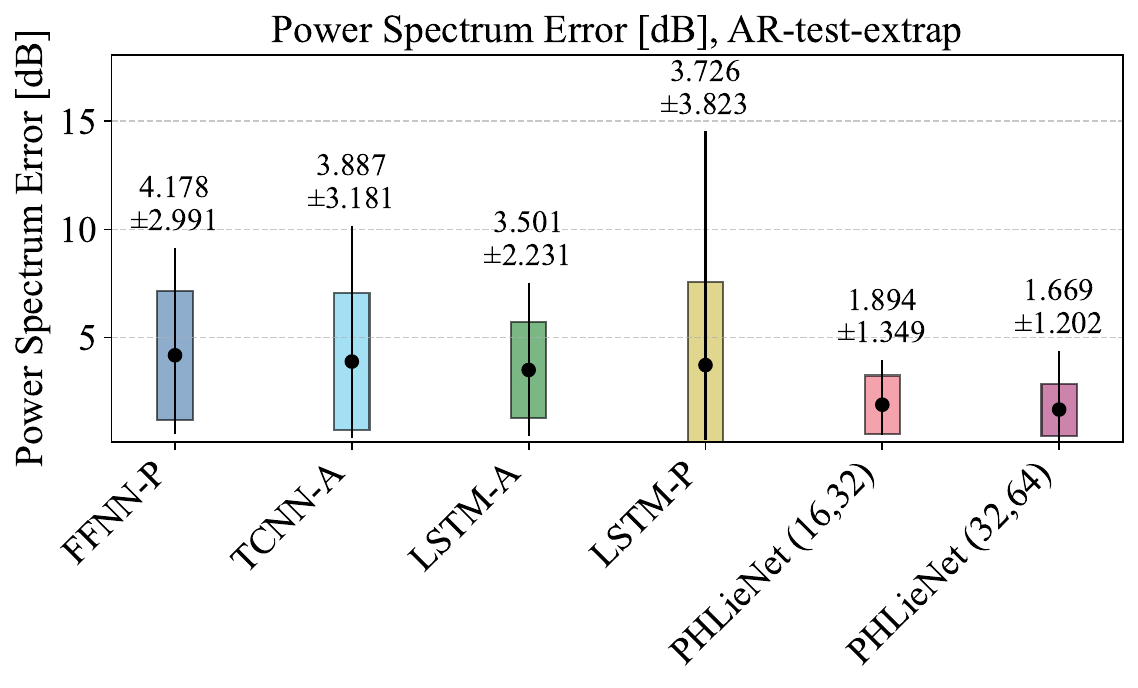}
        \caption{Power spectrum error.}
        \label{fig:finance:extrap:psd_error}
    \end{subfigure}
    \caption{
        Model performance on the Finance system for the extrapolation task.
        (a) Time-to-Threshold (TtT) metric.
        (b) Power spectrum error.
    }
    \label{fig:finance:extrap}
\end{figure}

On the parameter extrapolation task in~\Cref{fig:finance:extrap:rmse_evolution,fig:finance:extrap}, PHLieNet's advantage in parameter generalization becomes more pronounced.
The NRMSE evolution in~\Cref{fig:finance:extrap:rmse_evolution} shows LSTM-P maintaining the highest aggregate $\ttt_{0.2}$ in the legend (25.9), with PHLieNet variants close behind (16,32: 20.1, 32,64: 19.3) and the agnostic baselines trailing.
The per-parameter $\ttt_{0.2}$ values in~\Cref{fig:finance:extrap:ttt} reveal a different ordering: PHLieNet\textsubscript{(32,64)} leads with 30.5 time units, outperforming LSTM-P (25.9), PHLieNet\textsubscript{(16,32)} (20.2), LSTM-A (20.3), TCNN-A (15.1), and FFNN-P (11.9).
The power spectrum error in~\Cref{fig:finance:extrap:psd_error} shows the clearest separation: PHLieNet variants achieve substantially lower spectral error (1.67--1.89) compared to all baselines (LSTM-A 3.50, LSTM-P 3.73, TCNN-A 3.89, FFNN-P 4.18).
This confirms that weight-space interpolation faithfully reproduces the spectral structure of unseen regimes.

\subsection{The Lorenz 3D System}

The Lorenz system, introduced by Edward Lorenz in 1963~\cite{lorenz1963deterministic}, is a three-dimensional nonlinear dynamical system governed by
\begin{align}
\dot{x_1} &= \sigma (x_2 - x_1), \\
\dot{x_2} &= x_1 (\rho - x_3) - x_2, \\
\dot{x_3} &=  x_1 x_2 - \beta x_3,
\end{align}
where \( \state = [x_1, x_2, x_3]^T \in \mathbb{R}^3 \) and \( \sigma, \beta, \rho \in \mathbb{R} \) control the dynamics.
We use this system to demonstrate that PHLieNet can capture qualitatively different dynamical regimes within a single learned model.
Varying the bifurcation parameter $\rho$ produces drastically different long-run behaviors.
For $\rho < \rho_c \approx 24.74$, the system settles onto one of two stable fixed points $\mathbf{C}^\pm = (\pm\sqrt{\beta(\rho-1)},\,\pm\sqrt{\beta(\rho-1)},\,\rho-1)$, while for $\rho > \rho_c$ it exhibits deterministic chaos and the iconic butterfly attractor.
A single parametric model must therefore represent both convergent fixed-point dynamics and sensitive dependence on initial conditions.

We fix $\sigma = 10$ and $\beta = \frac{8}{3}$, and vary $\rho \in [10.0,\, 35.0]$, sampling $N_\text{train}=50$ values via Sobol quasi-random sequences.
An RK45 integrator with solver step $\delta t = 0.001$ is used, with observations sampled every $\Delta t = 0.02$ time units and a context window of $L=32$ steps.
For each training parameter, we simulate $\nicsTrain = 5$ initial conditions up to $t_\text{end}=20$ time units.
Training noise is set to $\trainnoise=5\%$.
For interpolation evaluation, we use ten parameter values $\rho \in \{12, 16, 20, 22, 24, 26, 28, 30, 32, 34\}$, which span both sides of the bifurcation, and simulate $\nicsTest=20$ trajectories per value up to $t_\text{end}=30$ time units.
\begin{figure}[!hbt]
    \centering
    \begin{subfigure}{0.45\textwidth}
        \centering
        \includegraphics[width=\textwidth]{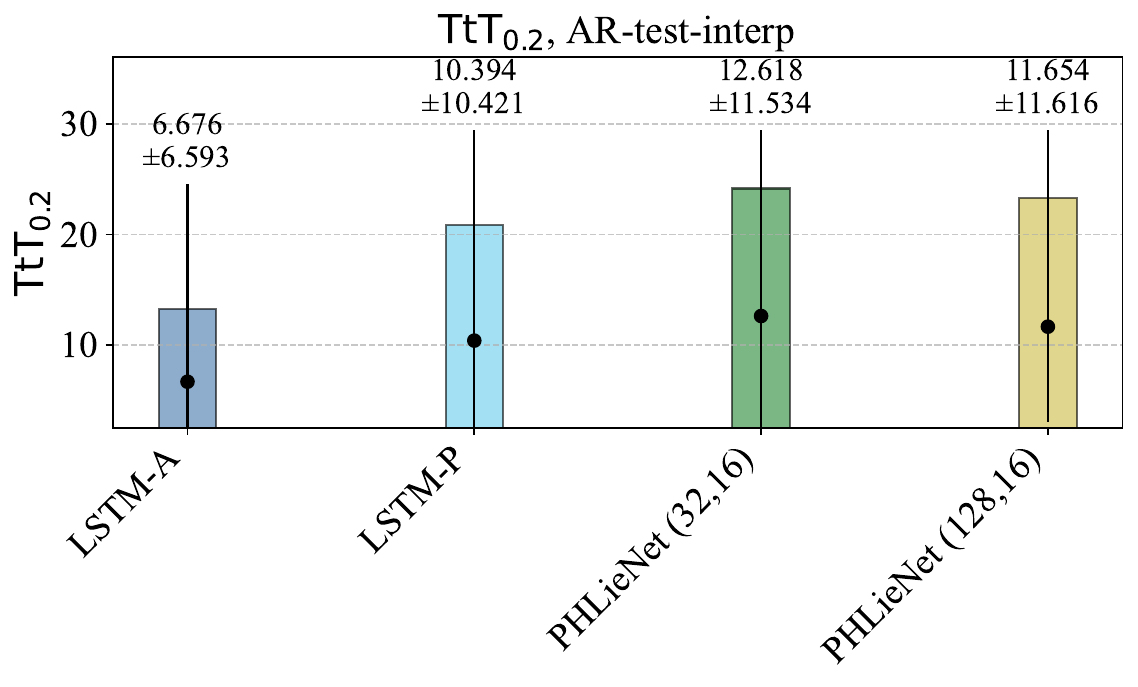}
        \caption{Time-to-Threshold ($\ttt_{0.2}$) metric.}
        \label{fig:lorenz:interp:ttt}
    \end{subfigure}
    \quad
    \begin{subfigure}{0.45\textwidth}
        \centering
        \includegraphics[width=\textwidth]{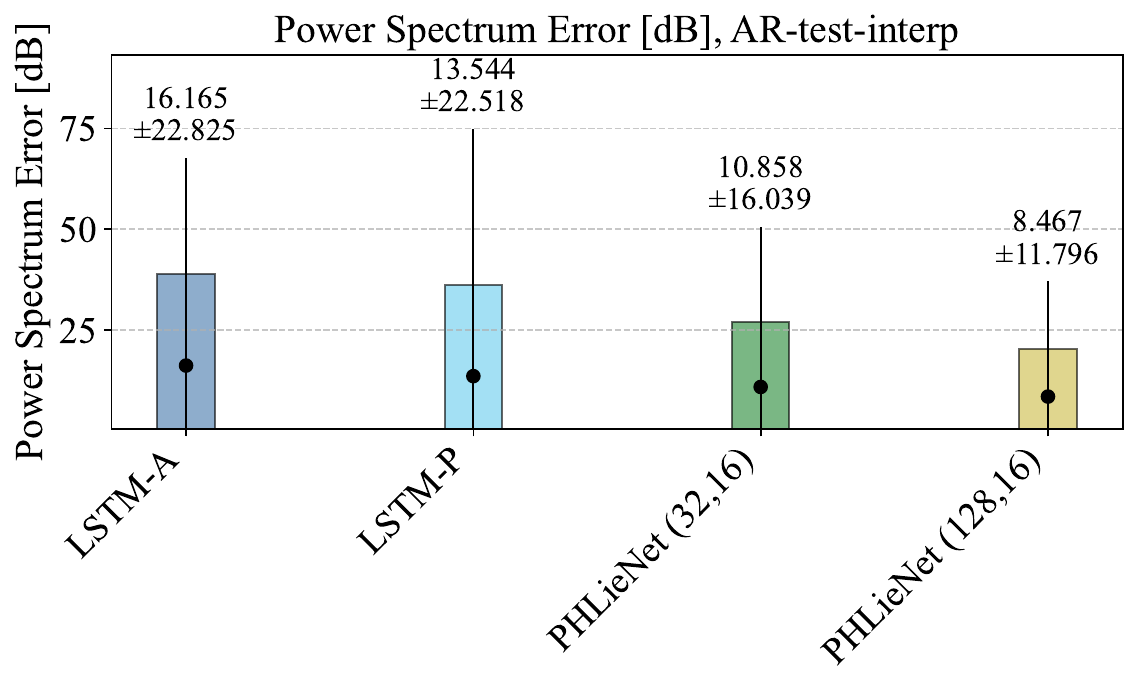}
        \caption{Power spectrum error.}
        \label{fig:lorenz:interp:psd_error}
    \end{subfigure}
    \caption{
        Model performance on the Lorenz 3D system for the interpolation task.
        (a) $\ttt_{0.2}$ per parameter value, averaged over initial conditions, with mean $\pm$ std across the ten $\rho$ values.
        (b) Power spectrum error.
    }
    \label{fig:lorenz:interp}
\end{figure}

\begin{figure}[!hbt]
    \centering
    \includegraphics[width=0.75\textwidth]{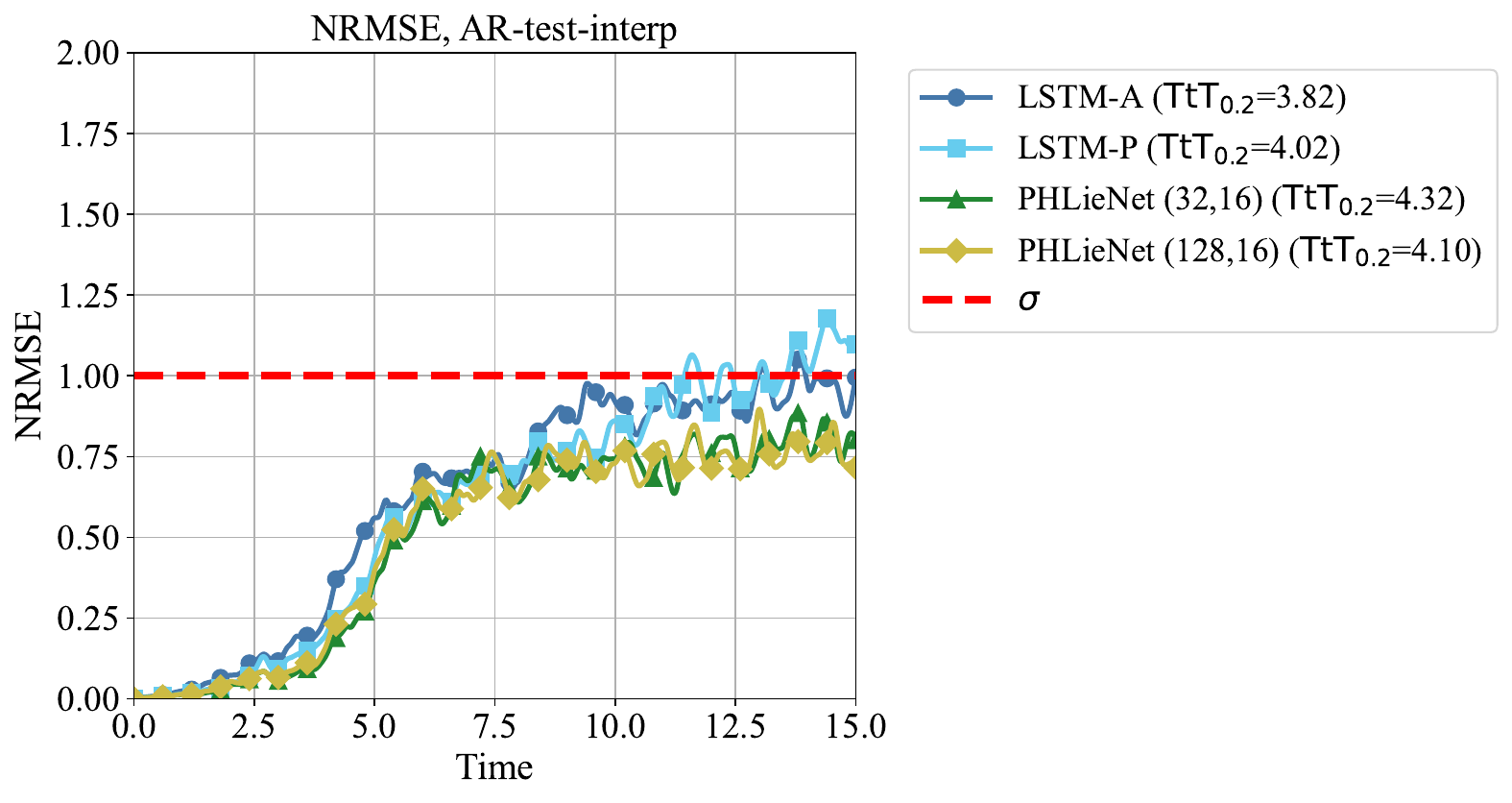}
    \caption{NRMSE evolution in time for the interpolation task, averaged over all $\rho$ values and initial conditions.}
    \label{fig:lorenz:interp:rmse_evolution}
\end{figure}

\begin{figure}[!hbt]
    \centering
    \includegraphics[width=\textwidth]{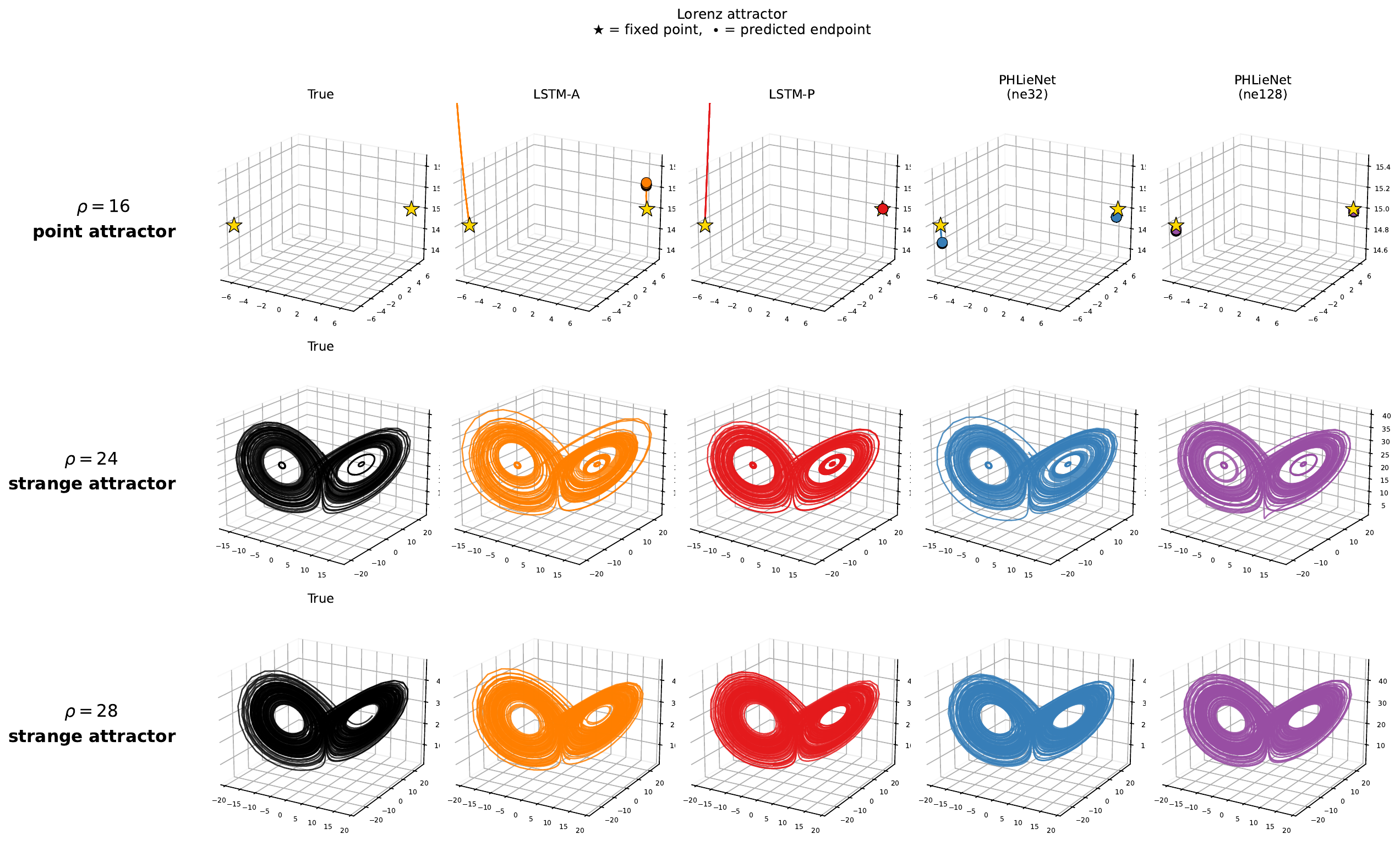}
    \caption{
        Predicted vs.\ true 3D Lorenz attractors for three representative interpolation parameters, showing 20 initial conditions per subplot.
        Top row ($\rho=16$, point attractor): true trajectories converge to $\mathbf{C}^\pm$ ($\bigstar$), with predicted endpoints (colored $\bullet$) landing nearby.
        Middle and bottom rows ($\rho=24$, $\rho=28$, strange attractor): trajectories explore the Lorenz butterfly.
        Axis limits are shared within each row.
    }
    \label{fig:lorenz:attractors}
\end{figure}

Quantitative results are shown in~\Cref{fig:lorenz:interp:rmse_evolution,fig:lorenz:interp}.
PHLieNet\textsubscript{(32,16)} achieves the highest mean $\ttt_{0.2}$ of 12.6 across all ten $\rho$ values, outperforming LSTM-P (10.4) by $+$21\% and LSTM-A (6.7) by $+$88\%.
PHLieNet\textsubscript{(128,16)} achieves a comparable mean $\ttt_{0.2}$ of 11.7.
Both PHLieNet variants demonstrate strong performance across all dynamical regimes, successfully capturing fixed-point convergence and chaotic attractor dynamics where the baselines diverge. PHLieNet\textsubscript{(128,16)} reaches the maximum observable $\ttt_{0.2}$ across all regimes, indicating near-perfect trajectory tracking.
In the chaotic regime ($\rho > \rho_c$), all models degrade to short $\ttt_{0.2}$ values of $2.5$--$4.2$ time units, reflecting the fundamental predictability limit imposed by the positive Lyapunov exponent of chaotic trajectories. The PHLieNet advantage narrows accordingly.
The NRMSE evolution in~\Cref{fig:lorenz:interp:rmse_evolution} confirms these trends: PHLieNet maintains lower mean error over the roll-out horizon, with the gap most visible in the early portion of the trajectory before Lyapunov divergence dominates.
The power spectrum error in~\Cref{fig:lorenz:interp:psd_error} shows PHLieNet also better captures the spectral structure of the dynamics across all regimes.

\Cref{fig:lorenz:attractors} shows the predicted attractor structure for three representative values of $\rho$.
At $\rho=16$, true trajectories converge to one of the two stable fixed points $\mathbf{C}^\pm$ (marked with $\bigstar$).
PHLieNet correctly reproduces this convergent behavior, with predicted endpoints (circles $\bullet$) landing near the analytical fixed points, while the LSTM baselines fail to converge.
At $\rho=24$ and $\rho=28$, the system is in the strange attractor regime. PHLieNet reproduces the correct attractor geometry, whereas the baselines diverge from the true attractor structure.
This shows that a single PHLieNet model, with sufficient embedding size, can represent both fixed-point and chaotic dynamics across the bifurcation.

\subsection{Embedding Space Analysis}
\label{sec:embedding_analysis}

To understand \emph{why} PHLieNet generalizes to unseen parameters, we analyze the structure of the learned embedding space for PHLieNet\textsubscript{(32,16)} across three benchmark systems: the Van der Pol oscillator discussed in~\Cref{sec:vdp}, the R\"{o}ssler system discussed in~\Cref{sec:rossler}, and the Finance system from~\Cref{sec:finance}.
For each system we present: (i) a Principal Component Analysis (PCA) projection of the interpolated query embeddings colored by the parameter value, with anchor positions shown as markers and representative attractor insets; (ii) the RBF activation profile as a function of the parameter; and (iii) the pairwise L2 distance matrix between the generated target network weights across parameter values.

\begin{figure}[!hbt]
    \centering
    \includegraphics[width=0.90\textwidth]{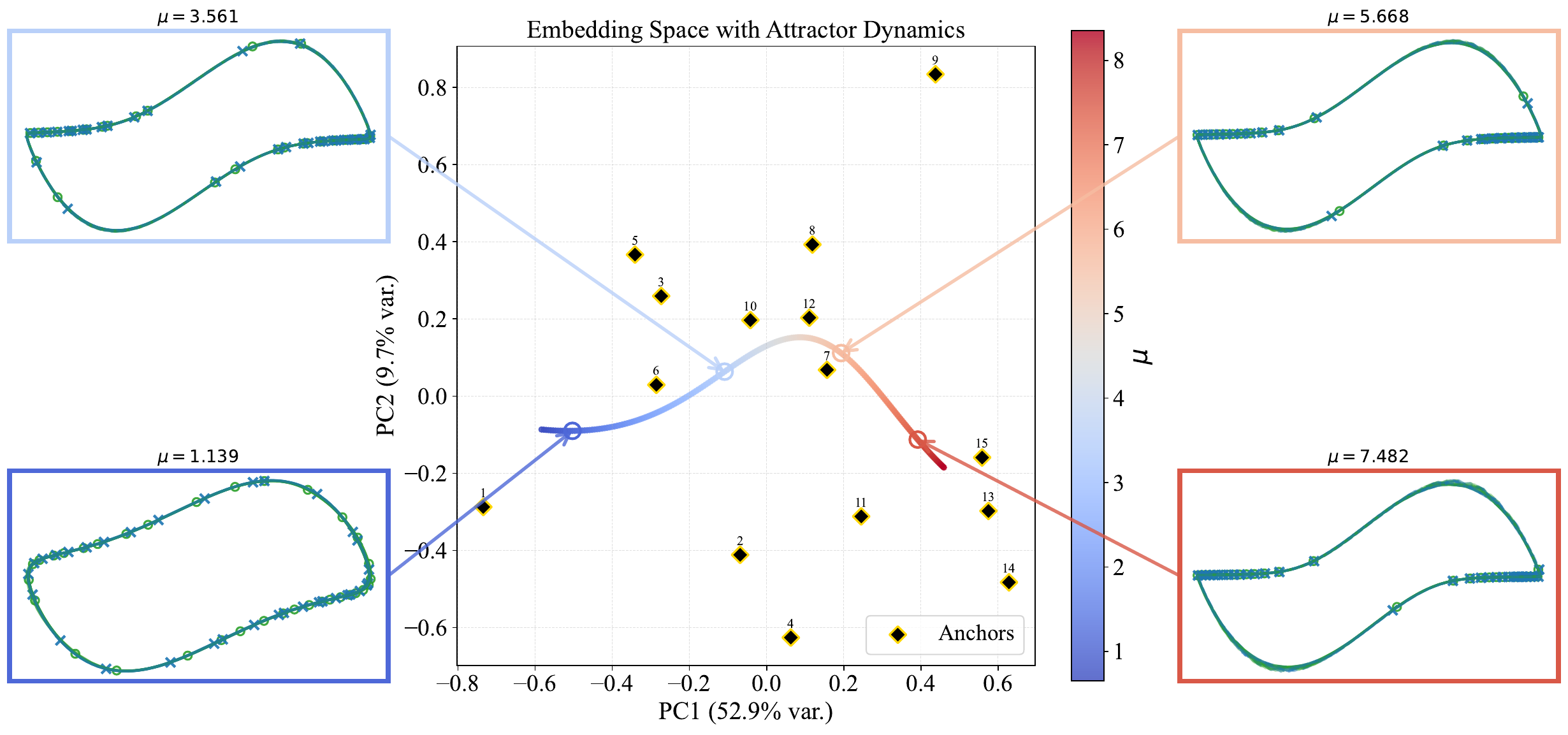}
    \caption{ Embedding space analysis for the Van der Pol oscillator: PCA projection of the learned anchor embeddings colored by $\mu$, with attractor insets at representative parameter values.}
    \label{fig:embedding:vdp:pca}
\end{figure}

\begin{figure}[!hbt]
    \begin{subfigure}{0.52\textwidth}
        \centering
        \includegraphics[width=\textwidth]{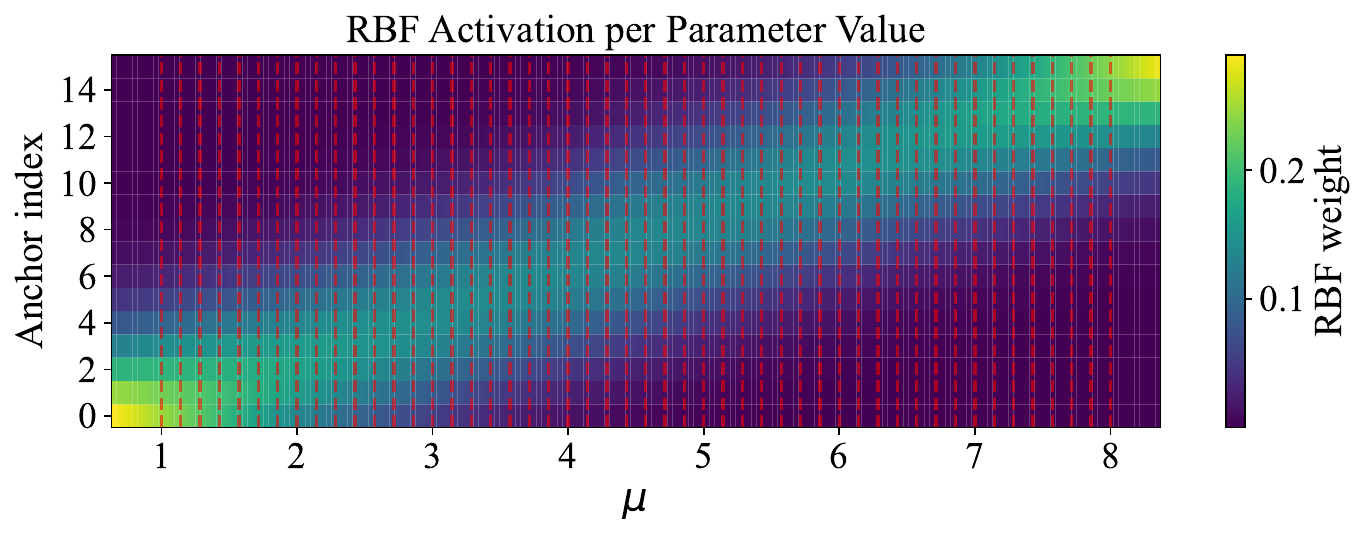}
        \caption{RBF activation profile vs.\ $\mu$.}
        \label{fig:embedding:vdp:rbf}
    \end{subfigure}
    \hfill
    \begin{subfigure}{0.32\textwidth}
        \centering
        \includegraphics[width=\textwidth]{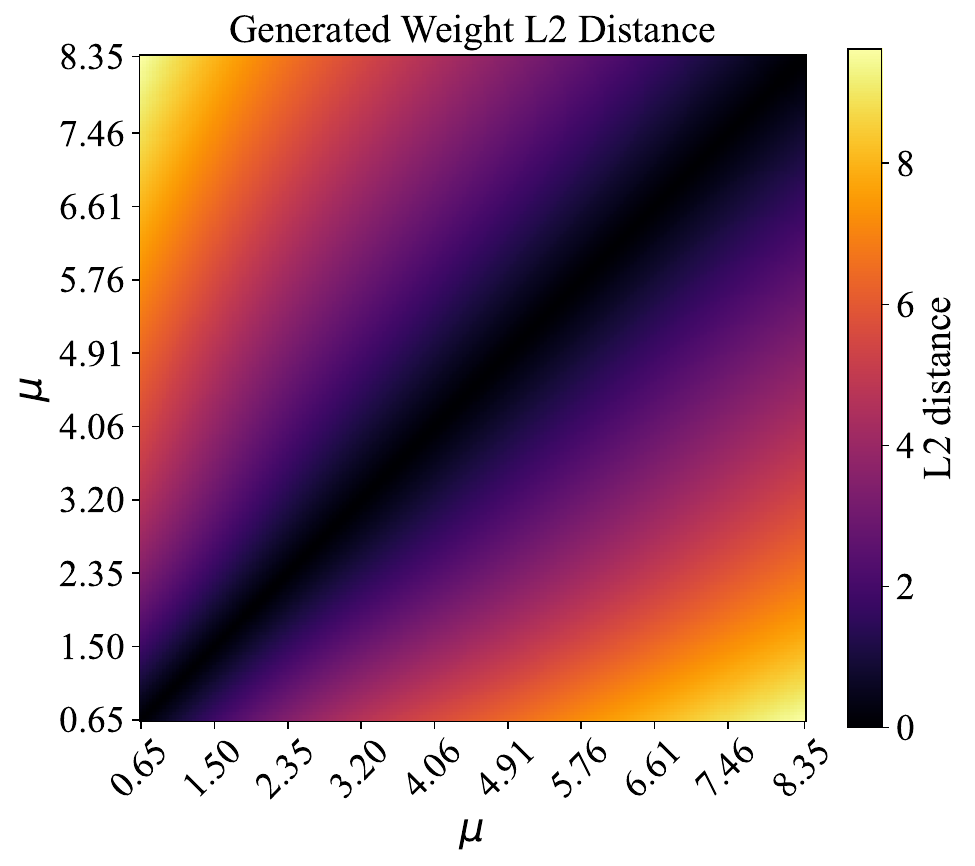}
        \caption{Pairwise weight distance matrix.}
        \label{fig:embedding:vdp:wdist}
    \end{subfigure}
    \caption{
        Embedding space analysis for the Van der Pol oscillator.
        \textit{Left}: RBF activation profile, with anchors activating in smooth, overlapping bands.
        \textit{Right}: Pairwise L2 distance matrix between generated weights, with near-diagonal structure confirming smooth weight variation with $\mu$.
    }\label{fig:embedding:vdp}
\end{figure}

The embedding space analysis for the Van der Pol oscillator is summarized in Figures~\ref{fig:embedding:vdp:pca} and~\ref{fig:embedding:vdp}.
The embeddings learned are organized in a nearly one-dimensional manifold in a monotonic order by system parameter. 
This structured learning was achieved without external supervision or constrained learning techniques. 
Furthermore, the activation profiles of the RBF in~\Cref{fig:embedding:vdp:rbf} demonstrate smooth and overlapping bands of anchor activation, which is crucial for continuous interpolation across the model space.
In addition, the pairwise weight distance matrices in~\Cref{fig:embedding:vdp:wdist} exhibit a near-diagonal structure.

\begin{figure}[!hbt]
    \centering
     \includegraphics[width=0.90\textwidth]{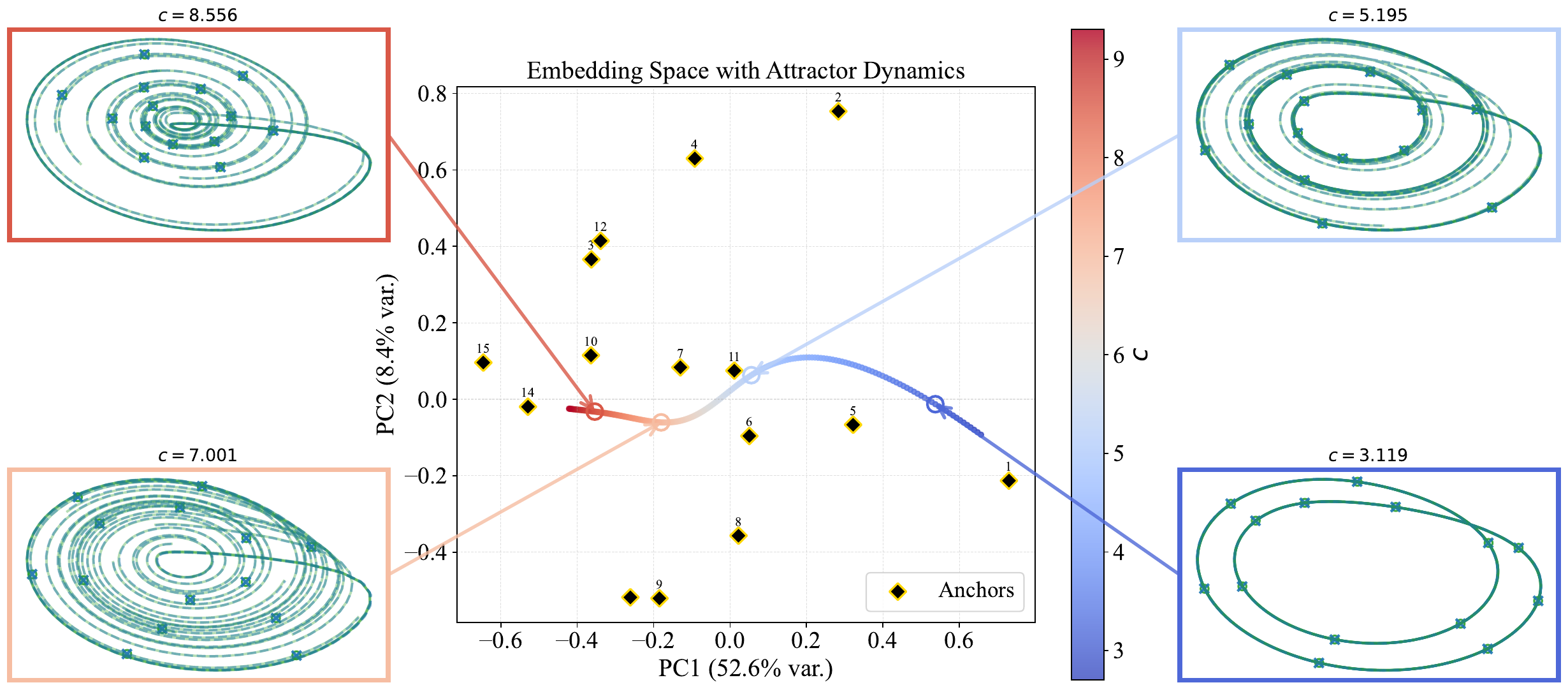}
        \caption{Embedding space analysis for the R\"{o}ssler system: PCA projection of the learned anchor embeddings colored by $c$, with attractor insets at representative parameter values.}
        \label{fig:embedding:roessler:pca}
\end{figure}

\begin{figure}[!hbt]
    \begin{subfigure}{0.54\textwidth}
        \centering
        \includegraphics[width=\textwidth]{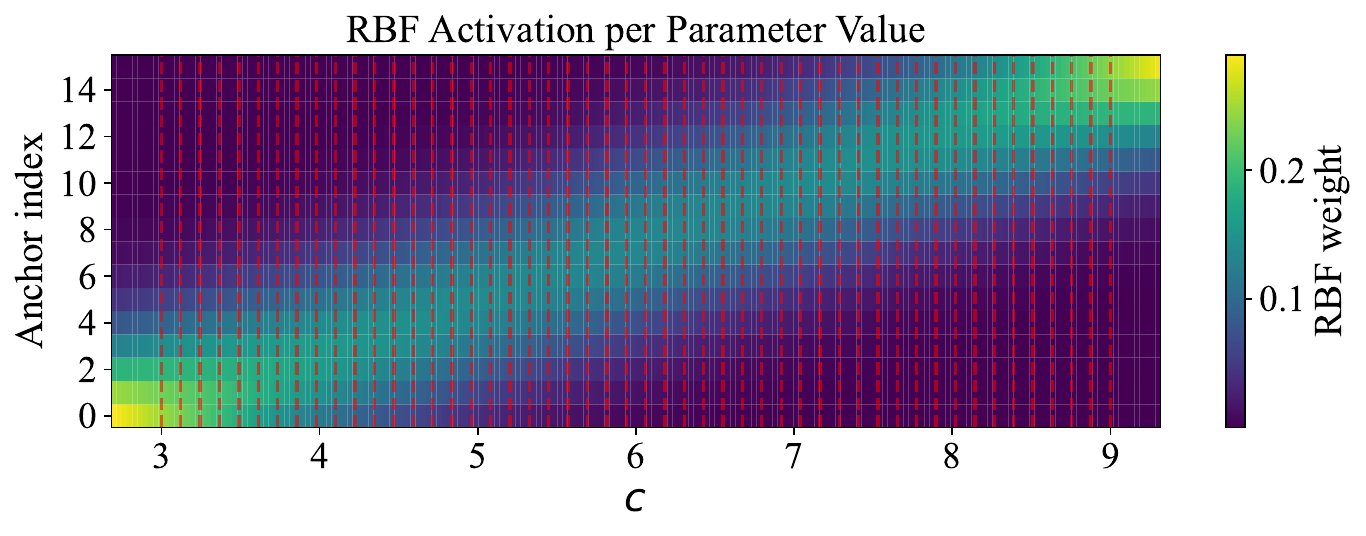}
        \caption{RBF activation profile vs.\ $c$.}
        \label{fig:embedding:roessler:rbf}
    \end{subfigure}
    \hfill
    \begin{subfigure}{0.34\textwidth}
        \centering
        \includegraphics[width=\textwidth]{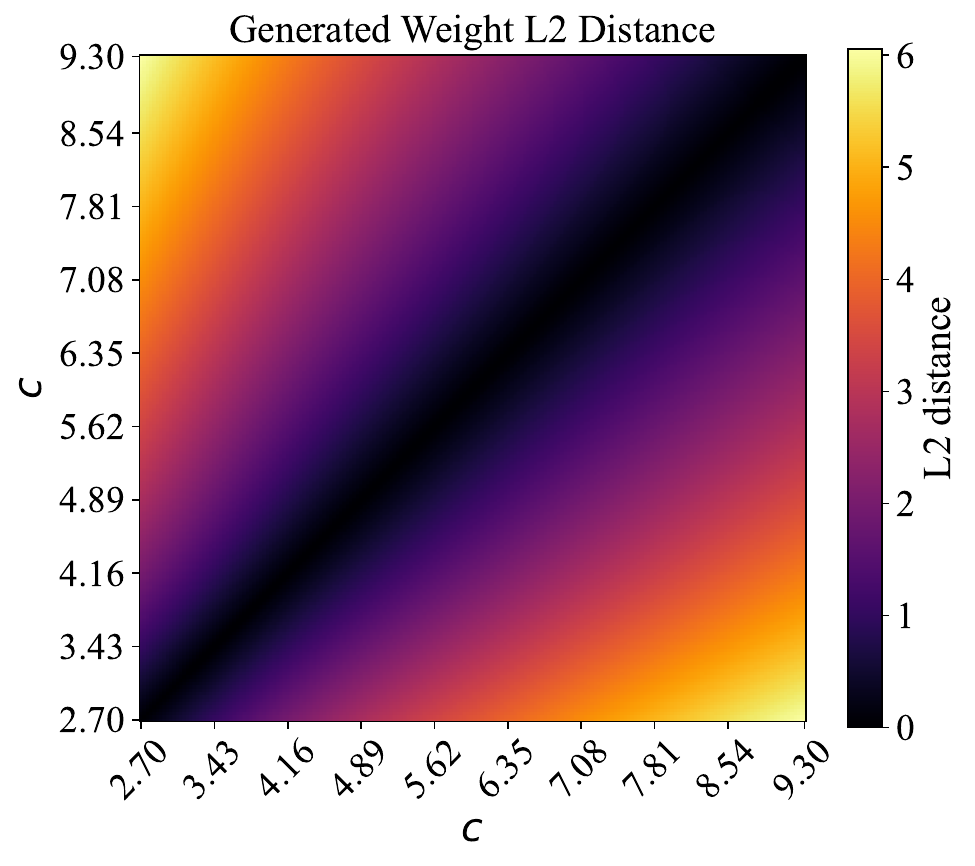}
        \caption{Pairwise weight distance matrix.}
        \label{fig:embedding:roessler:wdist}
    \end{subfigure}

    \caption{
        Embedding space analysis for the R\"{o}ssler system.
        \textit{Left}: RBF activation profile, with anchors activating in smooth, overlapping bands.
        \textit{Right}: Pairwise L2 distance matrix between generated weights, with near-diagonal structure confirming smooth weight variation with $c$.
    }
    \label{fig:embedding:roessler}
\end{figure}

\begin{figure}[!hbt]
    \centering
    \includegraphics[width=0.90\textwidth]{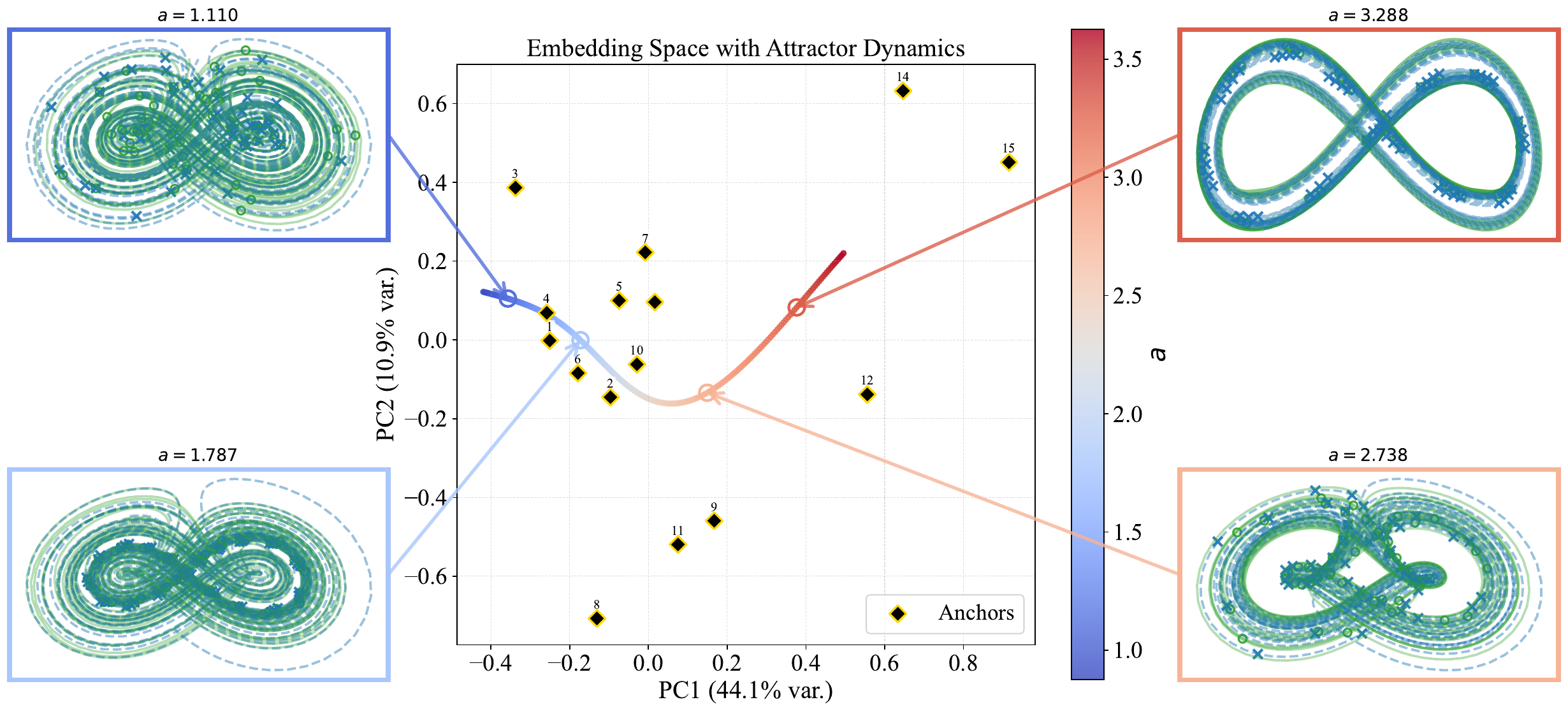}
        \caption{Embedding space analysis for the Finance system: PCA projection of the learned anchor embeddings colored by $a$, with attractor insets at representative parameter values.}
        \label{fig:embedding:finance:pca}
\end{figure}

\begin{figure}[!hbt]
    \begin{subfigure}{0.57\textwidth}
        \centering
        \includegraphics[width=\textwidth]{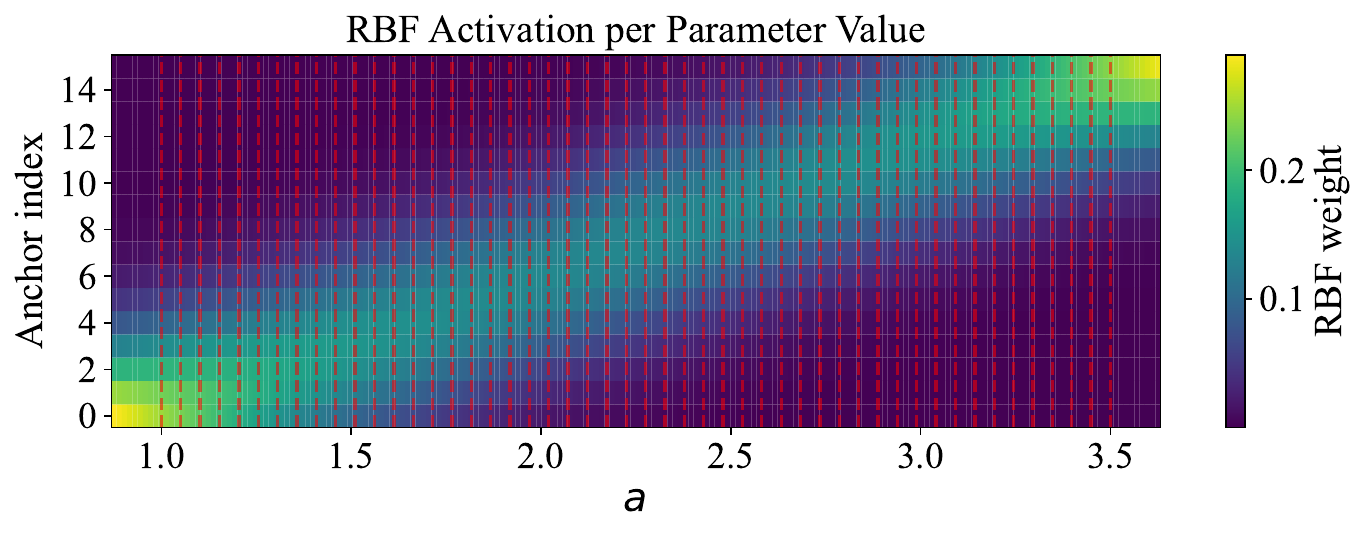}
        \caption{RBF activation profile vs.\ $a$.}
        \label{fig:embedding:finance:rbf}
    \end{subfigure}
    \hfill
    \begin{subfigure}{0.38\textwidth}
        \centering
        \includegraphics[width=\textwidth]{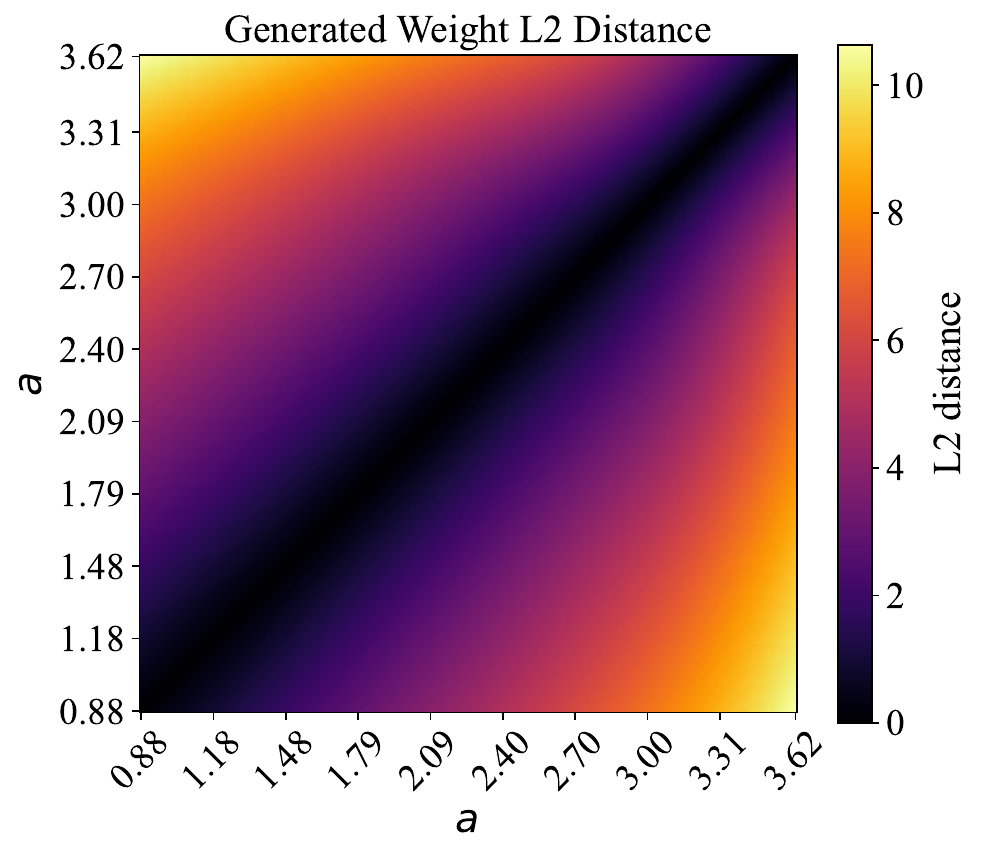}
        \caption{Pairwise weight distance matrix.}
        \label{fig:embedding:finance:wdist}
    \end{subfigure}

    \caption{
        Embedding space analysis for the Finance system.
        \textit{Left}: RBF activation profile, with anchors activating in smooth, overlapping bands.
        \textit{Right}: Pairwise L2 distance matrix between generated weights, with near-diagonal structure confirming smooth weight variation with $a$.
    }
    \label{fig:embedding:finance}
\end{figure}

Similar observations can be made for the R\"{o}ssler system, which is summarized in Figures~\ref{fig:embedding:roessler:pca} and~\ref{fig:embedding:roessler}, and the Finance system reported in Figures~\ref{fig:embedding:finance:pca} and~\ref{fig:embedding:finance}.
In all three cases, both the PCA projection plots and the visualizations of the RBF activation profile and the distance matrices exhibit similar patterns.

Thus, a consistent pattern emerges across all three systems.
The learned embeddings organize on a nearly one-dimensional manifold that is monotonically ordered by the system parameter, without any explicit supervision on that ordering.
This demonstrates that PHLieNet discovers the intrinsic geometry of the parameter axis and encodes it in a structured latent representation.
The RBF activation profiles confirm smooth, overlapping bands of anchor activation, implementing continuous interpolation across the parameter space.
The pairwise weight distance matrices exhibit a near-diagonal structure, confirming that the generated target network weights vary continuously and monotonically with the parameter.
This geometric regularity is the mechanism underlying PHLieNet's generalization advantage: interpolating in weight space between structured, parameter-ordered embeddings yields a well-posed dynamics model for any parameter value, including those outside the training range.
This consistent structure across all three systems demonstrates that PHLieNet learns a principled, parameter-ordered latent representation, providing the geometric basis for its generalization advantage.

\subsection{Summary}
\label{sec:summary}

\Cref{tab:summary} consolidates the $\ttt_{0.2}$ and power spectrum error (PSE) results across all benchmark systems, comparing the best PHLieNet variant against LSTM-P, the strongest parametric baseline.
On the interpolation task, PHLieNet outperforms LSTM-P across all systems, with $\ttt_{0.2}$ gains ranging from $+2\%$ to $+110\%$, and attains lower or matched PSE in all cases.
On the extrapolation task, it achieves large margins on two out of three evaluated systems, while LSTM-P retains an advantage on one.

\begin{table}[!hbt]
    \caption{
        Cross-system summary of $\ttt_{0.2}$ (higher is better) and power spectrum
        error (PSE, lower is better), comparing the best PHLieNet variant against LSTM-P,
        the strongest parametric baseline.
        $\Delta$ is the relative gain of PHLieNet over LSTM-P.
        Bold indicates the better result in each metric pair.
        $\dagger$~Results reported in \Cref{sec:appendix:chuaduffing}.
    }
    \label{tab:summary}
    \centering
    \small
    \begin{tabular}{l l r r r r r}
        \hline
        & & \multicolumn{2}{c}{$\ttt_{0.2}$} & & \multicolumn{2}{c}{PSE} \\
        \cline{3-4} \cline{6-7}
        System & Variant & PHLieNet & LSTM-P & $\Delta\ttt_{0.2}$ & PHLieNet & LSTM-P \\
        \hline
        \noalign{\vskip 2pt}
        \multicolumn{7}{l}{\textit{Interpolation}} \\
        \noalign{\vskip 2pt}
        Van der Pol             & $(32,16)$ & $\mathbf{22.0}$ & $10.5$          & $+110\%$        & $\mathbf{1.21}$ & $2.96$          \\
        Rössler                 & $(16,32)$ & $\mathbf{60.2}$ & $50.5$          & $+19\%$         & $\mathbf{1.10}$ & $1.58$          \\
        Finance                 & $(16,32)$ & $\mathbf{45.0}$ & $42.1$          & $+7\%$          & $\mathbf{1.14}$ & $1.28$          \\
        Lorenz 3D               & $(32,16)$  & $\mathbf{12.6}$ & $10.4$          & $+21\%$         & $\mathbf{10.86}$ & $13.54$        \\
        Chua$^{\dagger}$        & $(32,64)$ & $11.1$                & $10.9$   & $+2\%$          & $2.14$          & $2.15$          \\
        Duffing$^{\dagger}$     & $(16,32)$ & $\mathbf{24.2}$ & $22.0$          & $+10\%$         & $\mathbf{1.19}$ & $1.40$          \\
        \noalign{\vskip 4pt}
        \hline
        \noalign{\vskip 2pt}
        \multicolumn{7}{l}{\textit{Extrapolation}} \\
        \noalign{\vskip 2pt}
        Van der Pol             & $(32,16)$ & $\mathbf{4.61}$ & $1.78$          & $+159\%$        & $\mathbf{3.34}$ & $8.29$          \\
        Rössler                 & $(32,16)$ & $29.3$          & $\mathbf{36.1}$ & $-19\%$         & $1.26$          & $\mathbf{1.13}$ \\
        Finance                 & $(32,64)$ & $\mathbf{30.5}$ & $25.9$          & $+18\%$         & $\mathbf{1.67}$ & $3.73$          \\
        \hline
    \end{tabular}
\end{table}

\section{Discussion}
\label{sec:discussion}

Modeling dynamical systems that exhibit differentiated responses to varying external stimuli or parameters is a central challenge with broad practical relevance.
Previous approaches often fail to capture the full spectrum of behaviors within a unified framework.
To address this, we propose PHLieNet - Parametric Hypernetwork for Learning Interpolated Networks, a novel architecture which a) learns a continuous embedding of the parametric space and b) uses hypernetworks to map this embedding to the parameters of a latent dynamics network.
Unlike existing methods, PHLieNet does not impose theoretical constraints on the class of dynamics that can be represented.
The hypernetwork decouples parametric adaptation of the target network from temporal modeling of the state space dynamics, allowing the model to generate diverse dynamic behaviors.

We assess the performance of PHLieNet on five complex parametric dynamical systems that exhibit nonlinear dynamics: the Van der Pol oscillator, the Lorenz system, the Rössler attractor, the Finance system, and Chua's circuit (appendix).
We benchmark PHLieNet against other state-of-the-art methods, including parameter-agnostic models based on feedforward neural networks, LSTMs, and TCNN-CDs, as well as their adaptations that augment the hidden state with the parameter.
Evaluation metrics span both short-term prediction performance, such as time-to-threshold and NRMSE evolution, and the ability to reproduce the long-term spectral structure, as measured by the power spectrum error.
Across all benchmarks, PHLieNet consistently outperforms or matches the performance of other models while also exhibiting qualitatively distinct properties compared to existing methods.
We attribute this advantage to the weight-space interpolation mechanism: by generating distinct network weights per parameter value, PHLieNet adapts its full computational graph to each dynamical regime, rather than relying on a single shared set of weights to partition capacity across qualitatively different behaviors.

Although the use of hypernetworks for parametric dynamics modeling is still an emerging research area, the results in this work suggest that it is a direction worth pursuing.
Future research could extend the proposed method to high-dimensional parametric spaces, where multiple interacting parameters govern the system’s behavior.
Such an extension is conceptually straightforward within our framework and could be achieved through barycentric interpolation in the parameter space, enabling even more flexible and expressive modeling capabilities.

Another promising direction is to eliminate the interpolation step altogether and directly learn the mapping from parameters to embedding.
In an early version of our work, we explored this using a simple linear network.
We obtained promising preliminary results, although for a more limited range of dynamics compared to this work.
A similar linear mapping approach was also used in~\cite{brenner2024learning}, which also demonstrated limited generalization in various dynamical regimes.

However, PHLieNet's performance comes with an important caveat: it rests on the assumption that the dynamics vary smoothly with the parameters.
Specifically, PHLieNet relies on smooth activation functions to perform interpolation over parametric dynamics and weight space.
In scenarios involving abrupt transitions, such as bifurcations, shocks, or regime shifts, our framework would require adaptations to capture these discontinuities accurately.
This limitation is evident in the Lorenz system, where extrapolation to parameter values above the training range ($\rho \in \{36, \ldots, 40\}$) causes all models to fail, likely because the higher-$\rho$ attractors are substantially more complex than those encountered during training, making smooth weight-space interpolation insufficient.
Furthermore, in this work we used the same set of parameters for training and validation.
However, we also tested the ability of PHLieNet to generalize to unseen parametric dynamics.
Generalization to unseen parameters could be further improved by evaluating whether using a distinct parameter set for validation would enhance generalization.
A comprehensive analysis of how to select the validation set is left for future work.

The proposed framework poses no limitations on the structure or selection of the target network.
The target network in our framework could be a Neural ODE, a neural operator, a PINN, or any other dynamics propagator. 
Improving the proposed approach and benchmarking these different types of target network in PHLieNet while gaining a deeper understanding of their differences remain an open area of investigation.
It remains unclear which parametric weight space is more amenable to interpolation, as these methods exhibit distinct characteristics.
In our experiments, we observed that interpolating in the parametric space of temporal CNNs was easier than with RNNs such as LSTMs.

Another interesting avenue for future research is to apply PHLieNet to online adaptive modeling of dynamical systems.
Rather than training the model offline, this approach would account for parameters that change in real-time.
Such a framework would require online anomaly detection, the ability to detect when the dynamic regime has shifted, and automatic recalibration of the model.
Achieving this would likely involve combining PHLieNet with a state estimation framework.

All in all, PHLieNet marks a departure from the reductionist practice of training distinct models for each parameter setting to a more holistic and unified approach that learns the interplay between parameter and system dynamics.
This framework not only enables interpolation across a wide range of parameter regimes but also opens the door to adaptive and generalizable modeling of complex dynamical systems.
We believe that such a perspective, one that embraces the structure of the full parametric space, can inspire further research toward more flexible, robust, and insightful models of the dynamic phenomena that govern real-world systems.

\section*{Acknowledgements}

KV and EC gratefully acknowledge the funding from the European Commission under the Horizon Europe funding guaranty, for the projects ‘ReCharged - Climate-aware Resilience for Sustainable Critical and interdependent Infrastructure Systems enhanced by emerging Digital Technologies’ (grant agreement No: 101086413) and ‘TURING - Trustworthy Unified Robust Intelligent Generative Systems’ (grant agreement No: 101215032).
The authors further acknowledge the support provided by the High Performance Computing group at ETH Zurich operating the Euler cluster that was used for all numerical simulations.

\section{Declaration of generative AI and AI-assisted technologies in the writing process}

During the preparation of this work, the authors used ChatGPT 4o to improve the flow and clarity of the text, and Claude Sonnet (Anthropic) to assist with code development and figure preparation.
After using these tools, the authors reviewed and edited the content as needed and assume full responsibility for the content of the publication.

\appendix
\section{Hypernetwork Architecture Details}
\label{sec:appendix:hypernetwork}

We now describe the functional form of the hypernetwork used to generate the parameters of the target forecasting network, as defined in~\Cref{eq:hnn1}.

Let the target network parameters be denoted by \( w_f \), which includes all weights and biases.
By flattening and concatenating these, we obtain a vector of total length \( |w_f| \).
The goal of the hypernetwork is to generate \( w_f \) as a function of the system parameter vector \( \param \in \RR^{D_p} \).

\subsection{Embedding Interpolation Mechanism}

To enable smooth generalization across parameter space, we employ a radial basis function (RBF) interpolation mechanism over a fixed number of learned embedding vectors.
Specifically, we define a set of \( \nE \) learnable anchor embeddings \( \{ \embedding^{(i)} \}_{i=1}^{\nE} \subset \RR^{\dimE} \), implemented as a standard embedding layer.
The anchor positions \( \{ \param^{(i)} \}_{i=1}^{\nE} \) are placed uniformly in \( [0,1] \), with \( \param^{(i)} = (i-1)/(\nE - 1) \).
The input parameter is normalized to this range prior to embedding.

For a given input parameter \( p \in [0,1] \), we compute the interpolation weights over \emph{all} anchor positions using a Gaussian RBF kernel followed by softmax normalization:
\[
\alpha_i(p) = \frac{\exp\!\Big(-\frac{(p - \param^{(i)})^2}{2\sigma^2}\Big)}{\sum_{k=1}^{\nE} \exp\!\Big(-\frac{(p - \param^{(k)})^2}{2\sigma^2}\Big)},
\]
where \( \sigma > 0 \) is the bandwidth parameter (set to \( \sigma = 0.2 \) in our experiments).
The resulting embedding is a convex combination of all anchor embeddings:
\[
\embedding(p) = \sum_{i=1}^{\nE} \alpha_i(p) \, \embedding^{(i)} \in \RR^{\dimE}.
\]
This mechanism ensures that the embedding varies smoothly and differentiably with \( p \), with all anchors contributing to every query point, their influence decaying with distance in parameter space according to the Gaussian kernel.

Our method is inherently scalable to high-dimensional parameter vectors (\( D_p \gg 1 \)), with no architectural limitations.
The RBF kernel operates on the norm of the parameter vector and generalizes naturally to higher dimensions.
For very high-dimensional parameter spaces, alternatives such as barycentric interpolation over sparse simplexes could also be employed.

\subsection{Weight Generation via MLP}

The embedding is then passed through a multi-layer perceptron (MLP) to generate the flattened weight vector of the target network:
\begin{equation}
w_f = \mathrm{MLP}\left( \embedding(p) \right) \in \RR^{|w_f|},
\end{equation}
where the MLP is composed of several hidden layers with activation functions such as SiLU and a final output layer of size \( |w_f| \).
This architecture allows efficient and expressive mapping from the embedding space to the parameter space of the target network.

\subsection{Training}

The entire system consisting of the embedding layer, the MLP-based hypernetwork, and the target temporal forecasting model, is trained end-to-end using gradient-based optimization.
The hypernetwork remains fully differentiable with respect to the input parameter \( \param \), allowing backpropagation through both the embedding interpolation and the weight generation process.

This architecture enables dynamic generation of forecasting models tailored to each parameter configuration without requiring retraining for every new value of \( \param \). As such, it provides a flexible and efficient approach for modeling dynamical systems across wide parametric domains.
\section{Learnability of Smooth Parametric Families via Hypernetworks}
\label{sec:appendix:learnability}

In this appendix, we provide a formal justification in the form of a proof sketch for the weight-space interpolation adopted by PHLieNet.
We show that, under smoothness of the dynamics with respect to both the state and the parameter vector, the hypernetwork architecture can approximate a parametric family of dynamics to arbitrary precision.
Consider a parametric family of dynamical systems
\begin{equation}
\statedt = f(\state, \param), \quad \param \in \mathcal{P} \subset \RR^{\dimParam},
\end{equation}
where $\mathcal{P}$ is a compact parameter domain.
We denote by $\{f^{w} : w \in \mathcal{W} \subseteq \RR^{|w|}\}$ the class of functions realizable by the target network (e.g., a TCNN-CD), and by $\loss(w, \param)$ the expected approximation error of the target network $f^{w}$ with respect to the true dynamics $f(\state, \param)$ for a given parameter value $\param$:
\begin{equation}
\loss(w, \param) = \mathbb{E}_{
\state \sim \mu_{\mathbf{p}}
}\left[\left\| f^{w}(\state) - f(\state, \param) \right\|^2\right],
\end{equation}
where $\mu_{\param}$ is the state space distribution induced by the dynamics (e.g. the invariant measure of the attractor).
We assume that $\mu_{\param}$ depends smoothly on $\param$, which holds under assumption~(i) below in the absence of bifurcations.
The approximation error $\loss(w, \param)$ refers to the theoretical approximation quality averaged over the true state space distribution of the dynamics under a specific parametric regime, as opposed to the empirical loss computed over a finite training set.

\begin{proposition}[Learnability of smooth parametric families]
\label{prop:learnability}
Assume the following conditions hold:
\begin{enumerate}[label=(\roman*)]
\item \textbf{Smoothness of the dynamics.}
The map $(\state, \param) \mapsto f(\state, \param)$ is jointly smooth ($C^\infty$) on $\RR^{\dimState} \times \mathcal{P}$,
and the dynamics do not undergo bifurcations as $\param$ varies over $\mathcal{P}$.
Specifically, we assume that the qualitative structure of the attractor and the invariant measure $\mu_{\param}$ vary smoothly with $\param$.

\item \textbf{Existence and approximation capacity of the target network~\cite{hornik1991approximation}.}
For every $\param \in \mathcal{P}$, the loss $\loss(\cdot,\param)$ admits at least one local minimizer $w^*(\param) \in \mathcal{W}$, and for every $\varepsilon > 0$ there exists at least one such local minimizer satisfying $\loss(w^*(\param), \param) < \varepsilon$.

\item \textbf{Non-degeneracy of the loss landscape.}
For each $\param \in \mathcal{P}$, the minimizer $w^*(\param)$ satisfies $\nabla_w \loss(w^*(\param), \param) = 0$, and the
Hessian $\nabla^2_{ww} \loss(w^*(\param), \param)$ is positive definite after accounting for weight-space symmetries~\cite{brea2019weight,simsek2021geometry} (e.g., neuron permutations and scaling symmetries).

\item \textbf{Universal approximation of the hypernetwork.}
The hypernetwork MLP class is dense in $C(\mathcal{P}, \mathcal{W})$ (the space of all continuous functions from $\mathcal{P}$ to $\mathcal{W}$), i.e., it can approximate any continuous mapping from
the parameter space to the weight space to arbitrary precision.
\end{enumerate}
Then, for every $\varepsilon > 0$, there exists a hypernetwork $\hnn$ such that
\begin{equation}
\sup_{\param \in \mathcal{P}} \;
\loss \, \!\Big(\hnn\!\big(\embedding(\param); \, w_H\big), \, \param\Big) < \varepsilon,\end{equation}
where $\embedding(\param)$ is the learned interpolated embedding defined in~\Cref{eq:embedding}.
\end{proposition}

\begin{proof}[Proof sketch]
The argument proceeds in three steps.

\textit{Step 1 (Existence).}
By assumption (i), the dynamics $f(\cdot, \param)$ are smooth and hence continuous for all $\param \in \mathcal{P}$.
By assumption~(ii), for each $\param \in \mathcal{P}$ there exists a local minimizer $w^*(\param) \in \mathcal{W}$ such that the target network $f^{w^*(\param)}$ approximates $f(\cdot,\param)$ to arbitrary precision.

\textit{Step 2 (Regularity of the minimizer mapping).}
The goal of this step is to show that the mapping from the parameter $\param$ to the corresponding loss minimizer $w^*$ is smooth over the entire parameter domain $\mathcal{P}$, which is sufficient for the hypernetwork to approximate it as a single continuous function.
Consider the first-order optimality condition $\nabla_w \loss(w, \param) = 0$.
The smoothness of the dynamics $f(\cdot, \param)$ with respect to $\param$ (assumption (i)), combined with the smoothness of the target network $f^w$ with respect to its weights $w$ (which holds when smooth activation functions are used) ensures that the mapping $(\param, w) \mapsto \nabla_w \loss(w, \param)$ is smooth in both $w$ and $\param$.
By assumption~(iii), the Hessian $\nabla^2_{ww} \loss(w^*(\param_0), \param_0)$ is non-degenerate at any reference point $\param_0 \in \mathcal{P}$.

The implicit function theorem then guarantees that $\param \mapsto w^*(\param)$ is smooth in a neighborhood of each $\param_0$.
Assumption~(iii) ensures that these local branches are isolated.
Since $\mathcal{P}$ is compact and assumption~(i) excludes bifurcations that would induce degeneracies in the Hessian, a continuous selection of minimizers exists over $\mathcal{P}$, which is smooth almost everywhere and piecewise smooth on $\mathcal{P}$.

\textit{Step 3 (Approximation by the hypernetwork).}
Since $\param \mapsto w^*(\param)$ is continuous on the compact set $\mathcal{P}$,
and piecewise smooth almost everywhere, assumption~(iv) guarantees the existence of a hypernetwork $\hnn(\embedding(\param); w_H)$ that approximates $w^*(\param)$ uniformly over $\mathcal{P}$ to within arbitrary accuracy $\delta$.
By the continuity of the loss $\loss$ in $w$, for sufficiently small $\delta$, the composed approximation satisfies
$\loss(\hnn(\embedding(\param); w_H), \param) < \varepsilon$ uniformly over $\mathcal{P}$.
\end{proof}

\begin{remark}[Smooth activations]
The proof of Step 2 relies on the smoothness of the target network with respect to its weights.
This holds when smooth activations are employed, as is the case in our implementation.
\end{remark}

\begin{remark}[Role of the embedding layer]
The learned interpolated embedding $\embedding(\param)$ defined in~\Cref{eq:embedding} provides a smooth and differentiable mapping from $\mathcal{P}$ to $\RR^{\dimE}$ by construction (convex combination of learned anchors).
This does not affect the theoretical guarantee above, since the hypernetwork MLP composed with the embedding layer retains the universal approximation property.
In practice, the embedding layer provides a structured intermediate representation that facilitates learning, particularly when the mapping $\param \mapsto w^*(\param)$ exhibits nonlinear dependence.
\end{remark}

\begin{remark}[Limitations at bifurcation points]
\label{rem:bifurcations}
Assumption~(i) explicitly excludes bifurcations, i.e., qualitative changes in the dynamics as $\param$ varies (e.g., transitions from periodic orbits to chaos).
At such points, the loss landscape may undergo structural changes (local minima can merge, split, or disappear) violating the non-degeneracy condition~(iii) and breaking the smoothness of $\param \mapsto w^*(\param)$.
In such scenarios, the guarantee of~\Cref{prop:learnability} does not directly apply.
In practice, as discussed in~\Cref{sec:discussion}, denser sampling of the parameter space near bifurcation boundaries and an increased number of anchor embeddings $\nE$ can mitigate this limitation.
\end{remark}
\section{Hyperparameters of Network Architectures Used}
\label{sec:appendix:parameters}

The network architectures implemented for benchmarking in this study are summarized in~\autoref{tab:networks}.
For feedforward neural networks (FFNNs), we use two hidden layers with 40 neurons each.
For LSTMs, a single recurrent layer with 48 hidden units is employed.
The TCNN-CD (Temporal Convolutional Neural Network with Causal Dilations) is configured with a channel size of 22 and a kernel size of 5.
These values were chosen so that all baseline models have a comparable total number of trainable parameters, ensuring a fair comparison.

The target network in PHLieNet is also a TCNN-CD with the same configuration (channel size 22, kernel size 5).  
The hypernetwork of PHLieNet incorporates a Learned Interpolated Embedding (LIE) layer with embedding dimension $d_e$, tuned per system from values of 16, 32, and 64. The specific value used in each experiment is reported in the corresponding results section.
The number of anchor embeddings $n_e$ is tuned per system. Values in the range $[16, 32]$ are used for most systems. For the Lorenz 3D system, we additionally evaluate $n_e = 128$. The specific values used in each experiment are indicated in the corresponding results section.

The weight-generating component of the hypernetwork is a fully connected network with two hidden layers of size 64.
We did not observe substantial changes in performance when varying these hyperparameters within reasonable ranges, suggesting that the core architecture is the primary factor in the model's effectiveness.

An input sequence length of \( \isl = 64 \) was used for the Van der Pol oscillator, R\"ossler system, Finance system, and Chua’s circuit. For the Lorenz 3D system, we used \( \isl = 32 \).

While additional performance gains could potentially be achieved through further hyperparameter optimization, such tuning would come at increased computational cost and is beyond the scope of this work.

All networks were trained using truncated backpropagation through time with a batch size of 256 and 6 parallel data-loading workers.  
Training was run for a maximum of 1000 epochs.  
The initial learning rate was set to \( 10^{-2} \) or \( 10^{-3} \) depending on the system, and reduced by a factor of 0.25 if the validation loss did not improve by a relative margin of \( 10^{-5} \) over 15 consecutive epochs.  
If the validation loss failed to improve for 30 consecutive epochs, early stopping was triggered to prevent overfitting.

All input features were standardized using a feature-wise standard scaler.  
To improve robustness, additive Gaussian noise with standard deviation \( \trainnoise \) was applied during training.  
All models were trained using the Ranger optimizer~\cite{wright2021ranger21}.
\section{Additional benchmarking studies}
\label{sec:appendix:chuaduffing}

\subsection{Chua's Circuit}

Chua's circuit, first introduced by Leon O. Chua in 1983~\cite{chua1993global}, is a nonlinear electronic system built from a simple configuration of linear elements (resistors, capacitors, and inductors) and a nonlinear component known as the Chua diode.
The diode introduces a piecewise-linear characteristic that drives the circuit into chaotic behavior.
The dynamics of the circuit are described by a system of three first-order ordinary differential equations governing the voltage and current, which give rise to a wide range of dynamical phenomena such as periodic orbits, bifurcations, and chaotic attractors.

The equations governing the dynamics of Chua's circuit are given by:
\begin{align}
    \dot{x_1} &= a \big( x_2 - x_1 - h(x_1) \big), \\
    \dot{x_2} &= x_1 - x_2 + x_3, \\
    \dot{x_3} &= -b x_2,
\label{eq:chua}
\end{align}
where $\state=[x_1, x_2, x_3]^T \in \RR^3$ represents normalized voltages and currents, while $a$ and $b$ are parameters related to the circuit's components.
The nonlinearity is introduced by the piecewise-linear function
\begin{equation}
h(x) = \mu_1 x + 0.5(\mu_0 - \mu_1)(|x + 1| - |x - 1|),
\end{equation}
where $\mu_0$ and $\mu_1$ control the slope of the diode's characteristic.
We vary $a$ as it crucially determines the balance between linear and nonlinear dynamics through its coupling of the difference $x_1 - x_2$ to the nonlinearity.
We fix $b=15.0$, $\mu_0=-1.143$, and $\mu_1=-0.714$, following previous studies~\cite{chua1986double} that extensively analyzed the dynamics of the double-scroll attractor.

In our experiments, $a$ is sampled continuously from $[8.5,\, 10.5]$ using $N_\text{train}=50$ Sobol quasi-random points.
We use a Runge--Kutta integrator (\texttt{RK45}) with solver step $\delta t = 0.001$ and sampling interval $\Delta t = 0.05$ time units.
A transient of $t_\text{trans}=100$ time units is discarded before recording each trajectory.
For each parameter value, we simulate $\nicsTrain=10$ initial conditions, each integrated up to $t_\text{end}=20$ time units.
The noise level during training is set to $\trainnoise=5\%$.

For interpolation evaluation, we sample 10 parameter values within $[8.5,\, 10.5]$ using a held-out Sobol seed and simulate \( \nicsTest=10 \) trajectories per value up to $t_\text{end}=50$ time units.
For extrapolation evaluation, we use 10 parameter values outside the training range: $a \in \{8.1, 8.2, 8.25, 8.3, 8.4, 10.6, 10.7, 10.75, 10.8, 10.9\}$ (5 below and 5 above the training interval), simulated with the same protocol.

\begin{figure}[!hbt]
    \centering
    \includegraphics[width=0.75\textwidth]{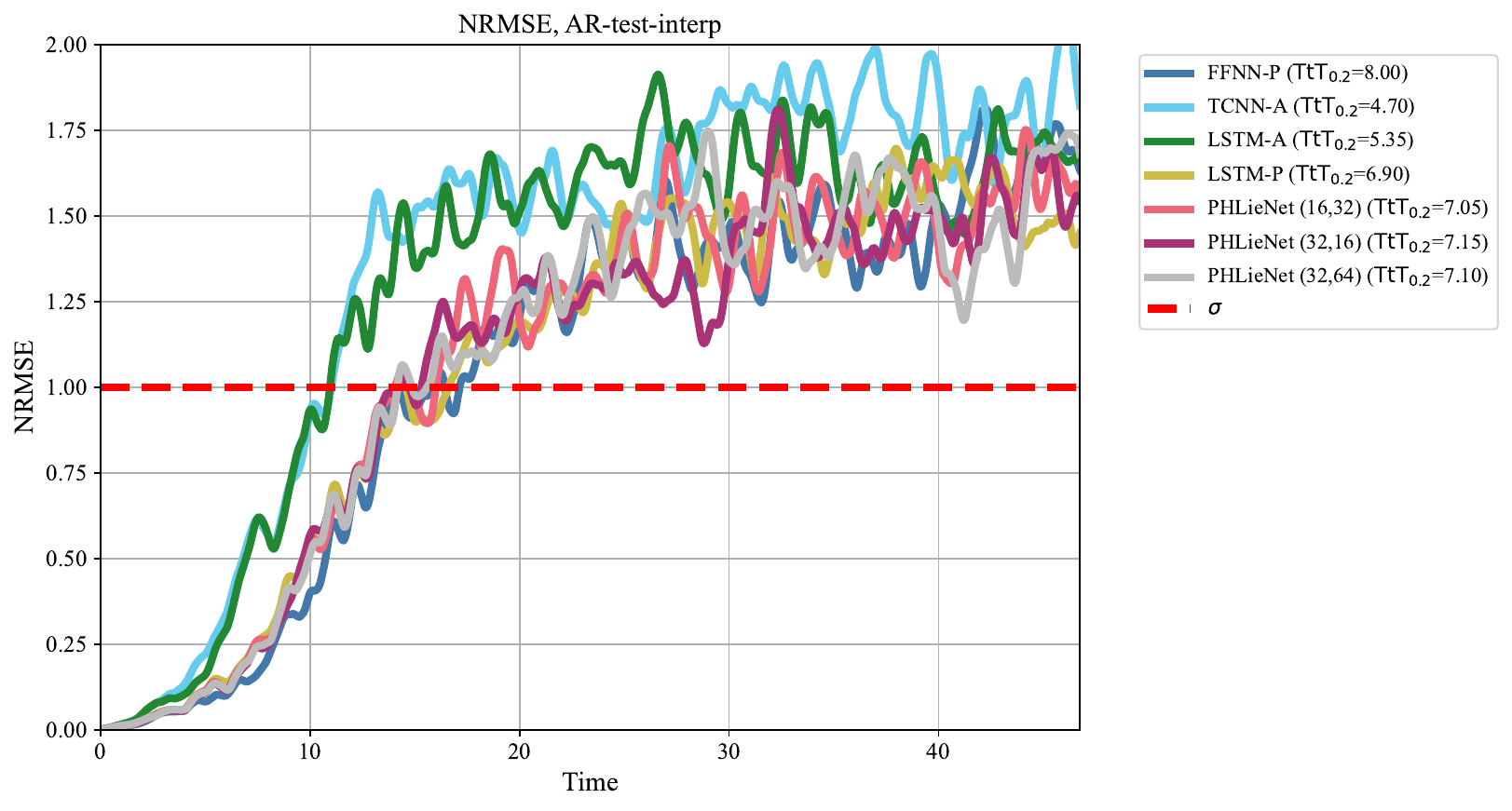}
    \caption{Normalized Root Mean Squared Error (NRMSE) evolution in time for the interpolation task.}
    \label{fig:chua:interp:rmse_evolution}
\end{figure}

\begin{figure}[H]
    \centering
    \begin{subfigure}{0.45\textwidth}
        \centering
        \includegraphics[width=\textwidth]{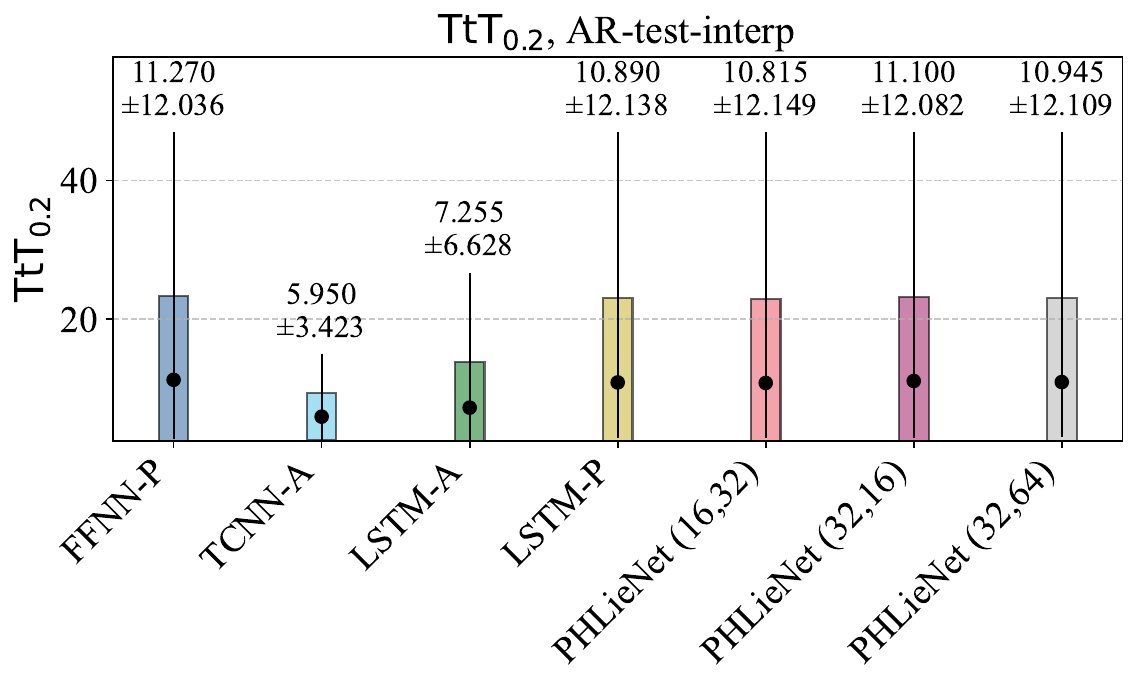}
        \caption{Time-to-Threshold (TtT) metric.}
        \label{fig:chua:interp:ttt}
    \end{subfigure}
    \quad
    \begin{subfigure}{0.45\textwidth}
        \centering
        \includegraphics[width=\textwidth]{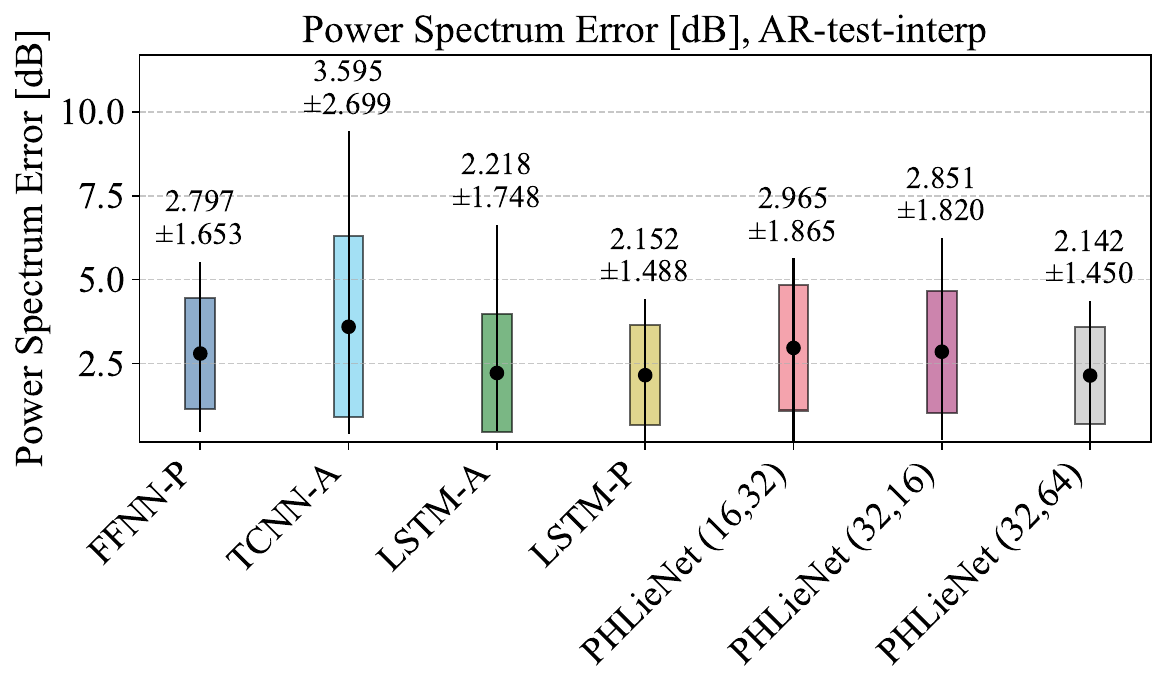}
        \caption{Power spectrum error.}
        \label{fig:chua:interp:psd_error}
    \end{subfigure}
    \caption{
        Model performance on Chua's circuit for the interpolation task.
        (a) Time-to-Threshold (TtT) metric.
        (b) Power spectrum error.
    }
    \label{fig:chua:interp}
\end{figure}

We include Chua's circuit in the appendix as a striking example where PHLieNet's core assumption is violated: the piecewise-linear Chua diode introduces hard kinks in the vector field, and the parameter $a$ drives the system through rapid bifurcations within the training range, so the attractor geometry does not vary smoothly with the parameter.
In~\Cref{fig:chua:interp:rmse_evolution,fig:chua:interp}, we benchmark performance on the interpolation task.
The NRMSE evolution in~\Cref{fig:chua:interp:rmse_evolution} shows that all parameter-informed models degrade at similar rates.
The aggregate $\ttt_{0.2}$ in the legend reflects this near-tie.
The per-parameter $\ttt_{0.2}$ values in~\Cref{fig:chua:interp:ttt} confirm this: all parameter-informed models cluster tightly (PHLieNet variants 10.8--11.1, LSTM-P 10.9), with no meaningful difference between architectures.
The large standard deviation ($\approx 12$) across test parameter values signals a bimodal distribution: some values of $a$ produce attractors that all models predict well, while others lie near bifurcation boundaries where all models fail quickly.
Parameter-agnostic baselines lag noticeably (LSTM-A 7.3, TCNN-A 6.0), confirming that conditioning on $a$ still helps at interpolation conditions.

\begin{figure}[!hbt]
    \centering
    \includegraphics[width=0.75\textwidth]{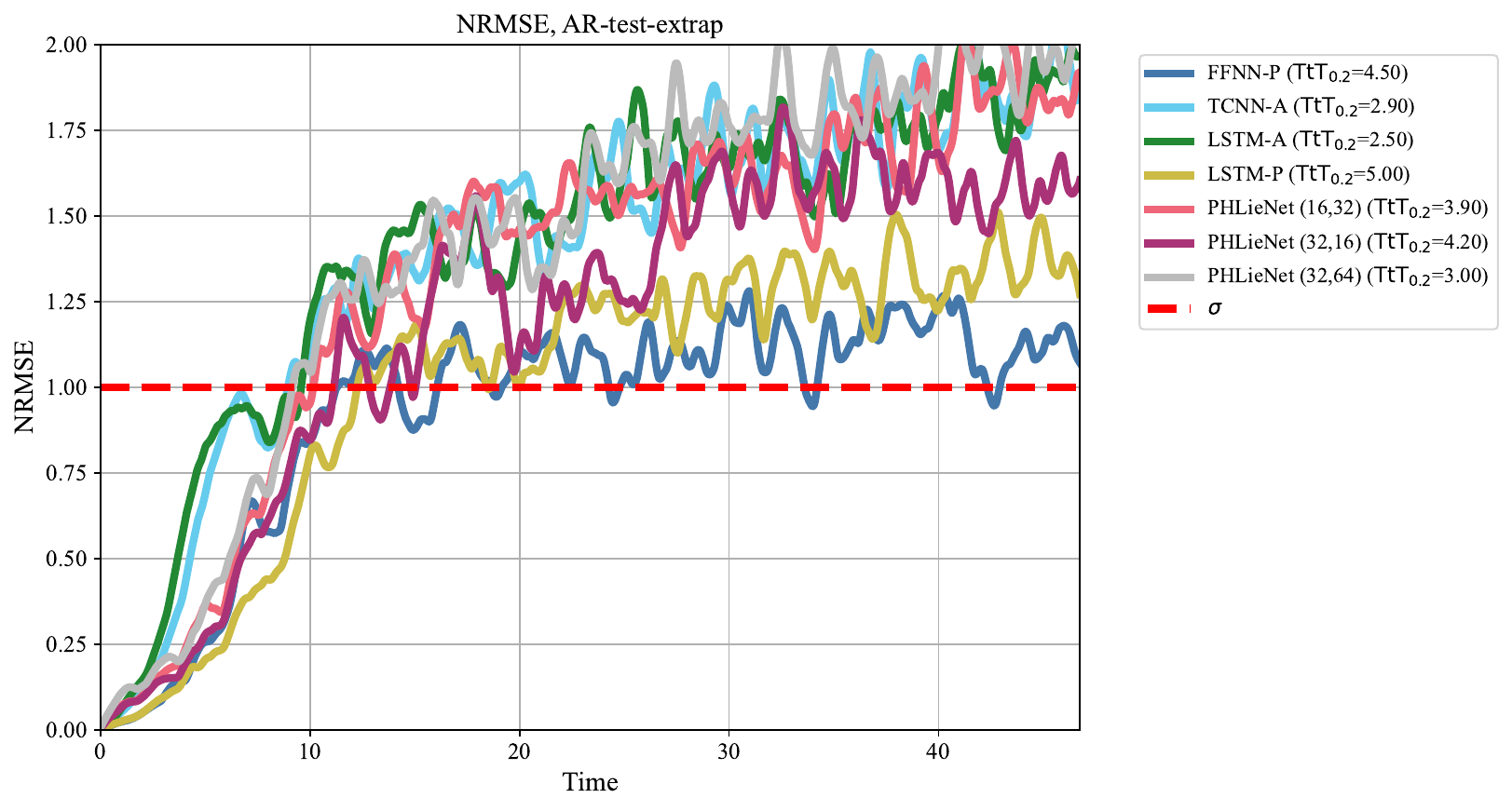}
    \caption{Normalized Root Mean Squared Error (NRMSE) evolution in time for the extrapolation task.}
    \label{fig:chua:extrap:rmse_evolution}
\end{figure}

\begin{figure}[H]
    \centering
    \begin{subfigure}{0.45\textwidth}
        \centering
        \includegraphics[width=\textwidth]{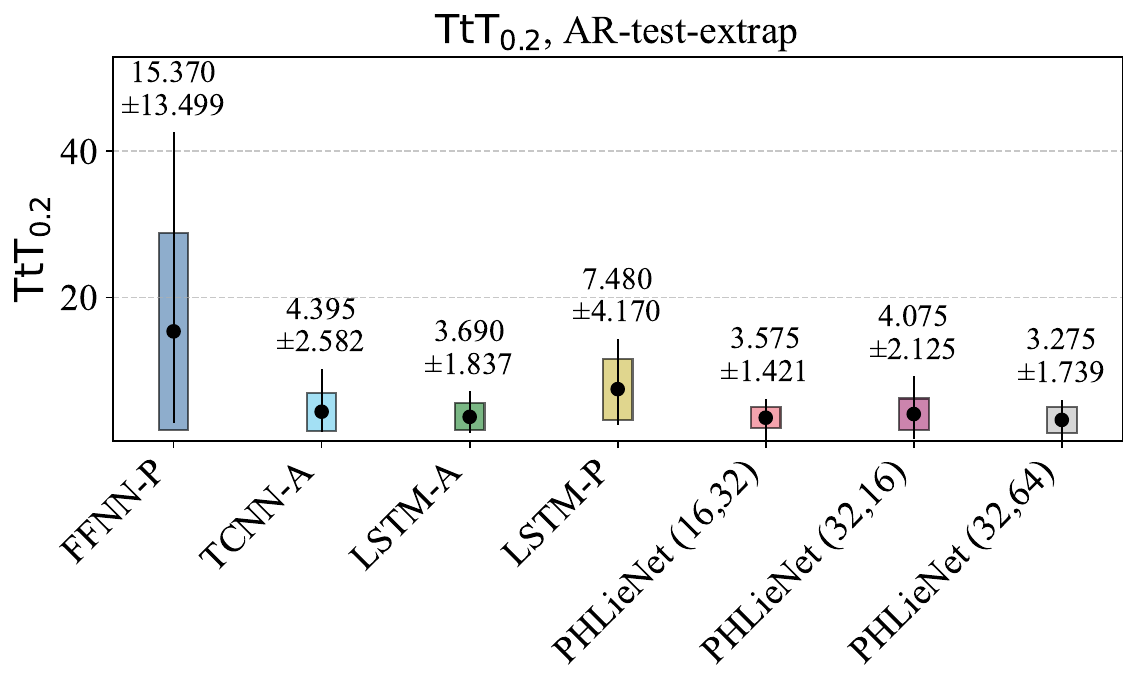}
        \caption{Time-to-Threshold (TtT) metric.}
        \label{fig:chua:extrap:ttt}
    \end{subfigure}
    \quad
    \begin{subfigure}{0.45\textwidth}
        \centering
        \includegraphics[width=\textwidth]{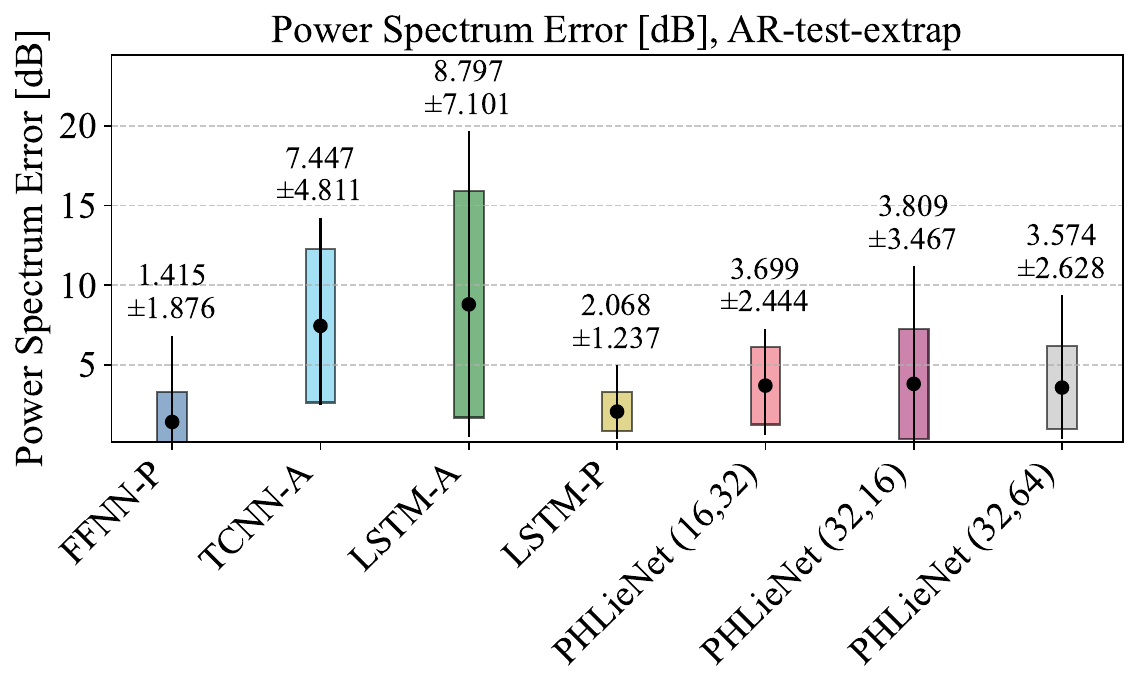}
        \caption{Power spectrum error.}
        \label{fig:chua:extrap:psd_error}
    \end{subfigure}
    \caption{
        Model performance on Chua's circuit for the extrapolation task.
        (a) Time-to-Threshold (TtT) metric.
        (b) Power spectrum error.
    }
    \label{fig:chua:extrap}
\end{figure}

Parameter extrapolation results in~\Cref{fig:chua:extrap:rmse_evolution,fig:chua:extrap} make the violation of PHLieNet's assumptions even more apparent.
The NRMSE evolution in~\Cref{fig:chua:extrap:rmse_evolution} shows PHLieNet variants degrading faster than LSTM-P.
The aggregate $\ttt_{0.2}$ in the legend reflects this ordering.
The per-parameter $\ttt_{0.2}$ values in~\Cref{fig:chua:extrap:ttt} quantify the gap: LSTM-P leads with 7.5 time units, while all three PHLieNet variants fall below even the agnostic baselines (PHLieNet variants 3.3--4.1 vs LSTM-A 3.7, TCNN-A 4.4).
This inversion (PHLieNet underperforming agnostic models on extrapolation) is a direct consequence of the smoothness assumption breaking down.
When weight-space interpolation is applied to a system whose attractor geometry changes non-smoothly with the parameter, the generated weights are a poor model of the unseen dynamics.
The power spectrum error in~\Cref{fig:chua:extrap:psd_error} tells the same story: LSTM-P achieves the lowest spectral error (2.1) among parameter-informed models, while PHLieNet variants show substantially higher values (3.6--3.8).
\subsection{Duffing Oscillator}

The Duffing oscillator is a periodically forced nonlinear oscillator that exhibits a rich variety of dynamical regimes depending on the forcing amplitude.
Its dynamics are governed by the second-order ODE
\begin{equation}
    \ddot{x} + \delta \dot{x} + \alpha x + \beta x^3 = \gamma \cos(\omega t),
\label{eq:duffing}
\end{equation}
where $\state = [x, \dot{x}]^T \in \RR^2$ represents position and velocity.
We vary the forcing amplitude $\gamma$, which controls the transition from near-harmonic oscillations to broadened, multi-loop limit cycles.
The remaining parameters are fixed at $\alpha = -1$, $\beta = 1$, $\delta = 0.3$, and $\omega = 1.2$.

In our experiments, $\gamma$ is sampled continuously from $[0.10,\, 0.80]$ using $N_\text{train} = 100$ Sobol quasi-random points.
We use a Runge--Kutta integrator (\texttt{RK45}) with solver step $\delta t = 0.001$ and sampling interval $\Delta t = 0.02$ time units.
A transient of $t_\text{trans} = 200$ time units is discarded before recording each trajectory.
For each parameter value, we simulate $\nicsTrain = 10$ initial conditions, each integrated up to $t_\text{end} = 50$ time units.
The noise level during training is set to $\trainnoise = 5\%$.

For interpolation evaluation, we sample 50 parameter values within $[0.10,\, 0.80]$ using a held-out Sobol seed and simulate $\nicsTest = 10$ trajectories per value up to $t_\text{end} = 50$ time units.

\begin{figure}[!hbt]
    \centering
    \includegraphics[width=0.75\textwidth]{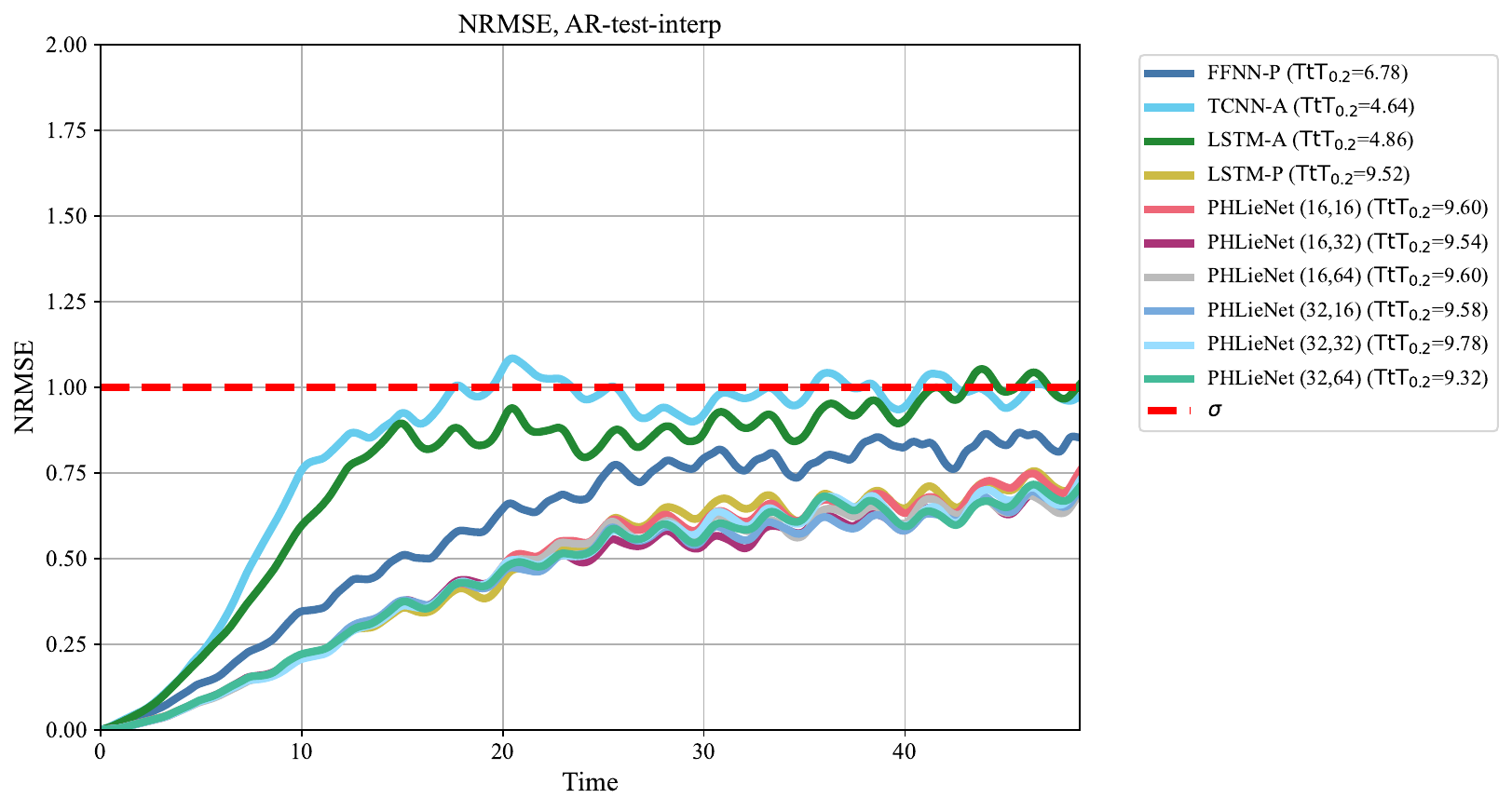}
    \caption{Normalized Root Mean Squared Error (NRMSE) evolution in time for the interpolation task on the Duffing oscillator.}
    \label{fig:duffing:interp:rmse_evolution}
\end{figure}

\begin{figure}[H]
    \centering
    \begin{subfigure}{0.45\textwidth}
        \centering
        \includegraphics[width=\textwidth]{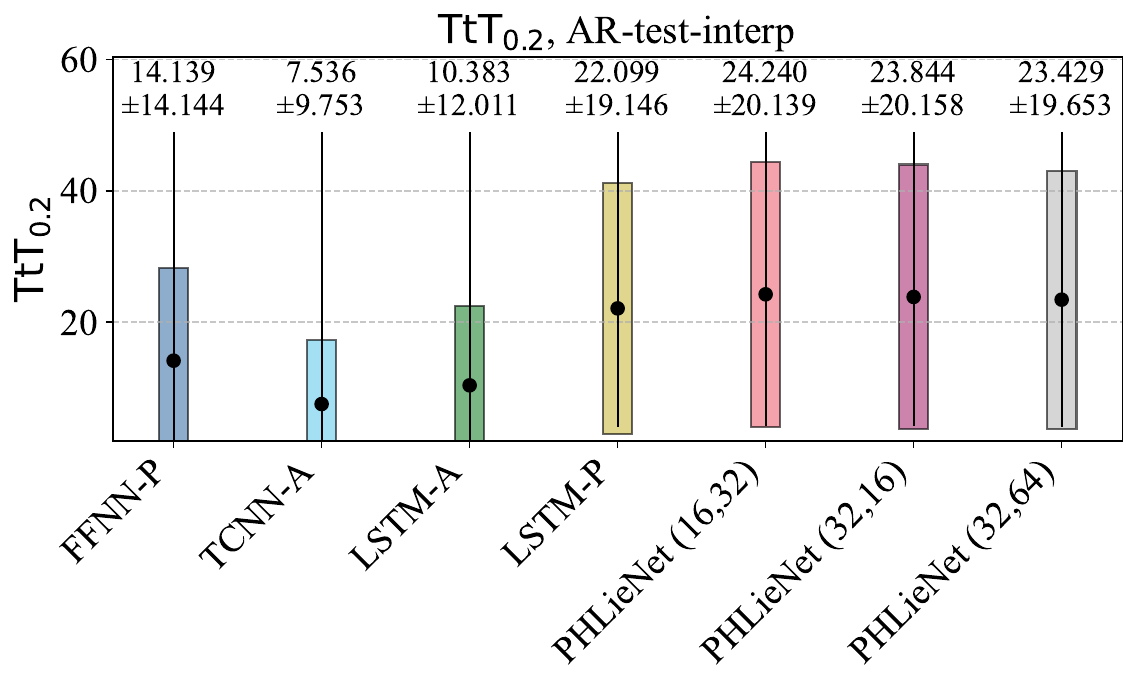}
        \caption{Time-to-Threshold (TtT) metric.}
        \label{fig:duffing:interp:ttt}
    \end{subfigure}
    \quad
    \begin{subfigure}{0.45\textwidth}
        \centering
        \includegraphics[width=\textwidth]{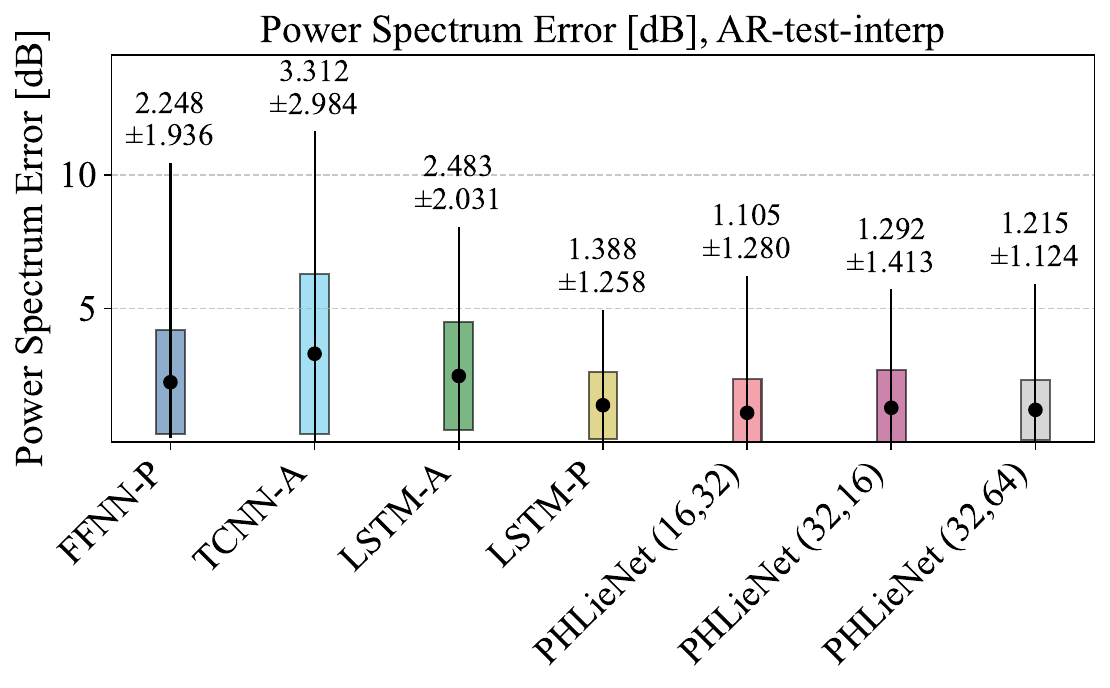}
        \caption{Power spectrum error.}
        \label{fig:duffing:interp:psd_error}
    \end{subfigure}
    \caption{
        Model performance on the Duffing oscillator for the interpolation task.
        (a) Time-to-Threshold (TtT) metric.
        (b) Power spectrum error.
    }
    \label{fig:duffing:interp}
\end{figure}

In~\Cref{fig:duffing:interp:rmse_evolution,fig:duffing:interp}, we evaluate the baselines and three representative PHLieNet configurations (PHLieNet\textsubscript{(16,32)}, PHLieNet\textsubscript{(32,16)}, and PHLieNet\textsubscript{(32,64)}) on autoregressive prediction at held-out parameter values within the training range.
All three PHLieNet variants outperform the baselines, with PHLieNet\textsubscript{(16,32)} achieving a per-parameter mean $\ttt_{0.2}$ of 24.2 compared to 22.0 for LSTM-P (a $+$10\% improvement).
The NRMSE evolution curves in~\Cref{fig:duffing:interp:rmse_evolution} confirm that PHLieNet maintains consistently lower prediction error throughout the rollout horizon.
This result is robust across seeds: every PHLieNet seed outperforms every LSTM-P seed on per-parameter $\ttt_{0.2}$.

Extrapolation results are omitted for this system.
The forcing amplitude $\gamma$ controls qualitative bifurcations in the Duffing oscillator, and parameter values far outside the training range induce fundamentally different dynamical regimes (e.g., nearly unforced motion at $\gamma \approx 0.02$ versus strongly forced multi-loop attractors at $\gamma > 0.9$), making extrapolation unreliable for all methods.

\begin{figure}[H]
    \centering
    \begin{subfigure}{0.90\textwidth}
        \centering
        \includegraphics[width=\textwidth]{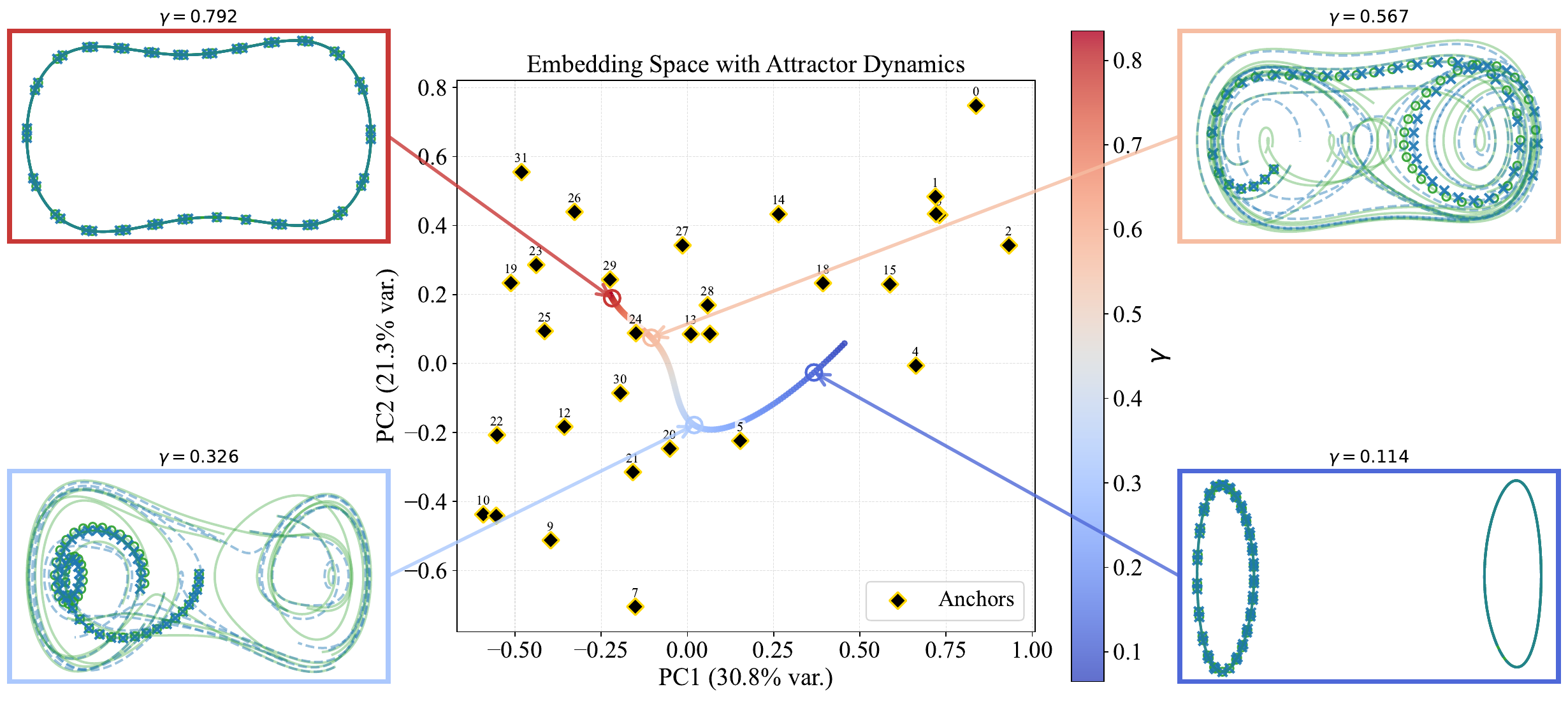}
        \caption{PCA of anchor embeddings coloured by $\gamma$, with attractor insets.}
        \label{fig:embedding:duffing:pca}
    \end{subfigure}

    \vspace{0.5em}

    \begin{subfigure}{0.54\textwidth}
        \centering
        \includegraphics[width=\textwidth]{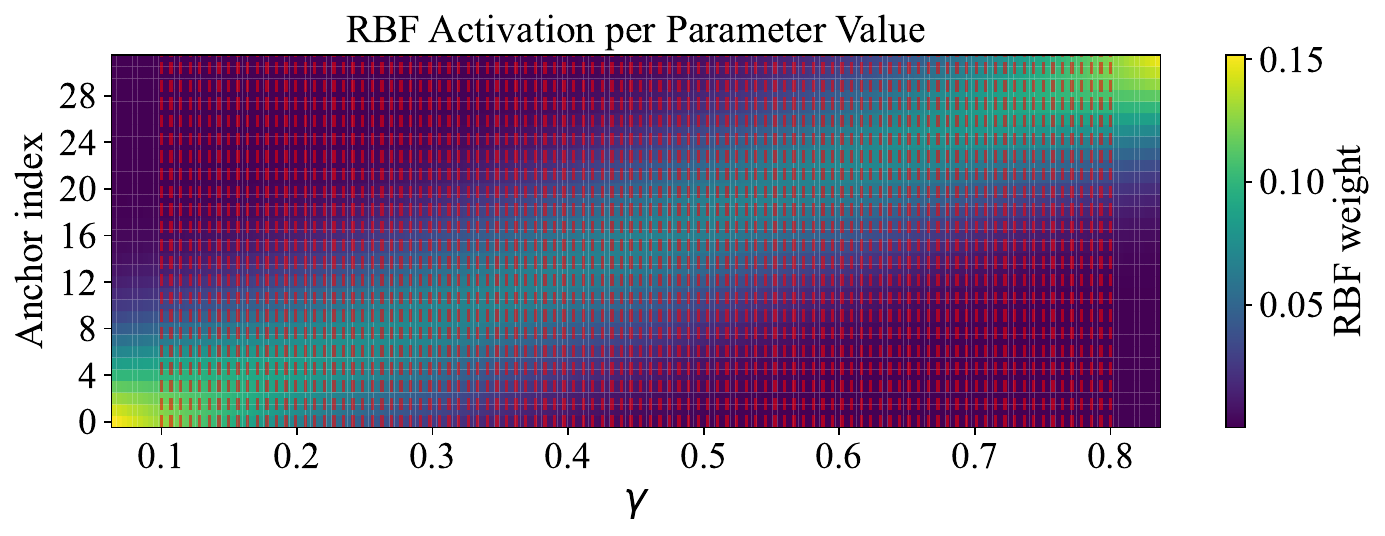}
        \caption{RBF activation profile vs.\ $\gamma$.}
        \label{fig:embedding:duffing:rbf}
    \end{subfigure}
    \hfill
    \begin{subfigure}{0.34\textwidth}
        \centering
        \includegraphics[width=\textwidth]{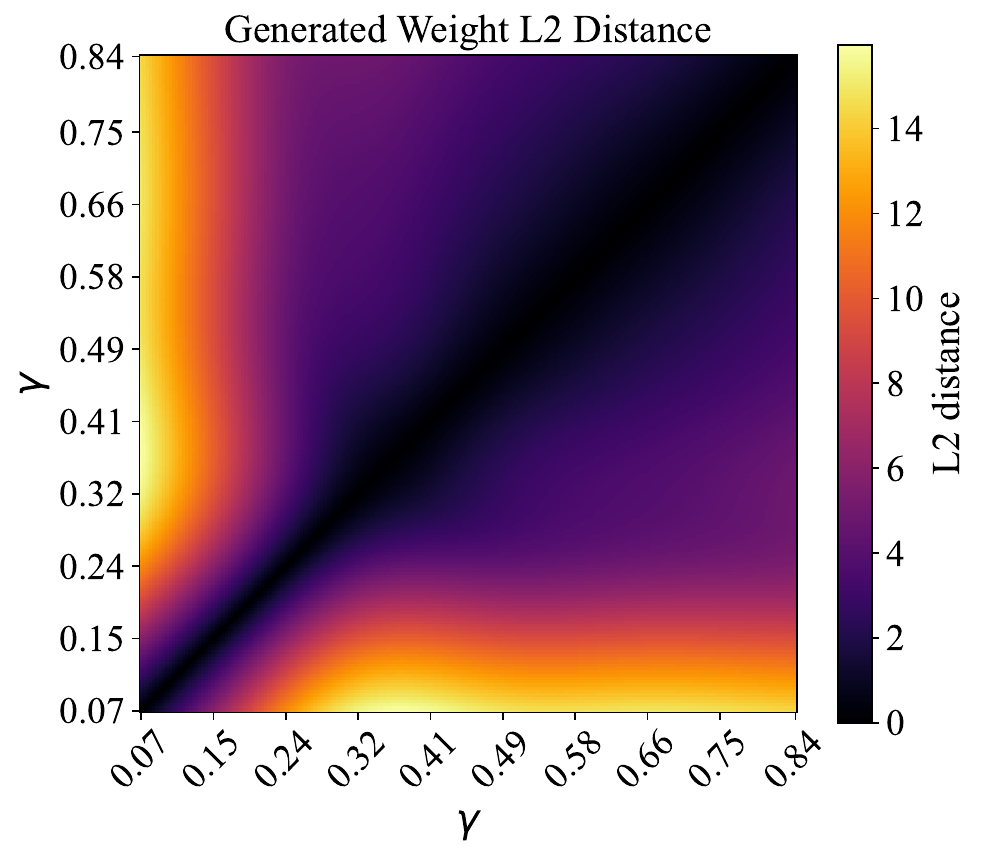}
        \caption{Pairwise weight distance matrix.}
        \label{fig:embedding:duffing:wdist}
    \end{subfigure}

    \caption{
        Embedding space analysis for the Duffing oscillator (PHLieNet\textsubscript{(32,16)}).
        \textit{Top}: PCA projection of the learned anchor embeddings coloured by $\gamma$, with attractor insets at representative parameter values showing the transition from simple elliptical orbits (low $\gamma$) to broadened multi-loop limit cycles (high $\gamma$).
        \textit{Bottom left}: RBF activation profile, with anchors activating in smooth, overlapping bands.
        \textit{Bottom right}: Pairwise L2 distance matrix between generated weights, with near-diagonal structure confirming smooth weight variation with $\gamma$.
    }
    \label{fig:embedding:duffing}
\end{figure}

The analysis in~\Cref{fig:embedding:duffing} provides further insight into PHLieNet's learned representation.
The PCA projection in~\Cref{fig:embedding:duffing:pca} shows that the learned anchor embeddings organize along a smooth manifold monotonically ordered by $\gamma$, with attractor insets illustrating how the dynamics transition from near-harmonic oscillations at low forcing to broadened limit cycles at higher amplitudes.
The RBF activation heatmap in~\Cref{fig:embedding:duffing:rbf} reveals smooth, overlapping bands of anchor activation, and the weight distance matrix in~\Cref{fig:embedding:duffing:wdist} confirms monotonically increasing weight divergence with parameter distance, consistent with the pattern observed for the main-text systems in~\Cref{sec:embedding_analysis}.

\bibliography{references}

@article{rostamijavanani2023dyncgan,
  title={Data-driven modeling of parameterized nonlinear dynamical systems with a dynamics-embedded conditional generative adversarial network},
  author={Rostamijavanani, Abdolvahhab and Li, Shanwu and Yang, Yongchao},
  journal={Journal of Engineering Mechanics},
  volume={149},
  number={11},
  pages={04023094},
  year={2023},
  publisher={American Society of Civil Engineers}
}

@article{li2023generative,
  title={A deep generative framework for data-driven surrogate modeling and visualization of parameterized nonlinear dynamical systems},
  author={Li, Shanwu and Yang, Yongchao},
  journal={Nonlinear Dynamics},
  volume={111},
  number={11},
  pages={10287--10307},
  year={2023},
  doi={10.1007/s11071-023-08391-0}
}

@article{li2025bifurcation,
  title={Data-driven modeling of bifurcation systems by learning the bifurcation parameter generalization},
  author={Li, Shanwu and Yang, Yongchao},
  journal={Nonlinear Dynamics},
  volume={113},
  pages={1163--1174},
  year={2025},
  doi={10.1007/s11071-024-10304-8}
}

@article{yin2021leads,
  title={Leads: Learning dynamical systems that generalize across environments},
  author={Yin, Yuan and Ayed, Ibrahim and de B{\'e}zenac, Emmanuel and Baskiotis, Nicolas and Gallinari, Patrick},
  journal={Advances in Neural Information Processing Systems},
  volume={34},
  pages={7561--7573},
  year={2021}
}

@inproceedings{kirchmeyer2022generalizing,
  title={Generalizing to new physical systems via context-informed dynamics model},
  author={Kirchmeyer, Matthieu and Yin, Yuan and Don{\`a}, J{\'e}r{\'e}mie and Baskiotis, Nicolas and Rakotomamonjy, Alain and Gallinari, Patrick},
  booktitle={International Conference on Machine Learning},
  pages={11283--11301},
  year={2022},
  organization={PMLR}
}

@inproceedings{karbasia2025parametric,
  title={A parametric LSTM neural network for predicting flow field dynamics across a design space},
  author={Karbasia, Hamid R and van Rees, Wim M},
  booktitle={Proceedings A},
  volume={481},
  number={2307},
  pages={20240055},
  year={2025},
  organization={The Royal Society}
}

@article{benner2015survey,
  title={A survey of projection-based model reduction methods for parametric dynamical systems},
  author={Benner, Peter and Gugercin, Serkan and Willcox, Karen},
  journal={SIAM review},
  volume={57},
  number={4},
  pages={483--531},
  year={2015},
  publisher={SIAM}
}

@article{agathos2024accelerating,
  title={Accelerating structural dynamics simulations with localised phenomena through matrix compression and projection-based model order reduction},
  author={Agathos, Konstantinos and Vlachas, Konstantinos and Garland, Anthony and Chatzi, Eleni},
  journal={International Journal for Numerical Methods in Engineering},
  volume={125},
  number={10},
  pages={e7445},
  year={2024},
  publisher={Wiley Online Library}
}

@article{vlachas2025utility,
  title={Utility-Driven Adaptive Model Selection for Digital Twinning},
  author={Vlachas, Konstantinos and Kamariotis, Antonios and Chatzi, Eleni},
  journal={SSRN},
  year={2025},
  publisher={Social Science Research Network}
}

@article{vlachas2024parametric,
  title={Parametric reduced-order modeling for component-oriented treatment and localized nonlinear feature inclusion},
  author={Vlachas, Konstantinos and Garland, Anthony and Quinn, D Dane and Chatzi, Eleni},
  journal={Nonlinear Dynamics},
  volume={112},
  number={5},
  pages={3399--3420},
  year={2024},
  publisher={Springer}
}

@article{proctor2016dynamic,
  title={Dynamic mode decomposition with control},
  author={Proctor, Joshua L and Brunton, Steven L and Kutz, J Nathan},
  journal={SIAM Journal on Applied Dynamical Systems},
  volume={15},
  number={1},
  pages={142--161},
  year={2016},
  publisher={SIAM}
}

@article{duthe2025mechanistic,
  title={A Mechanistic Analysis of Transformers for Dynamical Systems},
  author={Duth{\'e}, Gregory and Evangelou, Nikolaos and Liu, Wei and Kevrekidis, Ioannis G and Chatzi, Eleni},
  journal={arXiv preprint arXiv:2512.21113},
  year={2025}
}

@article{amoudruz2025bayesian,
  title={Bayesian Inference for PDE-based Inverse Problems using the Optimization of a Discrete Loss},
  author={Amoudruz, Lucas and Litvinov, Sergey and Papadimitriou, Costas and Koumoutsakos, Petros},
  journal={arXiv preprint arXiv:2510.15664},
  year={2025}
}

@article{karnakov2024solving,
  title={Solving inverse problems in physics by optimizing a discrete loss: Fast and accurate learning without neural networks},
  author={Karnakov, Petr and Litvinov, Sergey and Koumoutsakos, Petros},
  journal={PNAS nexus},
  volume={3},
  number={1},
  pages={pgae005},
  year={2024},
  publisher={Oxford University Press US}
  }

@article{de2026nonlinear,
  title={Nonlinear projection-based model order reduction with machine learning regression for closure error modeling in the latent space},
  author={De Parga, S Ares and Tezaur, Radek and Hern{\'a}ndez, Carlos G and Farhat, Charbel},
  journal={Computer Methods in Applied Mechanics and Engineering},
  volume={448},
  pages={118443},
  year={2026},
  publisher={Elsevier}
}

@article{barnett2022quadratic,
  title={Quadratic approximation manifold for mitigating the Kolmogorov barrier in nonlinear projection-based model order reduction},
  author={Barnett, Joshua and Farhat, Charbel},
  journal={Journal of Computational Physics},
  volume={464},
  pages={111348},
  year={2022},
  publisher={Elsevier}
}

@inproceedings{raonic2023convolutionalICLR,
  title={Convolutional neural operators},
  author={Raonic, Bogdan and Molinaro, Roberto and Rohner, Tobias and Mishra, Siddhartha and de Bezenac, Emmanuel},
  booktitle={ICLR 2023 Workshop on Physics for Machine Learning},
  year={2023}
}

@article{raonic2023convolutional,
  title={Convolutional neural operators for robust and accurate learning of PDEs},
  author={Raonic, Bogdan and Molinaro, Roberto and De Ryck, Tim and Rohner, Tobias and Bartolucci, Francesca and Alaifari, Rima and Mishra, Siddhartha and de B{\'e}zenac, Emmanuel},
  journal={Advances in Neural Information Processing Systems},
  volume={36},
  pages={77187--77200},
  year={2023}
}

@article{vinuesa2022enhancing,
  title={Enhancing computational fluid dynamics with machine learning},
  author={Vinuesa, Ricardo and Brunton, Steven L},
  journal={Nature Computational Science},
  volume={2},
  number={6},
  pages={358--366},
  year={2022},
  publisher={Nature Publishing Group US New York}
}

@article{van1926lxxxviii,
  title={LXXXVIII. On “relaxation-oscillations”},
  author={Van der Pol, Balth},
  journal={The London, Edinburgh, and Dublin Philosophical Magazine and Journal of Science},
  volume={2},
  number={11},
  pages={978--992},
  year={1926},
  publisher={Taylor \& Francis}
}

@article{brenner2024learning,
  title={Learning Interpretable Hierarchical Dynamical Systems Models from Time Series Data},
  author={Brenner, Manuel and Weber, Elias and Koppe, Georgia and Durstewitz, Daniel},
  journal={arXiv preprint arXiv:2410.04814},
  year={2024}
}

@article{liu2022hierarchical,
  title={Hierarchical deep learning of multiscale differential equation time-steppers},
  author={Liu, Yuying and Kutz, J Nathan and Brunton, Steven L},
  journal={Philosophical Transactions of the Royal Society A},
  volume={380},
  number={2229},
  pages={20210200},
  year={2022},
  publisher={The Royal Society}
}

@article{bai2018empirical,
  title={An empirical evaluation of generic convolutional and recurrent networks for sequence modeling},
  author={Bai, Shaojie and Kolter, J Zico and Koltun, Vladlen},
  journal={arXiv preprint arXiv:1803.01271},
  year={2018}
}

@article{takens1981dynamical,
  title={Dynamical systems and turbulence},
  author={Takens, Floris},
  journal={Warwick, 1980},
  pages={366--381},
  year={1981},
  publisher={Springer-Verlag}
}

@article{graves2013generating,
  title={Generating sequences with recurrent neural networks},
  author={Graves, Alex},
  journal={arXiv preprint arXiv:1308.0850},
  year={2013}
}

@book{sutskever2013training,
  title={Training recurrent neural networks},
  author={Sutskever, Ilya},
  year={2013},
  publisher={University of Toronto Toronto, ON, Canada}
}

@article{cho2024parameterized,
  title={Parameterized physics-informed neural networks for parameterized PDEs},
  author={Cho, Woojin and Jo, Minju and Lim, Haksoo and Lee, Kookjin and Lee, Dongeun and Hong, Sanghyun and Park, Noseong},
  journal={arXiv preprint arXiv:2408.09446},
  year={2024}
}

@article{eyring2019taking,
  title={Taking climate model evaluation to the next level},
  author={Eyring, Veronika and Cox, Peter M and Flato, Gregory M and Gleckler, Peter J and Abramowitz, Gab and Caldwell, Peter and Collins, William D and Gier, Bettina K and Hall, Alex D and Hoffman, Forrest M and others},
  journal={Nature Climate Change},
  volume={9},
  number={2},
  pages={102--110},
  year={2019},
  publisher={Nature Publishing Group UK London}
}

@article{schultz2013reynolds,
  title={Reynolds-number scaling of turbulent channel flow},
  author={Schultz, Michael P and Flack, Karen A},
  journal={Physics of Fluids},
  volume={25},
  number={2},
  year={2013},
  publisher={AIP Publishing}
}

@article{zhai2024reconstructing,
  title={Reconstructing dynamics from sparse observations with no training on target system},
  author={Zhai, Zheng-Meng and Huang, Jun-Yin and Stern, Benjamin D and Lai, Ying-Cheng},
  journal={arXiv preprint arXiv:2410.21222},
  year={2024}
}

@article{roy2022model,
  title={Model-free prediction of multistability using echo state network},
  author={Roy, Mousumi and Mandal, Swarnendu and Hens, Chittaranjan and Prasad, Awadhesh and Kuznetsov, NV and Dev Shrimali, Manish},
  journal={Chaos: An Interdisciplinary Journal of Nonlinear Science},
  volume={32},
  number={10},
  year={2022},
  publisher={AIP Publishing}
}

@article{langer2004modeling,
  title={Modeling parameter dependence from time series},
  author={Langer, G and Parlitz, Ulrich},
  journal={Physical Review E—Statistical, Nonlinear, and Soft Matter Physics},
  volume={70},
  number={5},
  pages={056217},
  year={2004},
  publisher={APS}
}

@article{luo2024reconstructing,
  title={Reconstructing bifurcation diagrams of chaotic circuits with reservoir computing},
  author={Luo, Haibo and Du, Yao and Fan, Huawei and Wang, Xuan and Guo, Jianzhong and Wang, Xingang},
  journal={Physical Review E},
  volume={109},
  number={2},
  pages={024210},
  year={2024},
  publisher={APS}
}

@article{geneva2022transformers,
  title={Transformers for modeling physical systems},
  author={Geneva, Nicholas and Zabaras, Nicholas},
  journal={Neural Networks},
  volume={146},
  pages={272--289},
  year={2022},
  publisher={Elsevier}
}

@article{wan2018data,
  title={Data-assisted reduced-order modeling of extreme events in complex dynamical systems},
  author={Wan, Zhong Yi and Vlachas, Pantelis and Koumoutsakos, Petros and Sapsis, Themistoklis},
  journal={PloS one},
  volume={13},
  number={5},
  pages={e0197704},
  year={2018},
  publisher={Public Library of Science San Francisco, CA USA}
}

@article{fresca2022pod,
  title={POD-DL-ROM: Enhancing deep learning-based reduced order models for nonlinear parametrized PDEs by proper orthogonal decomposition},
  author={Fresca, Stefania and Manzoni, Andrea},
  journal={Computer Methods in Applied Mechanics and Engineering},
  volume={388},
  pages={114181},
  year={2022},
  publisher={Elsevier}
}

@article{barnett2023neural,
  title={Neural-network-augmented projection-based model order reduction for mitigating the Kolmogorov barrier to reducibility},
  author={Barnett, Joshua and Farhat, Charbel and Maday, Yvon},
  journal={Journal of Computational Physics},
  volume={492},
  pages={112420},
  year={2023},
  publisher={Elsevier}
}

@article{vlachas2025reduced,
  title={Reduced Order Modeling conditioned on monitored features for response and error bounds estimation in engineered systems},
  author={Vlachas, Konstantinos and Simpson, Thomas and Garland, Anthony and Quinn, D Dane and Farhat, Charbel and Chatzi, Eleni},
  journal={Mechanical Systems and Signal Processing},
  volume={226},
  pages={112261},
  year={2025},
  publisher={Elsevier}
}

@book{benner2020model,
  title={Model Order Reduction: Volume 2: Snapshot-Based Methods and Algorithms},
  author={Benner, Peter and Schilders, Wil and Grivet-Talocia, Stefano and Quarteroni, Alfio and Rozza, Gianluigi and Miguel Silveira, Lu{\'\i}s},
  year={2020},
  publisher={De Gruyter}
}

@article{champaney2022engineering,
  title={Engineering empowered by physics-based and data-driven hybrid models: A methodological overview},
  author={Champaney, Victor and Chinesta, Francisco and Cueto, Elias},
  journal={International Journal of Material Forming},
  volume={15},
  number={3},
  pages={31},
  year={2022},
  publisher={Springer}
}

@article{amsallem2016pebl,
  title={PEBL-ROM: Projection-error based local reduced-order models},
  author={Amsallem, David and Haasdonk, Bernard},
  journal={Advanced Modeling and Simulation in Engineering Sciences},
  volume={3},
  pages={1--25},
  year={2016},
  publisher={Springer}
}

@article{huang2022meta,
  title={Meta-auto-decoder for solving parametric partial differential equations},
  author={Huang, Xiang and Ye, Zhanhong and Liu, Hongsheng and Ji, Shi and Wang, Zidong and Yang, Kang and Li, Yang and Wang, Min and Chu, Haotian and Yu, Fan and others},
  journal={Advances in Neural Information Processing Systems},
  volume={35},
  pages={23426--23438},
  year={2022}
}

@article{qin2022meta,
  title={Meta-pde: Learning to solve pdes quickly without a mesh},
  author={Qin, Tian and Beatson, Alex and Oktay, Deniz and McGreivy, Nick and Adams, Ryan P},
  journal={arXiv preprint arXiv:2211.01604},
  year={2022}
}

@article{galanti2020modularity,
  title={On the modularity of hypernetworks},
  author={Galanti, Tomer and Wolf, Lior},
  journal={Advances in Neural Information Processing Systems},
  volume={33},
  pages={10409--10419},
  year={2020}
}

@inproceedings{klein2015dynamic,
  title={A dynamic convolutional layer for short range weather prediction},
  author={Klein, Benjamin and Wolf, Lior and Afek, Yehuda},
  booktitle={Proceedings of the IEEE Conference on Computer Vision and Pattern Recognition},
  pages={4840--4848},
  year={2015}
}

@article{guo2025parametric,
  title={Learning Parametric Koopman Decompositions for Prediction and Control},
  author={Guo, Yue and Korda, Milan and Kevrekidis, Ioannis G and Li, Qianxiao},
  journal={SIAM Journal on Applied Dynamical Systems},
  volume={24},
  number={1},
  pages={744--781},
  year={2025},
  publisher={SIAM},
  doi={10.1137/23M1604576}
}

@article{franco2023deep,
  title={A deep learning approach to reduced order modelling of parameter dependent partial differential equations},
  author={Franco, Nicola and Manzoni, Andrea and Zunino, Paolo},
  journal={Mathematics of Computation},
  volume={92},
  number={340},
  pages={483--524},
  year={2023}
}

@article{wagner2023stacked,
  title={Stacked tensorial neural networks for reduced-order modeling of a parametric partial differential equation},
  author={Wagner, Caleb G},
  journal={arXiv preprint arXiv:2312.14979},
  year={2023}
}

@article{wright2021ranger21,
  title={Ranger21: a synergistic deep learning optimizer},
  author={Wright, Less and Demeure, Nestor},
  journal={arXiv preprint arXiv:2106.13731},
  year={2021}
}

@article{paszke2019pytorch,
  title={Pytorch: An imperative style, high-performance deep learning library},
  author={Paszke, Adam and Gross, Sam and Massa, Francisco and Lerer, Adam and Bradbury, James and Chanan, Gregory and Killeen, Trevor and Lin, Zeming and Gimelshein, Natalia and Antiga, Luca and others},
  journal={Advances in neural information processing systems},
  volume={32},
  year={2019}
}

@article{kovachki2023neural,
  title={Neural operator: Learning maps between function spaces with applications to pdes},
  author={Kovachki, Nikola and Li, Zongyi and Liu, Burigede and Azizzadenesheli, Kamyar and Bhattacharya, Kaushik and Stuart, Andrew and Anandkumar, Anima},
  journal={Journal of Machine Learning Research},
  volume={24},
  number={89},
  pages={1--97},
  year={2023}
}

@article{hochreiter1997long,
  title={Long Short-term Memory},
  author={Hochreiter, S},
  journal={Neural Computation MIT-Press},
  year={1997}
}

@article{pathak2017using,
  title={Using machine learning to replicate chaotic attractors and calculate Lyapunov exponents from data},
  author={Pathak, Jaideep and Lu, Zhixin and Hunt, Brian R and Girvan, Michelle and Ott, Edward},
  journal={Chaos: An Interdisciplinary Journal of Nonlinear Science},
  volume={27},
  number={12},
  year={2017},
  publisher={AIP Publishing}
}

@article{pathak2018model,
  title={Model-free prediction of large spatiotemporally chaotic systems from data: A reservoir computing approach},
  author={Pathak, Jaideep and Hunt, Brian and Girvan, Michelle and Lu, Zhixin and Ott, Edward},
  journal={Physical review letters},
  volume={120},
  number={2},
  pages={024102},
  year={2018},
  publisher={APS}
}

@article{kivcic2023adaptive,
  title={Adaptive learning of effective dynamics for online modeling of complex systems},
  author={Ki{\v{c}}i{\'c}, Ivica and Vlachas, Pantelis R and Arampatzis, Georgios and Chatzimanolakis, Michail and Guibas, Leonidas and Koumoutsakos, Petros},
  journal={Computer Methods in Applied Mechanics and Engineering},
  volume={415},
  pages={116204},
  year={2023},
  publisher={Elsevier}
}

@article{vlachas2022multiscale,
  title={Multiscale simulations of complex systems by learning their effective dynamics},
  author={Vlachas, Pantelis R and Arampatzis, Georgios and Uhler, Caroline and Koumoutsakos, Petros},
  journal={Nature Machine Intelligence},
  volume={4},
  number={4},
  pages={359--366},
  year={2022},
  publisher={Nature Publishing Group UK London}
}

@article{vlachas2018data,
  title={Data-driven forecasting of high-dimensional chaotic systems with long short-term memory networks},
  author={Vlachas, Pantelis R and Byeon, Wonmin and Wan, Zhong Y and Sapsis, Themistoklis P and Koumoutsakos, Petros},
  journal={Proceedings of the Royal Society A: Mathematical, Physical and Engineering Sciences},
  volume={474},
  number={2213},
  pages={20170844},
  year={2018},
  publisher={The Royal Society Publishing}
}

@article{vlachas2020backpropagation,
  title={Backpropagation algorithms and reservoir computing in recurrent neural networks for the forecasting of complex spatiotemporal dynamics},
  author={Vlachas, Pantelis-Rafail and Pathak, Jaideep and Hunt, Brian R and Sapsis, Themistoklis P and Girvan, Michelle and Ott, Edward and Koumoutsakos, Petros},
  journal={Neural Networks},
  volume={126},
  pages={191--217},
  year={2020},
  publisher={Elsevier}
}

@article{liu2018darts,
  title={Darts: Differentiable architecture search},
  author={Liu, Hanxiao and Simonyan, Karen and Yang, Yiming},
  journal={arXiv preprint arXiv:1806.09055},
  year={2018}
}

@inproceedings{pham2018efficient,
  title={Efficient neural architecture search via parameters sharing},
  author={Pham, Hieu and Guan, Melody and Zoph, Barret and Le, Quoc and Dean, Jeff},
  booktitle={International conference on machine learning},
  pages={4095--4104},
  year={2018},
  organization={PMLR}
}

@article{zoph2016neural,
  title={Neural architecture search with reinforcement learning},
  author={Zoph, B},
  journal={arXiv preprint arXiv:1611.01578},
  year={2016}
}

@inproceedings{ravi2017optimization,
  title={Optimization as a model for few-shot learning},
  author={Ravi, Sachin and Larochelle, Hugo},
  booktitle={International conference on learning representations},
  year={2017}
}

@inproceedings{finn2017model,
  title={Model-agnostic meta-learning for fast adaptation of deep networks},
  author={Finn, Chelsea and Abbeel, Pieter and Levine, Sergey},
  booktitle={International conference on machine learning},
  pages={1126--1135},
  year={2017},
  organization={PMLR}
}

@inproceedings{fanaskov2023spectral,
  title={Spectral neural operators},
  author={Fanaskov, Vladimir Sergeevich and Oseledets, Ivan V},
  booktitle={Doklady Mathematics},
  volume={108},
  number={Suppl 2},
  pages={S226--S232},
  year={2023},
  organization={Springer}
}

@article{li2020fourier,
  title={Fourier neural operator for parametric partial differential equations},
  author={Li, Zongyi and Kovachki, Nikola and Azizzadenesheli, Kamyar and Liu, Burigede and Bhattacharya, Kaushik and Stuart, Andrew and Anandkumar, Anima},
  journal={arXiv preprint arXiv:2010.08895},
  year={2020}
}

@article{raissi2019physics,
  title={Physics-informed neural networks: A deep learning framework for solving forward and inverse problems involving nonlinear partial differential equations},
  author={Raissi, Maziar and Perdikaris, Paris and Karniadakis, George E},
  journal={Journal of Computational physics},
  volume={378},
  pages={686--707},
  year={2019},
  publisher={Elsevier}
}

@article{shysheya2024conditional,
  title={On conditional diffusion models for PDE simulations},
  author={Shysheya, Aliaksandra and Diaconu, Cristiana and Bergamin, Federico and Perdikaris, Paris and Hern{\'a}ndez-Lobato, Jos{\'e} Miguel and Turner, Richard and Mathieu, Emile},
  journal={Advances in Neural Information Processing Systems},
  volume={37},
  pages={23246--23300},
  year={2024}
}

@article{lu2019deeponet,
  title={Deeponet: Learning nonlinear operators for identifying differential equations based on the universal approximation theorem of operators},
  author={Lu, Lu and Jin, Pengzhan and Karniadakis, George Em},
  journal={arXiv preprint arXiv:1910.03193},
  year={2019}
}

@article{karniadakis2021physics,
  title={Physics-informed machine learning},
  author={Karniadakis, George Em and Kevrekidis, Ioannis G and Lu, Lu and Perdikaris, Paris and Wang, Sifan and Yang, Liu},
  journal={Nature Reviews Physics},
  volume={3},
  number={6},
  pages={422--440},
  year={2021},
  publisher={Nature Publishing Group UK London}
}

@article{cuomo2022scientific,
  title={Scientific machine learning through physics--informed neural networks: Where we are and what’s next},
  author={Cuomo, Salvatore and Di Cola, Vincenzo Schiano and Giampaolo, Fabio and Rozza, Gianluigi and Raissi, Maziar and Piccialli, Francesco},
  journal={Journal of Scientific Computing},
  volume={92},
  number={3},
  pages={88},
  year={2022},
  publisher={Springer}
}

@misc{sudhakaran2022,
  author = {Sudhakaran, Shyam Sudhakaran},
  title = {hyper-nn},
  howpublished = {\url{https://github.com/shyamsn97/hyper-nn}},
  year = {2022},
}

@article{de2023physics,
  title={Physics-informed neural networks for data-driven simulation: Advantages, limitations, and opportunities},
  author={de la Mata, F{\'e}lix Fern{\'a}ndez and Gij{\'o}n, Alfonso and Molina-Solana, Miguel and G{\'o}mez-Romero, Juan},
  journal={Physica A: Statistical Mechanics and its Applications},
  volume={610},
  pages={128415},
  year={2023},
  publisher={Elsevier}
}

@article{huang2023limitations,
  title={On the limitations of physics-informed deep learning: Illustrations using first-order hyperbolic conservation law-based traffic flow models},
  author={Huang, Archie J and Agarwal, Shaurya},
  journal={IEEE Open Journal of Intelligent Transportation Systems},
  volume={4},
  pages={279--293},
  year={2023},
  publisher={IEEE}
}

@article{majumdar2023hyperlora,
  title={HyperLoRA for PDEs},
  author={Majumdar, Ritam and Jadhav, Vishal and Deodhar, Anirudh and Karande, Shirish and Vig, Lovekesh and Runkana, Venkataramana},
  journal={arXiv preprint arXiv:2308.09290},
  year={2023}
}

@article{zheng2024hypercan,
  title={HyperCAN: Hypernetwork-driven deep parameterized constitutive models for metamaterials},
  author={Zheng, Li and Kochmann, Dennis M and Kumar, Siddhant},
  journal={Extreme Mechanics Letters},
  pages={102243},
  year={2024},
  publisher={Elsevier}
}

@article{berman2024colora,
  title={CoLoRA: Continuous low-rank adaptation for reduced implicit neural modeling of parameterized partial differential equations},
  author={Berman, Jules and Peherstorfer, Benjamin},
  journal={arXiv preprint arXiv:2402.14646},
  year={2024}
}

@article{cho2023hypernetwork,
  title={Hypernetwork-based meta-learning for low-rank physics-informed neural networks},
  author={Cho, Woojin and Lee, Kookjin and Rim, Donsub and Park, Noseong},
  journal={Advances in Neural Information Processing Systems},
  volume={36},
  pages={11219--11231},
  year={2023}
}

@article{poli2020hypersolvers,
  title={Hypersolvers: Toward fast continuous-depth models},
  author={Poli, Michael and Massaroli, Stefano and Yamashita, Atsushi and Asama, Hajime and Park, Jinkyoo},
  journal={Advances in Neural Information Processing Systems},
  volume={33},
  pages={21105--21117},
  year={2020}
}

@article{chauhan2024brief,
  title={A brief review of hypernetworks in deep learning},
  author={Chauhan, Vinod Kumar and Zhou, Jiandong and Lu, Ping and Molaei, Soheila and Clifton, David A},
  journal={Artificial Intelligence Review},
  volume={57},
  number={9},
  pages={1--29},
  year={2024},
  publisher={Springer}
}

@article{brunton2024promising,
  title={Promising directions of machine learning for partial differential equations},
  author={Brunton, Steven L and Kutz, J Nathan},
  journal={Nature Computational Science},
  volume={4},
  number={7},
  pages={483--494},
  year={2024},
  publisher={Nature Publishing Group US New York}
}

@article{farenga2025latent,
  title={On latent dynamics learning in nonlinear reduced order modeling},
  author={Farenga, Nicola and Fresca, Stefania and Brivio, Simone and Manzoni, Andrea},
  journal={Neural Networks},
  pages={107146},
  year={2025},
  publisher={Elsevier}
}

@article{chen2018neural,
  title={Neural ordinary differential equations},
  author={Chen, Ricky TQ and Rubanova, Yulia and Bettencourt, Jesse and Duvenaud, David K},
  journal={Advances in neural information processing systems},
  volume={31},
  year={2018}
}

@article{chua1986double,
  title={The double scroll family},
  author={Chua, LEONO and Komuro, Motomasa and Matsumoto, Takashi},
  journal={IEEE transactions on circuits and systems},
  volume={33},
  number={11},
  pages={1072--1118},
  year={1986},
  publisher={IEEE}
}

@article{chua1993global,
  title={Global unfolding of Chua's circuit},
  author={Chua, Leon O},
  journal={IEICE Transactions on Fundamentals of Electronics, Communications and Computer Sciences},
  volume={76},
  number={5},
  pages={704--734},
  year={1993},
  publisher={The Institute of Electronics, Information and Communication Engineers}
}

@book{strogatz2018nonlinear,
  title={Nonlinear dynamics and chaos: with applications to physics, biology, chemistry, and engineering},
  author={Strogatz, Steven H},
  year={2018},
  publisher={CRC press}
}

@article{rossler1976equation,
  title={An equation for continuous chaos},
  author={R{\"o}ssler, Otto E},
  journal={Physics Letters A},
  volume={57},
  number={5},
  pages={397--398},
  year={1976},
  publisher={Elsevier}
}

@article{ha2016hypernetworks,
  title={Hypernetworks},
  author={Ha, David and Dai, Andrew and Le, Quoc V},
  journal={arXiv preprint arXiv:1609.09106},
  year={2016}
}

@inproceedings{li2020dhp,
  title={Dhp: Differentiable meta pruning via hypernetworks},
  author={Li, Yawei and Gu, Shuhang and Zhang, Kai and Van Gool, Luc and Timofte, Radu},
  booktitle={Computer Vision--ECCV 2020: 16th European Conference, Glasgow, UK, August 23--28, 2020, Proceedings, Part VIII 16},
  pages={608--624},
  year={2020},
  organization={Springer}
}

@inproceedings{alaluf2022hyperstyle,
  title={Hyperstyle: Stylegan inversion with hypernetworks for real image editing},
  author={Alaluf, Yuval and Tov, Omer and Mokady, Ron and Gal, Rinon and Bermano, Amit},
  booktitle={Proceedings of the IEEE/CVF conference on computer Vision and pattern recognition},
  pages={18511--18521},
  year={2022}
}

@inproceedings{chai2020supervised,
  title={Supervised and unsupervised learning of parameterized color enhancement},
  author={Chai, Yoav and Giryes, Raja and Wolf, Lior},
  booktitle={Proceedings of the IEEE/CVF Winter Conference on Applications of Computer Vision},
  pages={992--1000},
  year={2020}
}

@article{de2021hyperpinn,
  title={HyperPINN: Learning parameterized differential equations with physics-informed hypernetworks},
  author={de Avila Belbute-Peres, Filipe and Chen, Yi-fan and Sha, Fei},
  journal={The symbiosis of deep learning and differential equations},
  volume={690},
  year={2021}
}

@article{lorenz1963deterministic,
  title={Deterministic Nonperiodic Flow},
  author={Lorenz, Edward},
  journal={Journal of Atmospheric Sciences},
  volume={20},
  number={2},
  year={1963}
}

@book{hesthaven2016certified,
  title={Certified reduced basis methods for parametrized partial differential equations},
  author={Hesthaven, Jan S and Rozza, Gianluigi and Stamm, Benjamin and others},
  volume={590},
  year={2016},
  publisher={Springer}
}

@article{hornik1991approximation,
  title={Approximation capabilities of multilayer feedforward networks},
  author={Hornik, Kurt},
  journal={Neural networks},
  volume={4},
  number={2},
  pages={251--257},
  year={1991},
  publisher={Elsevier}
}

@inproceedings{simsek2021geometry,
  title={Geometry of the loss landscape in overparameterized neural networks: Symmetries and invariances},
  author={Simsek, Berfin and Ged, Fran{\c{c}}ois and Jacot, Arthur and Spadaro, Francesco and Hongler, Cl{\'e}ment and Gerstner, Wulfram and Brea, Johanni},
  booktitle={International Conference on Machine Learning},
  pages={9722--9732},
  year={2021},
  organization={PMLR}
}

@article{brea2019weight,
  title={Weight-space symmetry in deep networks gives rise to permutation saddles, connected by equal-loss valleys across the loss landscape},
  author={Brea, Johanni and Simsek, Berfin and Illing, Bernd and Gerstner, Wulfram},
  journal={arXiv preprint arXiv:1907.02911},
  year={2019}
}

@article{ma2001finance,
  title={Study for the bifurcation topological structure and the global complicated character of a kind of nonlinear finance system (I)},
  author={Ma, Junhai and Chen, Yushu},
  journal={Applied Mathematics and Mechanics},
  volume={22},
  number={11},
  pages={1240--1251},
  year={2001},
  publisher={Springer}
}

\end{document}